\documentclass[journal ]{new-aiaa}
\usepackage[utf8]{inputenc}
\usepackage{textcomp}

\usepackage{mathtools}
\usepackage[version=4]{mhchem}
\usepackage{siunitx}
\usepackage{longtable,tabularx}
\newcommand{\bx}{\textbf{\emph{x}}}
\newcommand{\bPhi}{\boldsymbol{\Phi}}

\newcommand{\beps}{\boldsymbol{\epsilon}}
\newcommand{\ba}{\textbf{\emph{a}}}
\newcommand{\bb}{\textbf{\emph{b}}}

\newcommand{\bd}{\textbf{\emph{d}}}
\newcommand{\bm}{\textbf{\emph{m}}}
\newcommand{\bl}{\boldsymbol{\ell}}

\newcommand{\be}{\textbf{\emph{e}}}
\newcommand{\bt}{\textbf{\emph{t}}}

\newcommand{\bw}{\textbf{\emph{w}}}

\newcommand{\br}{\textbf{\emph{r}}}

\newcommand{\bu}{\textbf{\emph{u}}}
\newcommand{\bv}{\textbf{\emph{v}}}
\newcommand{\by}{\textbf{\emph{y}}}

\newcommand{\bp}{\textbf{\emph{p}}}
\newcommand{\bq}{\textbf{\emph{q}}}
\newcommand{\bA}{\textbf{\emph{A}}}
\newcommand{\bC}{\textbf{\emph{C}}}
\newcommand{\bD}{\textbf{\emph{D}}}
\newcommand{\bF}{\textbf{\emph{F}}}

\newcommand{\bI}{\textbf{\emph{I}}}
\newcommand{\bP}{\textbf{\emph{P}}}
\newcommand{\bK}{\textbf{\emph{K}}}
\newcommand{\bH}{\textbf{\emph{H}}}
\newcommand{\bR}{\textbf{\emph{R}}}

\newcommand{\bk}{\textbf{\emph{k}}}
\newcommand{\bB}{\textbf{\emph{B}}}
\newcommand{\bT}{\textbf{\emph{T}}}
\newcommand{\bW}{\textbf{\emph{W}}}
\newcommand{\bE}{\textbf{\emph{E}}}
\newcommand{\bM}{\textbf{\emph{M}}}
\newcommand{\bL}{\textbf{\emph{L}}}
\newcommand{\bS}{\textbf{\emph{S}}}

\newcommand{\bxi}{\boldsymbol{\xi}}
\newcommand{\bnu}{\boldsymbol{\nu}}

\newcommand{\boldeta}{\boldsymbol{\eta}}

\newcommand{\PP}{{\mathbb P}}
\newcommand{\RR}{{\mathbb R}}

\newcommand{\hide}[1]{}

\title{Absolute Triangulation Algorithms for Space Exploration}

\author{S\'{e}bastien Henry\footnote{Graduate Research Assistant, Guggenheim School of Aerospace Engineering.} and John A. Christian\footnote{Associate Professor, Guggenheim School of Aerospace Engineering. Associate Fellow AIAA.} }

\affil{Georgia Institute of Technology, Atlanta, GA 30332}

\begin{document}

\maketitle

\begin{abstract}
Images are an important source of information for spacecraft navigation and for three-dimensional reconstruction of observed space objects. Both of these applications take the form of a triangulation problem when the camera has a known attitude and the measurements extracted from the image are line of sight (LOS) directions. This work provides a comprehensive review of the history and theoretical foundations of triangulation. A variety of classical triangulation algorithms are reviewed, including a number of suboptimal linear methods (many LOS measurements) and the optimal method of Hartley and Sturm (only two LOS measurements). It is shown that the optimal many-measurement case may be solved without iteration as a linear system using the new Linear Optimal Sine Triangulation (LOST) method. Both LOST and the polynomial method of Hartley and Sturm provide the same result in the case of only two measurements. The various triangulation algorithms are assessed with a few numerical examples, including planetary terrain relative navigation, angles-only optical navigation at Uranus, 3-D reconstruction of Notre-Dame de Paris, and angles-only relative navigation.
\end{abstract}

\section{Introduction}

Cameras, telescopes, and other optical instruments are an important source of information for modern space vehicles \cite{Christian:2019b}. Information extracted from the images produced by these sensors may be used for many different purposes, with shape modeling \cite{Gaskell:2008,Palmer:2022} and spacecraft navigation \cite{Owen:2008,Christian:2021} being two of the most common. In both of these applications, the objective is achieved by exploiting the geometry that relates the three-dimensional (3-D) location of objects in the observed scene to their apparent pixel locations in a digital image.

There are a variety of observables one might extract from an image and a variety of ways in which these observables may be interpreted. Perhaps the simplest situation is when one obtains line of sight (LOS) measurements from an optical instrument to a point in 3-D space. Oftentimes, the absolute location of either the camera or the observed point is known. When the camera location is known and the observed point's location is unknown, then this is a shape modeling (i.e., 3-D reconstruction) problem. Conversely, when the observed point's location is known and the camera location is unknown, this is a navigation problem. The geometry of these two problems is identical and the algorithms developed for one are equally applicable to the other. For the sake of brevity and clarity, this work primarily develops algorithms from the navigation perspective (though all results are also valid for 3-D reconstruction).

Point localization using a collection of LOS measurements (or angles) is generically referred to as triangulation [for the specific application of spacecraft navigation this is also sometimes called ``angles-only'' optical navigation (OPNAV)].
In recent years, there has been substantial interest in the utility of triangulation methods for spacecraft localization in cislunar space using sightings of other well-known satellites \cite{Driedger:2021,Bradley:2020} or for interplanetary exploration using sightings of asteroids/planets \cite{Karimi:2015,Broschart:2019,Bradley:2020,Hinga:2020,Franzese:2022}. The concepts of triangulation, however, apply to a much broader class of spaceflight navigation problems. For example, some of these other applications include terrain relative navigation (TRN) with known landmarks \cite{Cheng:2003}, angles-only relative navigation (RelNav) \cite{Woffinden:2009}, and star-based interstellar navigation \cite{McKee:2022}.

This work reviews the present state of triangulation algorithms with applicability to space exploration. Special consideration is given to the relationships between the different algorithms, and some algorithms are shown to be equivalent under certain conditions. Many of the most popular methods (e.g., the Direct Linear Transform) are geometrically exact, but do not produce statistically optimal localization estimates when provided noisy LOS measurements. When there are only two LOS measurements, the classic method of Hartley and Sturm  \cite{Hartley:1997} finds the statistically optimal answer as the solution to a polynomial of degree six. This work shows how Hartley and Sturm's solution can be simplified to a polynomial of degree two when all of the LOS measurements come from a single image (as happens with monocular vision-based navigation). Unfortunately, the extension of Hartley and Sturm's method to three or more LOS measurements is not practical \cite{Stewenius:2005} and past work has relied on iterative numerical schemes to perform statistically optimal triangulation with many measurements. To address this problem, the new Linear Optimal Sine Triangulation (LOST) method is introduced in Section~\ref{Sec:LOST}. The LOST method solves the statistically optimal triangulation problem without iteration as the solution to a linear system. When only two measurements are available the error covariance of the triangulated point is exactly the same for the LOST method and the Hartley and Sturm method. Moreover, as compared to the Hartley and Sturm method, the LOST method is faster (small linear system instead of a polynomial of degree six) and scalable (can directly find optimal solution for any number of measurements as a linear system). This work concludes with a number of examples to illustrate the similarities, advantages, and disadvantages of the various triangulation methods.

\section{Background and the Lexicon of Triangulation}
\label{Sec:Lexicon}

The idea of using optical observations to find an object's location is not new. A sophisticated understanding of the underlying geometry was known in antiquity, and has been used for over two thousand years by navigators, cartographers, astronomers, civil engineers, and many others. Unfortunately, many critical ideas have been forgotten and rediscovered over the centuries---with this apparent forgetfulness being strangely prevalent today. Harry S. Truman famously said ``There is nothing new in the world except the history you do not know.'' It will soon become apparent that Truman's astute observation certainly applies to the field of triangulation.

\begin{figure}[b!]
\centering
\includegraphics[width=0.3\columnwidth,trim=0in 0in 0in 0in,clip]{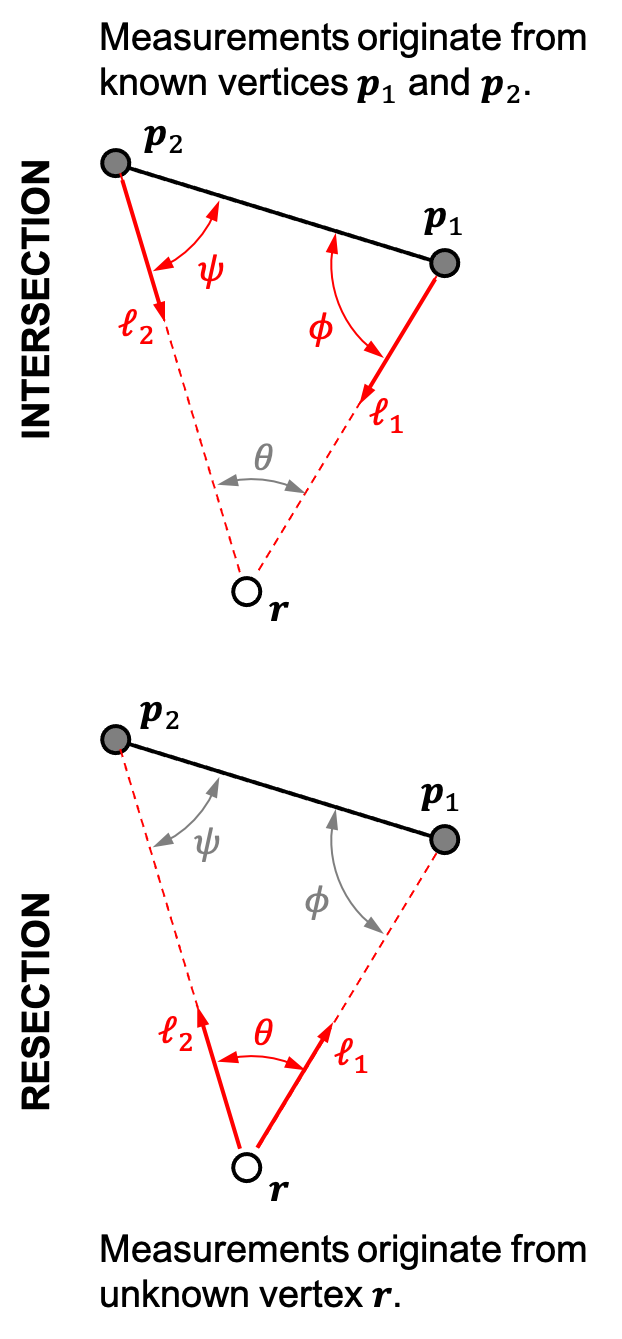}
	\caption{Illustration of intersection (top) and resection (bottom) forms of the triangulation problem. The location of vertices $\bp_1$ and $\bp_2$ are known and the objective is to solve for the unknown location of vertex $\br$. Measurements are shown in red.}
	\label{fig:IntersectionResection}
\end{figure}

To begin, let the term \emph{triangulation} refer to the task of localizing a point using the angles that describe a triangle (see Fig.~\ref{fig:IntersectionResection}). In the simplest formulation there are two points of known position (forming two vertices of the triangle), such that the angles describing a triangle uniquely determine the position of an unknown point (the third vertex of the triangle). Note that the ``tri'' in triangulation refers to the use of triangles and not to the number of points. Thus, the term triangulation remains precise when observations of many known points are used to localize a single unknown point, since the solution then consists of many triangles.

\begin{figure}[b!]
\centering
\includegraphics[width=0.35\columnwidth,trim=0in 0in 0in 0in,clip]{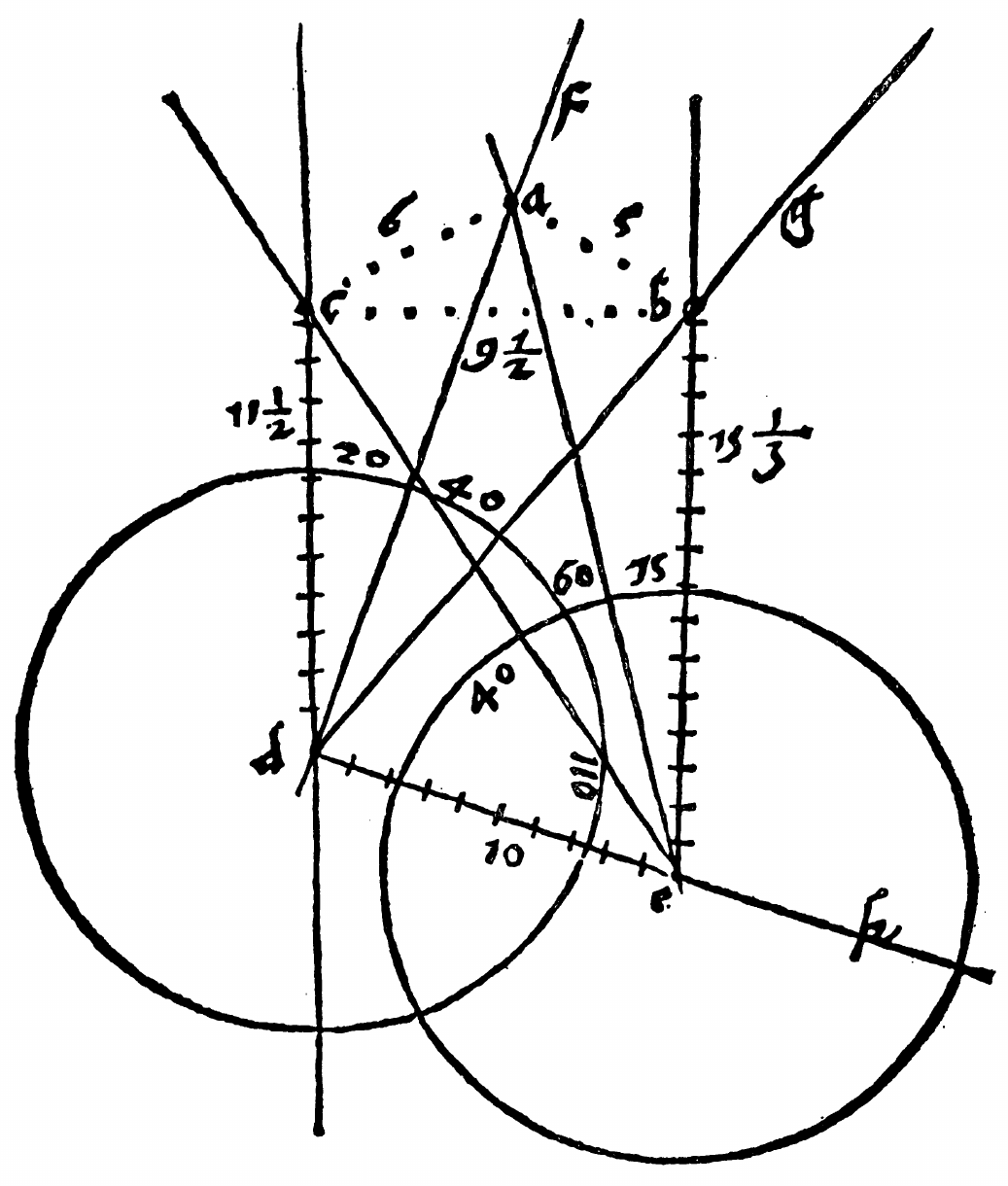}
	\caption{Illustration by G. Frisius (c. 1533) showing how to determine the location of unknown (or inaccessible) points $a,b,c$ using only the bearings from two known points ($d$ and $e$) separated by a known baseline. This is an example of the intersection form of the triangulation problem. Reproduced from Ref.~\cite{Frisius:1533} as it appears at the end of Ref.~\cite{Pogo:1935}.}
	\label{fig:FrisiusTri}
\end{figure}

The origins of triangulation are difficult to establish, largely due to the loss of primary sources from antiquity. Indeed, attribution of primacy seems to depend as much on whose works survived to the modern era as it does on who was truly the first to ``discover'' an idea. Regardless, a few things are known. The concept of using triangles to estimate distance and size was known as early as the 16th century BC, as can be seen by the problem/solution sets (especially P57-P60) in the Rhind Mathematical Papyrus \cite{Chace:1927} from the Second Intermediate Period of ancient Egypt. Sophistication of understanding increased significantly by 6th centuty BC, when the Greek mathematician Thales purportedly used triangles to estimate the height of the Egyptian pyramids and the distance to ships at sea \cite{Molinsky:2015}. All of the geometric principles necessary for point localization by triangulation were known by the 3rd century BC, when they were clearly laid out in Euclid's \emph{Elements} \cite{EuclidElements,Archibald:1950}. Moreover, by the 2nd century BC, Hipparchus (amongst the first known to posses a trigonometric table \cite{Toomer:1974b}) used triangulation principles to estimate the distance to the Sun and Moon (as described by Ptolemy in the Almagest Book V) \cite{Almagest,Toomer:1974}. Thus, there can be little doubt that the classical Greek geometers were quite familiar with how triangles and a known baseline could be used to infer the location of a point at a distance. While this line of study continued to flourish in the Islamic world during the first millennium CE \cite{Lorch:2002}, the knowledge was lost to European scholars for over a thousand years (until Euclid's \emph{Elements} was translated from Arabic into Latin in the 12th century CE \cite{Lorch:2002,Archibald:1950}). The modern understanding of triangulation (e.g., Fig.~\ref{fig:FrisiusTri}) can be traced to a booklet published in 1533 by Gemma Frisius \cite{Frisius:1533,Pogo:1935,Haasbroek:1968}. Notable improvements in the practice were made by Snellius in 1617 as part of his effort to measure the circumference of the Earth \cite{Haasbroek:1968,Snellius:1617}, and triangulation was put into a least squares framework by Gauss himself while conducting a survey of Hanover in the 1820s \cite{Breitenberger:1984, Gauss:1828}. At a conceptual level, the work of Frisius, Snellius, and Gauss brings us to the modern era---where new sensors, computational resources, and applications have made room for additional innovations.

The modern triangulation problem takes on one of two forms: intersection or resection \cite{McCaw:1918}. This is illustrated in Fig.~\ref{fig:IntersectionResection}. The first form is \emph{intersection}, which occurs when angle measurements collected by observers at two (or more) known locations are used to estimate the location of an unknown point. The intersection form of triangulation has many practical applications, such as satellite orbit determination \cite{Bernard:2018,Chen:2021} and 3-D scene reconstruction from multiple images \cite{Hartley:1997,Pollefeys:2000,Snavely:2008}. The second form is \emph{resection}, which occurs when an observer estimates its unknown location using angle measurements to two (or more) points at known locations. The resection form of triangulation describes the vehicle (e.g., robot, ship, aircraft, spacecraft) localization problem that commonly appears in navigation applications \cite{Broschart:2019,Poirot:1976}. The intersection and resection problems have the same underlying geometry, with the primary difference being whether the triangle vertex of unknown location is the observer or is being observed.

Both the intersection and resection forms of the triangulation problem find the location of the unknown triangle vertex (denoted as $\br$ in Fig.~\ref{fig:IntersectionResection}) relative to the locations of the known triangle vertices (denoted as $\bp_1$ and $\bp_2$ in Fig.~\ref{fig:IntersectionResection}). Thus, one typically chooses to express the known locations of $\bp_1$ and $\bp_2$ in a common frame and then triangulation usually results in an estimate of $\br$ this same frame. The frame in which all this occurs is generically called the \emph{localization frame}. The choice of the localization frame depends on the application (see Section~\ref{Sec:Applications} for three different choices), but common examples for spacecraft localization include an inertial frame (e.g., International Celestial Reference Frame \cite{Feissel:1998, Ma:1998,Fey:2015}), planet-fixed frame, spacecraft-fixed frame, or the orbital local-vertical, local-horizontal (LVLH) frame.

The angles in the triangulation problem often come from optical instruments.
For example, in land surveying and laboratory metrology applications the optical instrument is commonly a theodolite. In spacecraft navigation applications the angles usually come from camera or telescope images. The angles themselves are often referred to as \emph{bearings} and come in pairs for 3-D problems (e.g., azimuth and elevation, right ascension and declination). Regardless of the application, the angle measurements (i.e., the bearings) produced by such optical instruments describe the direction from the sensor to an observed point (i.e., the direction from one triangle vertex to another). This direction corresponds to a straight line connecting two points that is commonly referred to as the \emph{line of sight} (LOS).

Crucially important is the frame(s) in which the measured LOS directions may be expressed. Most optical instruments fundamentally measure the LOS direction in a frame fixed to the sensor, referred to here as the \emph{sensor frame}. If the orientation of this sensor frame relative to the localization frame is known, then it is possible to transform the LOS direction from the sensor frame (where it is measured) into the localization frame (where the locations of the reference points are known). When expressed in the localization frame, these measurements are referred to as \emph{absolute} LOS directions. For usual optical instruments, the absolute LOS is coincident with the \emph{line of position} (LOP)---where the LOP describes a straight line in the localization frame on which the unknown point must lie. Moreover, following the terminology of Kaplan \cite{Kaplan:2011}, define \emph{absolute triangulation} as triangulation performed exclusively with absolute LOS directions and having a solution at the intersection of two (or more) LOPs. Most spacecraft have excellent attitude knowledge (e.g., from star trackers \cite{Liebe:1995,Liebe:2002,Christian:2021star}) and so the absolute triangulation problem occurs frequently in spaceflight applications. Absolute triangulation is the primary topic of this manuscript and numerous algorithms are presented to accomplish this task (see Sections~\ref{Sec:TrigSolutions} and \ref{Sec:OptimalTri}).

\begin{figure}[b!]
\centering
\includegraphics[width=0.3\columnwidth,trim=0in 0in 0in 0in,clip]{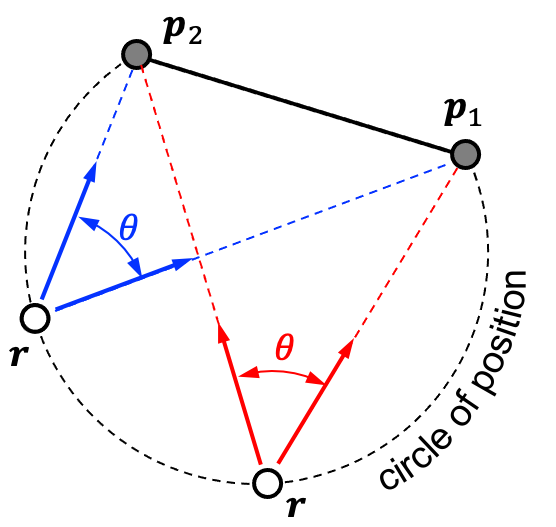}
	\caption{Illustration of the 2-D resection problem with unknown instrument orientation. The observer lies somewhere on a circle of position (COP) when only the apparent angle between the LOS directions to two known stations is measured.}
	\label{fig:ResectionCOP}
\end{figure}

\begin{figure}[t!]
\centering
\includegraphics[width=0.35\columnwidth,trim=0in 0in 0in 0in,clip]{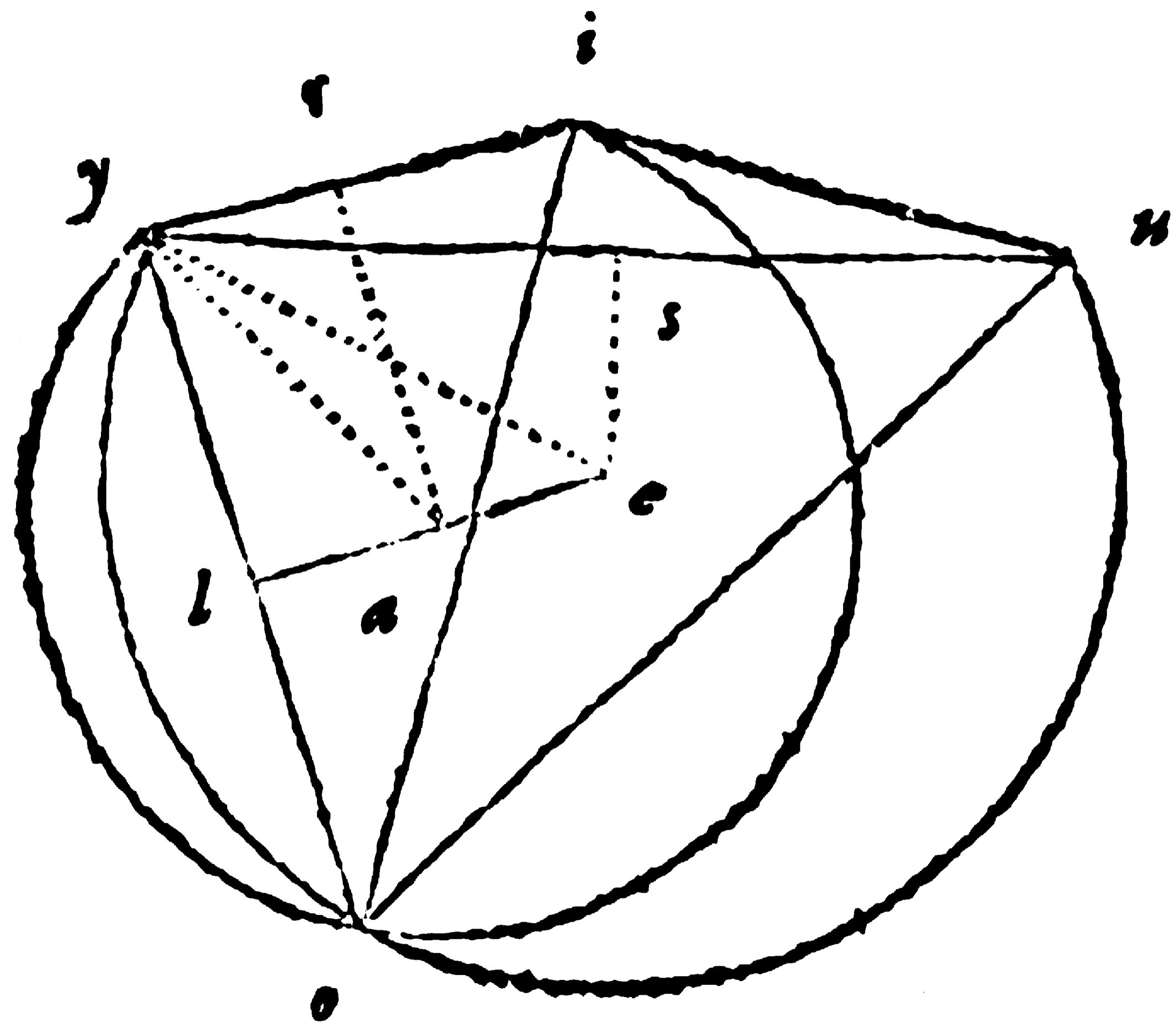}
	\caption{Illustration by W. Snellius (c. 1617) showing how to determine the location of unknown observer point $o$  using only measured angles $\angle yoi$ and $\angle you$ to the three known points $y,i,u$. This is a 2-D example of the resection form of the triangulation problem. Reproduced from Ref.~\cite{Snellius:1617}, p. 204, and digitally restored by the authors of this manuscript.}
	\label{fig:SnelliusTri}
\end{figure}

Although absolute triangulation is the subject of the technical discussions that follow, additional insight may be gained (and common mistakes avoided) by also understanding what absolute triangulation is not. Thus, consider the alternative case where the orientation of the optical instrument is not known and the absolute LOS directions cannot be found---instead, only the triangle interior angles may be measured. In this case, the differences between the intersection and resection forms of the triangulation problem are especially important. For example, consider a two-dimensional (2-D) system. In this example, the intersection form of the triangulation problem may be solved with a single triangle using observations of the unknown point from two known observer locations. This is because measurement of the angles $\phi$ and $\psi$ (see top diagram in Fig.~\ref{fig:IntersectionResection}) produce two LOS directions relative to the known observer baseline that intersect at a unique 2-D point. This, however, is not the case for the resection problem with two known points. If absolute orientation is unknown and the only measurement is the angle $\theta$, then the observer's unknown location is somewhere along a circular arc (see Fig.~\ref{fig:ResectionCOP}) called the \emph{circle of position} (COP). This is a well-known result. The observer position may be made unique in the 2-D resection problem by measuring angle pairs between three points, which is classically referred to as the \emph{three-point resection problem} or the \emph{Snellius-Pothenot problem} (named after Snellius's seminal 1617 work mentioned earlier \cite{Snellius:1617,Haasbroek:1968}). As an example, consider Fig.~\ref{fig:SnelliusTri} where Snellius documented how he found the location of point $o$ (roof of his house) using only measured angles $\angle yoi = 32^{\circ} 57'$ and $\angle you = 64^{\circ} 40'$ to the known points $y$ (Pieterskerk), $i$ (Town Hall), $u$ (Hooglandse Kerk) in Leiden. Clearly shown are the two COPs (one from the baseline $yi$ and the other from the baseline $yu$) that intersect at the unique location of $o$. Even after 400 years, the Snellius-Pothenot problem remains an important problem in land surveying, with various algorithms still being published today \cite{Hassan:2002}.

Spacecraft and other aerospace vehicles operate in a 3-D world, and so it is instructive to expand upon the intuitive 2-D results from the Snellius-Pothenot problem. 
The 3-D case of resection, where a sensor of unknown attitude lies at the point of unknown location, has sometimes been called the \emph{space resection problem} in land surveying and photogrammetry applications. The astute roboticist, engineer, or computer scientist will recognize this as identical to the \emph{perspective-n-point (PnP) problem}. When $n=3$ one has the famous P3P problem, which admits 4 unique solutions with a direct (non-iterative) method existing as early as 1841 \cite{Grunert:1841}. Since then, driven in large part by interest from the computer vision community, there have been great improvements in algorithms for the P3P \cite{Haralick:1994,Gao:2003,Tanygin:2014}, P4P \cite{PascualEscudero:2021}, and PnP \cite{Hartley:2003,Quan:1999,Ansar:2003,Lepetit:2009} problems. Similar problems (i.e., position and attitude from LOS measurements to points) have been considered within the space community for some time \cite{Calhoun:1995,Crassidis:2000}. Note that the PnP problem simply describes a particular means of solving the \emph{pose estimation problem}. Pose estimation is the general problem of finding the position and attitude of an object, which may be achieved with optical observations of points (e.g., PnP problem) or with other sensors (e.g., LIDAR point cloud registration \cite{Besl:1992,Yang:2021}). Finally, the reader should note that the pose estimation problem cannot be solved for the intersection form of the triangulation problem since the measured angles (collected by observers at the triangle's known points) do not depend on the orientation of the unknown point. Hence, the unknown point's location may be found, but the attitude of the object residing at the unknown point is unobservable by triangulation alone.

\begin{figure}[bt!]
\centering
\includegraphics[width=0.4\columnwidth,trim=0in 0in 0in 0in,clip]{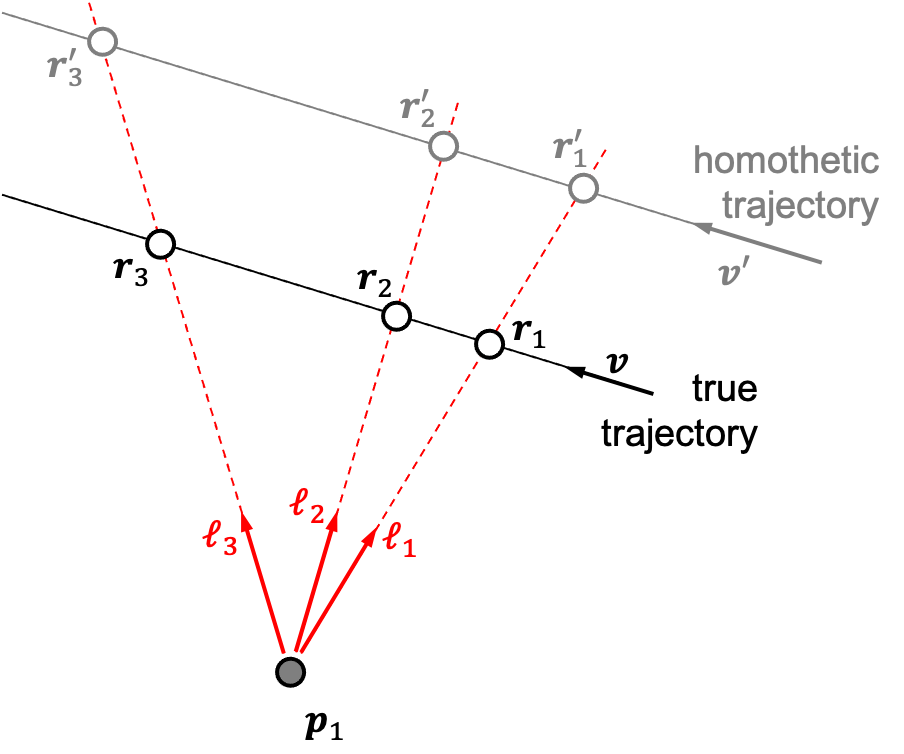}
	\caption{If the motion between two particles is governed by linear dynamics, then then LOS measurements from one particle to the other can be explained by an infinite number of homothetic trajectories.}
	\label{fig:HomotheticTraj}
\end{figure}

The triangulation problems discussed so far imply that the point of unknown location is either (1) stationary or (2) that the LOS measurements are obtained at the same instant in time. Spacecraft, however, are not stationary vehicles. Moreover, while it is possible in many situations to obtain (nearly) simultaneous LOS measurements, this is not always the case. Fortunately, the tools of triangulation may also be used for a moving observer---a concept referred to here as \emph{dynamic triangulation}. The ideas of dynamic triangulation were pioneered for maritime applications, such as ship navigation \cite{Poirot:1976} and bearings-only target motion analysis (BO-TMA) \cite{Nardone:1984}. The topic gained much attention after the wide-spread deployment of radar systems in the 1940s and 1950s, where it immediately became apparent that a ship moving along a straight line cannot navigate by bearings to a single station \cite{Harper:1955}. Likewise, a single observer that is either stationary or moving along a straight line cannot uniquely determine the trajectory of a target ship with bearings alone. In these cases, it is impossible to distinguish between the object's true trajectory and an infinite number of \emph{homothetic trajectories}. Any one of the homothetic trajectories is related to the true trajectory by a homothety \cite{Casey:1888}, which is described by a contraction or expansion of the true trajectory about the homothetic center. Thus, the observer is left with scale ambiguity on the trajectory, as illustrated on  Fig.~\ref{fig:HomotheticTraj}. The usual solution is to collect LOS observations from multiple stations or to maneuver the observer \cite{Lindgren:1977,Nardone:1981}. 

Dynamic triangulation plays a central role in spacecraft ``angles-only'' relative navigation (RelNav). Recall that two spacecraft in near circular orbits that are close to one another exhibit nearly linear dynamics (e.g., as described with the Clohessy-Wiltshire equations \cite{Clohessy:1960}). Due to the linear dynamics, the exact same observability challenges found for the BO-TMA problem were found to exist for problems of spacecraft angles-only RelNav \cite{Woffinden:2007,Woffinden:2009}. Unsurprisingly, the solutions (e.g., maneuvering the observer) for space applications are largely the same as for maritime applications. Indeed, many of the novel approaches for spacecraft angles-only RelNav are actually rediscoveries of the dynamic triangulation techniques known for decades within the context maritime BO-TMA.

\section{Projective Geometry, Camera Models, and Images}

Proper development and analysis of triangulation methods requires a model of the instrument used to acquire the LOS measurements. This work assumes an optical instrument obeying projective geometry and its mathematical representation is referred to here as the \emph{camera model}. The subsections that follow first develop the camera model from the perspective of algebraic projective geometry \cite{Semple:1952} and then consider the measurement statistics.

\subsection{Ideal Camera Model}
\label{Sec:CamModel}
Suppose a spacecraft at position $\br$ views an object located as position $\bp_i$ with an optical instrument. In many cases the sensor only provides a measurement of the LOS direction $\bl_i$
\begin{equation}
    \label{eq:BasicPinholeNoFrame}
    \bl_i \propto \bp_i - \br
\end{equation}
Here, $\bl_i$ is of arbitrary scale and describes a point in $\PP^2$. For additional information on projective space (e.g., $\PP^2,\PP^3,\PP^n$), see Refs.~\cite{Christian:2021,Hartley:2003,Gallier:2011}. The relationship in Eq.~\eqref{eq:BasicPinholeNoFrame} is true in any consistent frame. 

When the optical instrument is a camera (or telescope) obeying projective geometry, the LOS direction $\bl_i$ is produced by back-projecting a ray out of the camera that corresponds to a specific pixel coordinate $\bu^T_i = [u_i,v_i]$ in a digital image. 
Although formation of the LOS direction $\bl_i$ may be accomplished in any frame, it is convenient to choose the sensor frame (which, in this case, is the \emph{camera frame}). The most popular camera frame convention---and the one chosen here---places the origin at the center of the camera's entrance pupil (i.e., apparent pinhole location to an observer outside the camera). The camera frame $z$-axis is along the camera boresight direction (positive out of the camera). The orientation of the $x$- and $y$-axes vary in the literature, but this work adopts the convention of Ref.~\cite{Christian:2021} that aligns the $x$-axis with the image $u$-direction and the $y$-axis with the image $v$-direction. Details and diagrams are provided in Ref.~\cite{Christian:2021}.

Now, in the case where the camera attitude is known, it is possible to express everything in the camera frame. Different normalizations of the direction $\bl_i$ in the camera frame (denoted by the subscript $C$) describe the pinhole camera geometry \cite{Christian:2021}, such that
\begin{equation}
    \label{eq:CameraPinhole}
    \begin{bmatrix} x_i \\ y_i \\ 1 \end{bmatrix} = \bar{\bx}_i \propto \ba_{C_i}
    \propto \bl_{C_i} \propto \bp_{C_i} - \br_C
\end{equation}
where $\ba_{C_i} = \bl_{C_i} / \| \bl_{C_i} \|$ is a unit vector pointing from the camera to the observed object and where $\bx^T_i = [x_i,y_i]$ are the 2-D coordinates where the direction $\bl_{C_i}$ pierces the camera frame's $z=1$ plane (which is sometimes called the image plane). The 3-D camera frame coordinates of the $z=1$ image plane intersection point is the same as $\bx_i$ written in homogeneous coordinates, which is denoted here as $\bar{\bx}^T_i = [\bx^T_i, 1]$. As will become apparent, efficiently moving between the different choices of normalization in Eq.~\eqref{eq:CameraPinhole} provides numerous algorithmic benefits and insights. 

Contemporary space cameras most commonly produce a digital image. A 2-D point $\bx^T_i = [x_i,y_i]$ on the image plane corresponds to a 2-D point $\bu^T_i = [u_i,v_i]$ in the digital image. Writing $\bu_i$ in homogemeous coordinates, $\bar{\bu}^T_i = [\bu^T_i, 1]$, the affine transformation between the two image point representations is given by,
\begin{equation}
    \label{eq:IPtoPixAffine}
    \begin{bmatrix}
        u_i \\ v_i \\ 1
    \end{bmatrix} = 
    \bar{\bu}_i = \bK \bar{\bx}_i
\end{equation}
where $\bK$ is the so-called \emph{camera calibration matrix} (or, sometimes, the camera intrinsics matrix),
\begin{equation}
    \bK = 
    \begin{bmatrix}
        d_x & \alpha & u_p \\ 0 & d_y & v_p \\ 0 & 0 & 1
    \end{bmatrix}
\end{equation}
In most cases $\alpha \approx 0$ and this transformation simply shifts the origin ($u_p$ and $v_p$) and converts from units of length to units of pixels ($d_x$ and $d_y$). More details are provided in Ref.~\cite{Christian:2021}.

The matrix $\bK$ is almost always known for space cameras. The five calibration terms contained within this matrix may be estimated prior to launch, and it is straightforward to recalibrate the camera on-orbit using images of starfields \cite{Christian:2021star,Christian:2016,Bos:2020}. When $\bK$ is known, the apparent pixel coordinates of a point (e.g., star, distant celestial body, surface landmark) in an image may be transformed to the camera frame LOS direction $\bl_{C_i}$ as
\begin{equation}
    \label{eq:PixCoordToLOS}
    \bl_{C_i} \propto \ba_{C_i} \propto \bar{\bx}_i = \bK^{-1} \bar{\bu}_i
\end{equation}
where the matrix inverse $\bK^{-1}$ may be analytically computed as discussed in Ref.~\cite{Christian:2021}.

\subsection{LOS Representations and Measurement Models}
\label{eq:MeasModels}
The choice of LOS representation and its covariance model is one of the most important decisions the designer of a triangulation algorithm must make. A good choice will often allow for an elegant and statistically optimal solution. Conversely, a poor choice will lead to a complicated (and usually iterative) solution. 

The most common choice of LOS representation for spacecraft navigation is a unit vector. The popularity of this choice can be traced to the great success of unit vectors for star-based attitude determination. The problem of spacecraft attitude determination from unit vectors in correspondence was posed by Grace Wahba in 1965 \cite{Wahba:1965} and a number of elegant solutions have been proposed since then \cite{Farrell:1966,Shuster:1981,Markley:1988,Christian:2021star}. These algorithms are the workhorse of modern star-based attitude determination and play a central role in commercially available star trackers. 

Proceeding in a similar way, consider the LOS direction represented by the unit vector $\ba$. Let the noisy measurement of that unit vector be $\tilde{\ba} = \ba + \boldeta$, where $\boldeta \sim \mathcal{N}(\textbf{0}_{3 \times 1}, \bR_{QMM})$. In what is now known as the QUEST measurement model (QMM), the covariance of a unit vector (to first order) with isotropic direction uncertainty is \cite{Shuster:1981}
\begin{equation}
    \label{eq:QUESTcov}
    \bR_{QMM} = E[\boldeta \boldeta^T] = \sigma_{\theta}^2 \left( \bI_{3 \times 3} - \ba \ba^T \right)
\end{equation}
where $\sigma_{\theta}^2$ is the variance of the unit vector pointing direction (in radians). This is true regardless of the frame in which $\ba$ is expressed. Since both $\ba$ and $\tilde{\ba}$ are constrained to be unit vectors, the covariance matrix $\bR_{QMM}$ is rank two and describes a plane tangent to the unit sphere at the point $\ba$. Importantly, solutions to Wahba’s problem are known to provide a maximum likelihood estimate (MLE) of the attitude when unit vector measurements are assumed to follow the QMM \cite{Shuster:1989}. This justifiably led to widespread adoption of the unit vector and QMM convention.

Building on past successes with spacecraft attitude determination, it would seem logical to represent LOS directions for spacecraft triangulation with the unit vector $\ba$ and covariance from Eq.~\eqref{eq:QUESTcov}. However, these are generally poor choices in practice. There are two problems. The first problem is that the QMM constrains $\boldeta$ to lie on the unit sphere's tangent plane such that $\tilde{\ba}$ is not a unit vector (specifically, $\| \tilde{\ba} \| > 1$). Although it is a second-order effect, these difficulties with additive noise are well-known and have led some to consider a multiplicative error model \cite{Mortari:2009}. Avoiding the (usually) artificial unit vector normalization eliminates the need for the multiplicative error model. The second (and more important) problem is that the QMM from Eq.~\eqref{eq:QUESTcov} is not the correct noise model when the sensor is a camera or telescope. 

To build the correct error covariance for an LOS measurement, begin by considering how these measurements are constructed from a digital image.
Recall that LOS directions originate as image points with pixel coordinates $\{u_i,v_i\}_{i=1}^n$. Depending on the type of object and image processing technique used (e.g., centroiding of an unresolved celestial body \cite{Kaasalainen:2004}, cross-correlation of a terrain maplet \cite{Gaskell:2008,Olds:2022}, localization of a SIFT keypoint \cite{Lowe:2004}), there will be some uncertainty in the pixel coordinates $[u_i,v_i]$ used to create the LOS measurement. Therefore, let the pixel coordinate $\bu^T_i = [u_i,v_i]$ produce the camera frame LOS measurement $\bl_{C_i}$ using Eq.~\eqref{eq:PixCoordToLOS}. Describe the zero-mean, Gaussian measurement noise in the image point location by $\bnu_i \sim \mathcal{N}(0,\bR_{\bu_i})$, such that
\begin{equation}
    \label{eq:NoisyPixel2D}
   \tilde{\bu}_i = \bu_i + \bnu_i
\end{equation}
where the measurement covariance $\bR_{\bu_i} = E[\bnu_i \bnu_i^T]$ is a $2 \times 2$ matrix. 
Thus, following Eq.~\eqref{eq:PixCoordToLOS},
\begin{equation}
    \label{eq:LOSnoisy}
   \tilde{\bl}_{C_i} \propto \tilde{\bar{\bx}}_i = \bK^{-1} \tilde{\bar{\bu}}_i = \bK^{-1}\bar{\bu}_i + \bK^{-1}\bS^T \bnu_i
\end{equation}
where 
\begin{equation}
    \begin{bmatrix}
        \tilde{u}_i \\ \tilde{v}_i \\ 1
    \end{bmatrix} = 
   \tilde{\bar{\bu}}_i = \bar{\bu}_i + \bS^T \bnu_i
\end{equation}
\begin{equation}
    \label{eq:DefBoldS}
   \bS = \begin{bmatrix} \bI_{2 \times 2} & \textbf{0}_{2 \times 1} \end{bmatrix}
\end{equation}
Moreover, defining $\bw_i= \bK^{-1}\bS^T \bnu_i$, one has
\begin{equation}
    \label{eq:xbarnoisy}
   \tilde{\bar{\bx}}_i = \bar{\bx}_i + \bw_i
\end{equation}
Thus, making use of Eq.~\eqref{eq:LOSnoisy}, it follows that the error covariance of $\tilde{\bar{\bx}}$ is
\begin{equation}
    \label{eq:DefRxbxb}
   \bR_{\bar{\bx}_i} = E[\bw_i \bw^T_i] =  \bK^{-1}\bS^T \bR_{\bu_i} \bS \bK^{-T}
\end{equation}
Since $\bx = \bS \bar{\bx}$, the $2 \times 2$ covariance on the image plane is
\begin{equation}
   \bR_{\bx_i}= \bS \bR_{\bar{\bx}_i} \bS^T = \bS \bK^{-1}\bS^T \bR_{\bu_i} \bS \bK^{-T} \bS^T
\end{equation}
Finally, as was shown in Ref.~\cite{Cheng:2006}, the covariance of the unit vector LOS direction is given by
\begin{equation}
    \bR_{\ba_i} = \left( \frac{\partial \ba_i }{\partial \bx_i} \right) \bR_{ \bx_i} \left( \frac{\partial \ba_i }{\partial \bx_i } \right)^T
\end{equation}
where the Jacobian $\partial \ba_i / \partial \bx_i$ is the $3 \times 2$ matrix
\begin{equation}
    \frac{\partial \ba_i }{\partial \bx_i } = \left(\frac{1}{\| \bar{\bx}_i \|} \bI_{3 \times 3} - \frac{1}{\| \bar{\bx}_i \|^3  } \bar{\bx}_i \bar{\bx}^T_i \right) \bS^T = \frac{-1}{\| \bar{\bx}_i \|}\left[ \ba_i \times \right]^2 \bS^T
\end{equation}
and where $\left[ \, \cdot \, \times \right]$ is  the skew-symmetric cross-product matrix, $\left[\ba \times \right] \bb = \ba \times \bb$. 
It was shown in Ref.~\cite{Cheng:2006} that $\bR_{QMM} \geq \bR_{\ba}$, suggesting that the QMM provides a conservative bound of the measurement uncertainty. However, to use the QMM is to assume measurement statistics that are inconsistent with the optical instrument's actual behavior---ultimately leading to reduced state estimation performance. 

The development of the triangulation methods that follow can be equivalently achieved with any normalization of the LOS direction $\bl_i$ and its corresponding error covariance. The authors prefer to use $\bar{\bx}_i = \bK^{-1} \bar{\bu}_i$ in almost all cases, since this is often easier than working with the unit vector $\ba_i$.

\section{Trigonometric Solutions}
\label{Sec:TrigSolutions}

Simple trigonometric relations may be used to develop two of the classical solutions to the triangulation problem. Referred to here as the \emph{trigonometric solutions}, one solution method follows from the Law of Sines and the other follows from the Law of Cosines. Both of these solutions have found widespread use in computer vision. 
  
In the discussions that follow, the development for both of the trigonometric solutions are first presented from a vector analysis perspective. This is done since the vector interpretation has clear algorithmic advantages. These solutions are then shown to be equivalent to either the Law of Sines or Law of Cosines. The reader is reminded that idea behind the Law of Sines was understood by the classical Greek geometers (Euclid's \emph{Elements}, Prop. I.18 and I.19 \cite{EuclidElements}) and was known in its modern form by at least the 13th century CE (when it appears in the work of Nasir al-Din al-Tusi, ``Treatise on the Quadrilateral''  \cite{Tusi:lawofsines}). Similarly, the idea underpinning the Law of Cosines is also found in Euclid's \emph{Elements} (Prop II.12 and II.13\cite{EuclidElements}) and appeared in its modern form by at least the 15th century CE (notably in the work Jamsh\={\i}d al-K\={a}sh\={\i} \cite{Al-Kashi:lowofcos}). Thus, while the vector notation used here is modern, the fundamental relationships necessary to construct the trigonometric solutions to the triangulation problem would have been quite familiar to classical Greek, medieval Islamic, and Renaissance European geometers.  
  
The trigonometric solutions are geometrically exact and produce a perfect solution when provided perfect (noise free) measurements. They have the advantage of not requiring any \emph{a priori} position knowledge, admitting a closed-form solution, and being easily scalable to many ($n>2$) measurements (with the methods based on Law of Sines being considerably better at scaling than the Law of Cosines). The classical trigonometric solutions both have the disadvantage of being statistically suboptimal when challenged with noisy measurements obtained from a camera, telescope, or other optical sensor. This suboptimality arises because they are conventionally written as an ordinary (unweighted) least squares problems, though the Law of Sines method may be easily modified to overcome this drawback (see Section~\ref{Sec:LOST}). Finally, under consistent normalization assumptions with two LOS measurements ($n=2$), the two trigonometric solutions produce identical results. Thus, of the two methods, the Law of Sines approach is usually preferred since it is simpler and scales better with many LOS measurements.

The discussions that follow are primarily written from the perspective of a resection problem (e.g., spacecraft navigation). If desired, a simple change in notation will transform these results into the intersection problem (e.g., 3-D reconstruction).

\subsection{Law of Sines: The Direct Linear Transform (DLT)}
\label{Sec:DLT}
Optical instruments such as cameras and telescopes measure LOS directions $\bl_i$. However, the distance between the instrument and the observed object along that LOS direction is not directly measured---thus giving rise to the proportional relationship in Eq.~\eqref{eq:BasicPinholeNoFrame}. Rather than attempting to explicitly solve for these unknown distances, it is sometimes easier to remove them entirely by use of the Direct Linear Transform (DLT) \cite{Hartley:2003,AbdelAziz:2015}. In recent years, this approach has been considered for both camera-based navigation with planetary landmarks (a resection problem) \cite{Christian:2019} and for orbit determination from ground-based telescope observations (an intersection problem) \cite{Kaplan:2011}.

\subsubsection{Development by Vector Analysis}
The DLT is the simplest of the triangulation algorithms and its development consists of a few steps. To begin, consider the case where one obtains $n \geq 2$ LOS measurements $\{ \bl_{i} \}^n_{i=1}$ to points with known 3-D locations $\{ \bp_{i} \}^n_{i=1}$. Now, to remove the scale (i.e., range) ambiguity in Eq.~\eqref{eq:BasicPinholeNoFrame}, take the cross-product of both sides with the known LOS direction $\bl_{i}$
\begin{equation}
    \label{eq:DLT1}
    \left[ \bl_{i} \times \right] \bl_{i} \propto \left[ \bl_{i} \times \right] \left(\bp_i - \br \right) = \textbf{0}_{3 \times 1}
\end{equation}
Both Eq.~\eqref{eq:BasicPinholeNoFrame} and Eq.~\eqref{eq:DLT1} are true in any consistent frame, and so no frame-specific notation is required. Thus, after simplification,
\begin{equation}
    \left[ \bl_{i} \times \right] \br = \left[ \bl_{i}  \times \right]\bp_{i}
    \label{eq:DLT1a}
\end{equation}
If many LOS measurements $\{ \bl_i \}_{i=1}^n$ are available, they may be stacked into a linear system as
\begin{equation}
    \begin{bmatrix}
        \left[ \bl_1  \times \right] \\
        \left[ \bl_2  \times \right] \\
        \vdots \\
        \left[ \bl_n  \times \right]
    \end{bmatrix}
    \br = 
    \begin{bmatrix}
        \left[ \bl_1 \times \right]\bp_{1} \\
        \left[ \bl_2 \times \right]\bp_{2} \\
        \vdots \\
        \left[ \bl_n \times \right]\bp_{n}
    \end{bmatrix}
    \label{eq:DLT_LS_NoFrame}
\end{equation}
This linear system may be solved in the least squares sense for the estimated position $\hat{\br}$. The DLT equations [e.g., Eq.~\eqref{eq:DLT1a} or Eq.~\eqref{eq:DLT_LS_NoFrame}] are the same for both resection and intersection forms of the triangulation problem since $\bl_i \propto \bp_i - \br \propto \br - \bp_i$. Though Eq.~\eqref{eq:DLT_LS_NoFrame} is true in any frame, some frame choices have algorithmic advantages in certain situations. 

It is desirable for a triangulation algorithm to produce a consistent solution regardless of the choice of localization frame \cite{Hartley:1997}. This condition is satisfied for the DLT regardless of the frame chosen. To see this, suppose that one knows the locations of the $n$ observed points $\{ \bp_{I_i} \}_{i=1}^n$ as expressed in some localization frame denoted by $I$. Further suppose the triangulation $\tau$ produces the solution $\bar{\br}_I = \tau(\{\bar{\bu}_i\}_{i=1}^n, \{\bar{\bp}_{I_i}\}_{i=1}^n)$, where $\bar{\br}_I^T = [\br^T_I, 1]$ and $\bar{\bp}^T_{I_i} = [\bp^T_{I_i}, 1]$. If $\bM^I_A$ is a $4\times4$ matrix describing a general frame transformation (rotation and translation) between two different localization frames $I$ and $A$, then a consistent triangulation solution would imply $\bM^I_A \bar{\br}_I = \tau(\{\bar{\bu}_i\}_{i=1}^n, \bM^I_A \{\bar{\bp}_{I_i}\}_{i=1}^n)$. Simple substitution will show this condition to be satisfied for Eq.~\eqref{eq:DLT_LS_NoFrame}.

Since almost any reasonable frame convention may be used with Eq.~\eqref{eq:DLT_LS_NoFrame}, a simple convention is usually best. As an example, the specific convention preferred by the authors is now provided. Following the sensor model from Section~\ref{Sec:CamModel}, the LOS measurements originate as pixel coordinates $\bar{\bu}_i$ in a digital image. These image points produce LOS measurements in the camera frame as described by Eq.~\eqref{eq:PixCoordToLOS}. Recognizing that different LOS measurements may come from different cameras, denote the camera frame producing image point $\bar{\bu}_i$ as $C_i$ and the corresponding camera calibration matrix as $\bK_i$. If all the measurements come from one image, then they all come from the same camera and $C=C_i, \, \bK = \bK_i \; \forall i$. Thus, taking the most straightforward approach, one may write
\begin{equation}
    \left[ \bK_i^{-1} \bar{\bu}_i \times \right] \bT^I_{C_i}  \br_I = \left[ \bK_i^{-1} \bar{\bu}_i  \times\right] \bT^I_{C_i}  \bp_{I_i} 
    \label{eq:DLT2}
\end{equation}
where $\bT^I_{C_i}$ is the attitude transformation matrix that transforms a vector expressed in frame $I$ to that same vector expressed in frame $C_i$. The result in Eq.~\eqref{eq:DLT2} makes use of the LOS normalization $\bl_{C_i} \propto \bar{\bx}_i = \bK_i^{-1} \bar{\bu}_i$. Although any LOS normalization may be used, some care is required since (re)normalization acts as a weighting on the affected rows of the linear system from Eq.~\eqref{eq:DLT_LS_NoFrame}. Inconsistent normalization can result in inconsistent weighting of the measurements, which can result in poor triangulation performance. This observation is the motivation for the statistically optimal algorithm presented in Section~\ref{Sec:LOST}.

If many image points $\{ \bar{\bu}_i \}_{i=1}^n$ are available, the corresponding vector equations from Eq.~\eqref{eq:DLT2} may be stacked into a linear system as
\begin{equation}
    \begin{bmatrix}
        \left[ \bK_1^{-1} \bar{\bu}_1 \times \right] \bT^I_{C_1}  \\
        \left[ \bK_2^{-1} \bar{\bu}_2 \times \right] \bT^I_{C_2}  \\
        \vdots \\
        \left[ \bK_n^{-1} \bar{\bu}_n \times \right] \bT^I_{C_n} 
    \end{bmatrix}
    \br_I = 
    \begin{bmatrix}
        \left[ \bK_1^{-1} \bar{\bu}_1 \times \right] \bT^I_{C_1} \bp_{I_1} \\
        \left[ \bK_2^{-1} \bar{\bu}_2 \times \right] \bT^I_{C_2} \bp_{I_2} \\
        \vdots \\
        \left[ \bK_n^{-1} \bar{\bu}_n \times \right] \bT^I_{C_n} \bp_{I_n}
    \end{bmatrix}
    \label{eq:DLT_Hxy}
\end{equation}
Similar to Eq.~\eqref{eq:DLT_LS_NoFrame}, the linear system in Eq.~\eqref{eq:DLT_Hxy} may be solved in the least squares sense (or total least squares sense) for the estimated position $\hat{\br}_I$. An analytic expression for the covariance of this position estimate is given in the Appendix.

\subsubsection{Relationship to Law of Sines}
\begin{figure}[b!]
\centering
\includegraphics[width=0.25\columnwidth,trim=0in 0in 0in 0in,clip]{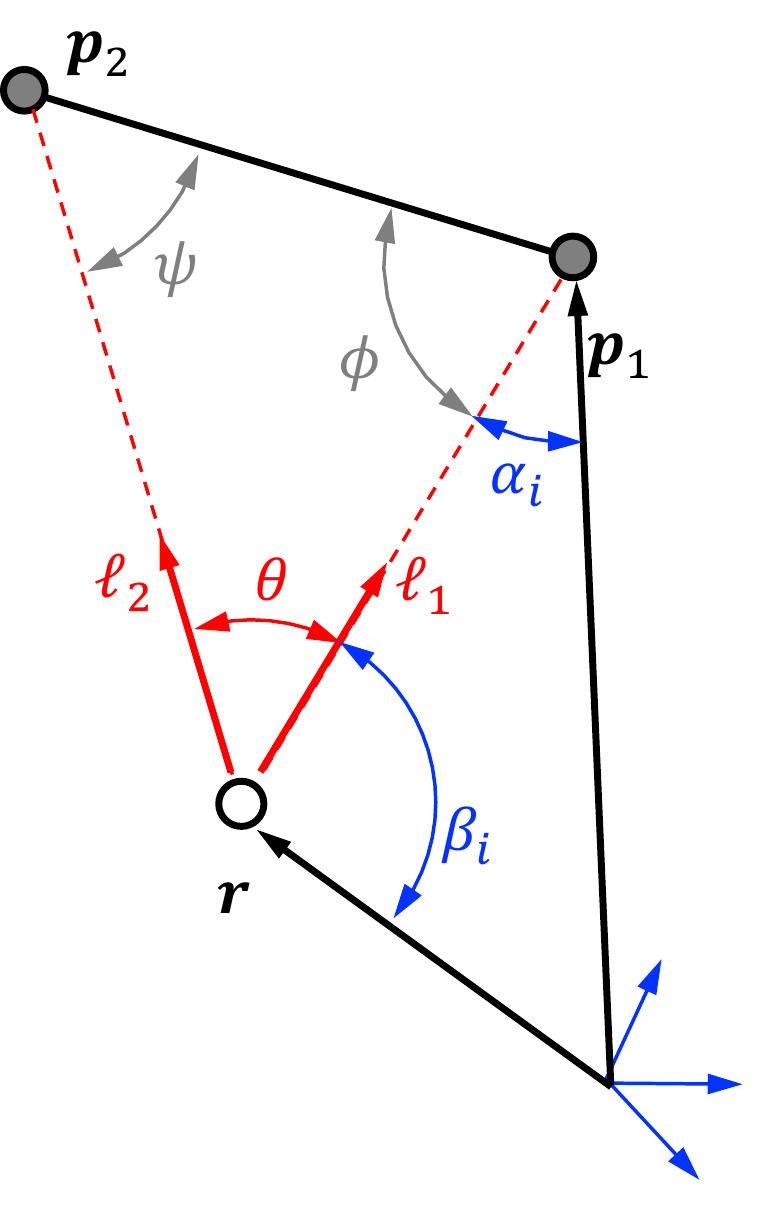}
	\caption{The DLT is a vector representation of the Law of Sines using angles $\alpha_i$ and $\beta_i$.}
	\label{fig:LawOfSinesGeom}
\end{figure}

It was shown in Ref.~\cite{Christian:2019} that the key relation from Eq.~\eqref{eq:DLT1a} is simply the Law of Sines in disguise. Making use of the angles from Fig.~\ref{fig:LawOfSinesGeom}, write the Law of Sines for the triangle formed by the localization frame origin, $\bp_i$, and $\br$,
\begin{equation}
    \frac{\sin \alpha_i}{\| \br \|} = \frac{\sin \beta_i}{ \| \bp_i \| }
\end{equation}
which is the same as
\begin{equation}
    \| \br \| \sin \beta_i = \| \bp_i \| \sin \alpha_i
\end{equation}
This expression is true regardless of the frame in which the vector quantities are expressed. Recalling that $\| \ba \times \bb \| = \|\ba \| \, \| \bb \| \sin \theta$ and accounting for the sign of the cross-product,
\begin{equation}
    \| \br \| \sin \beta_i = \| \br \| \frac{\bl_i \times \br}{ \| \br \| \, \| \bl_i \|} = \| \bp_i \| \frac{\bl_i \times \bp_i}{\| \bl_i \| \, \| \bp_i \| } = \| \bp_i \| \sin \alpha_i
\end{equation}
Simplifying the middle two terms yields
\begin{equation}
    \bl_i \times \br = \bl_i \times \bp_i
\end{equation}
which is exactly the same as Eq.~\eqref{eq:DLT1a}. Hence, the familiar DLT solution is a vector representation of the Law of Sines.

\subsubsection{Relationship to the Collinearity Equations}
\label{Sec:CollinearityEqns}
There are other ways of representing the triangulation problem as a linear system, and some of these other representations are the same as the DLT. For example, consider triangulation with the collinearity equations. 

To arrive at the collinearity equations, rewrite the pinhole camera model from Eq.~\eqref{eq:CameraPinhole} in homogeneous coordinates and remove the proportionality constraint by introducing the scaling $\alpha_i \neq 0$
\begin{equation}
    \alpha_i \bar{\bx}_i = \bT^I_{C_i} ( \bp_{I_i} - \br_I )
\end{equation}
or, equivalently,
\begin{equation}
    \begin{bmatrix} \alpha_i x_i \\ \alpha_i y_i \\ \alpha_i \end{bmatrix} = 
    \begin{bmatrix} -\bT^I_{C_i} & \bT^I_{C_i} \bp_{I_i} \end{bmatrix} 
    \begin{bmatrix} \br_I \\ 1 \end{bmatrix} 
\end{equation}
Define the rows of $\bT^I_{C_i}$ as
\begin{equation}
    \bT^I_{C_i} = \begin{bmatrix} \bt^T_1 \\ \bt^T_2 \\ \bt^T_3 \end{bmatrix}
\end{equation}
and the elements of $\bT^I_{C_i} \bp_{I_i}$ as
\begin{equation}
    \bT^I_{C_i} \bp_{I_i} = \begin{bmatrix} p_1 & p_2 & p_3 \end{bmatrix}^T
\end{equation}
The three rows of the pinhole camera model are now
\begin{subequations}
\begin{align}
    \alpha_i x_i & = -\bt^T_1 \br_I + p_1 \label{eq:Collx}\\
    \alpha_i y_i & = -\bt^T_2 \br_I + p_2 \label{eq:Colly}\\
    \alpha_i  & = -\bt^T_3 \br_I + p_3 \label{eq:Collz}
\end{align}
\end{subequations}
Dividing Eqs.~\eqref{eq:Collx} and \eqref{eq:Colly} by Eq.~\eqref{eq:Collz} gives the classical collinearity equations seen in textbooks and widely-used in a variety of applications (e.g., the so-called CAHVOR model for planetary photogrammetry \cite{Gennery:1992,Di:2004})
\begin{subequations}
\begin{align}
    x_i & = \frac{-\bt^T_1 \br_I + p_1}{-\bt^T_3 \br_I + p_3} = \frac{\bt^T_1 (\bp_{I_i}- \br_I)}{\bt^T_3 (\bp_{I_i}- \br_I)} \\
    y_i & = \frac{-\bt^T_2 \br_I + p_2}{-\bt^T_3 \br_I + p_3} = \frac{\bt^T_2 (\bp_{I_i}- \br_I)}{\bt^T_3 (\bp_{I_i}- \br_I)}
\end{align}
\end{subequations}
Within the context of triangulation, the collinearity equations may be used to construct a linear system for $\br$. To do this, substitute Eq.~\eqref{eq:Collz} into Eqs.~\eqref{eq:Collx} and \eqref{eq:Colly}
\begin{subequations}
\begin{align}
     x_i (-\bt^T_3 \br_I + p_3 ) & = -\bt^T_1 \br_I + p_1 \\
     y_i (-\bt^T_3 \br_I + p_3 ) & = -\bt^T_2 \br_I + p_2 
\end{align}
\end{subequations}
Rearranging this pair of equations and stacking them into matrix form yields (writing the equation for $y_i$ first) a linear system that may be solved for $\br_I$ 
\begin{equation}
    \begin{bmatrix} y_i \bt^T_3  -\bt^T_2   \\ \bt^T_{1}  - x_i \bt^T_3 \end{bmatrix} \br_I =\begin{bmatrix} y_i p_3 - p_2 \\ p_1 - x_i p_3 \end{bmatrix}
\end{equation}
Which is easily verified to be the same as
\begin{equation}
    \bS \left[ \bar{\bx} \times \right] \bT^I_{C_i} \br_I = 
    \begin{bmatrix} y_i \bt^T_3 \br_I -\bt^T_2 \br_I  \\ \bt^T_{1} \br_I - x_i \bt^T_3 \br_I \end{bmatrix} =\begin{bmatrix} y_i p_3 - p_2 \\ p_1 - x_i p_3 \end{bmatrix} = \bS \left[ \bar{\bx} \times \right] \bT^I_{C_i} \bp_{I_i}
\end{equation}
where $\bS$ is from Eq.~\eqref{eq:DefBoldS}. Hence, comparing this result to Eq.~\eqref{eq:DLT2} and recalling that $\bar{\bx}_i = \bK^{-1}_i \bar{\bu}_i$, the collinearty equations for each measured image point are exactly the first two rows of the corresponding part of the DLT. Since the skew-symmetric matrix $\left[ \bar{\bx} \times \right]$ is only rank two, the third row is redundant (linearly dependent on the first two). Thus, solving the DLT is equivalent to solving the collinearity equations (and vice versa).

\subsubsection{Relationship to Pl\"{u}cker Coordinates}
\label{Sec:Plucker}

Given the algebraic projective geometry framework considered here, it is logical to consider triangulation through the use of Pl\"{u}cker coordinates. Introduced in 1865 \cite{Plucker:1865}, the so-called Pl\"{u}cker coordinates are a way of representing a 3-D line in $\PP^3$ using the ratios between six parameters (a point in $\PP^5$). This representation of 3-D lines has become popular for a variety of computer vision applications \cite{Hartley:2003,Bartoli:2005}.

The LOS direction $\bl_i \in \PP^2$ is formed by the line passing through the origin (camera location in the camera frame) and the camera-relative point location $\bp_i - \br$. The corresponding LOP is a 3-D line (not necessarily passing through the origin) passing through two points $\br \in \RR^3$ and $\bp_i \in \RR^3$. Therefore, consider the triangulation problem as expressed in $\PP^3$. In this case, let the spacecraft position be denoted by $\bar{\br}^T = [\br^T, 1]$ and the known point's position be denoted by $\bar{\bp}_i^T = [\br^T_i, 1]$. Describe the LOP as the line in $\PP^3$ formed by the join of the points $\bar{\br} \in \PP^3$ and $\bar{\bp}_i \in \PP^3$, which may be written directly using the Pl\"{u}cker matrix $\bL_i$ \cite{Hartley:2003}
\begin{equation}
    \bL_i \propto \bar{\br} \bar{\bp}_i^T - \bar{\bp}_i \bar{\br}^T
\end{equation}
The present application requires the dual representation $\bL^{\ast}_i$, which may be computed as
\begin{equation}
    \bL^{\ast}_i \propto
    \begin{bmatrix}
        \left[ (\br-\bp_i) \times \right] & (\bp_i \times \br) \\
        (\br \times \bp_i)^T & 0
    \end{bmatrix}
\end{equation}
Since the same line is formed by any two points along the line, one may use the relation from  Eq.~\eqref{eq:BasicPinholeNoFrame} to find
\begin{equation}
    \bL^{\ast}_i \propto
    \begin{bmatrix}
        \left[ \bl_i \times \right] & (\bp_i \times \bl_i) \\
        (\bl_i \times \bp_i)^T & 0
    \end{bmatrix}
\end{equation}
This solution is valid in any frame, and one may easily choose to express this problem in the camera frame, localization frame, or any other frame. 
Now, recall that any point $\bar{\bq}^T \propto [\bq^T,1] \in \PP^3$ lies on the line when $\bL^{\ast}_i \bar{\bq} = \textbf{0}_{4 \times 1}$. Thus, it follows that
\begin{equation}
    \label{eq:PluckerSolution}
    \bL^{\ast}_{i} \bar{\br} = \textbf{0}_{4 \times 1}
\end{equation}
The matrix $\bL^{\ast}_i$ has rank two. Hence when $n \geq 2$ unique LOS measurements are available one may construct the linear system
\begin{equation}
    \label{eq:PluckerSolutionAll}
    \begin{bmatrix}
        \bL^{\ast}_1 \\
        \vdots \\
        \bL^{\ast}_n
    \end{bmatrix} \bar{\br} = \textbf{0}_{4n \times 1}
\end{equation}
The result in Eq.~\eqref{eq:PluckerSolutionAll} is a null space problem and the least squares solution for $\bar{\br}$ may be found directly via the Singular Value Decomposition (SVD). The resulting $\hat{\bar{\br}}\in\PP^3$ is a $4 \times 1$ vector with ambiguous scale (it represents the null space direction) and one may recover the camera location $\hat{\br} \in \RR^3$ by scaling the fourth component to unity.

It is now observed that the Pl\"{u}cker coordinate solution shown here is the same as the DLT solution. To see this, begin by noting that the first three rows of Eq.~\eqref{eq:PluckerSolution} are
\begin{equation}
    \begin{bmatrix}
        \left[ \bl_i \times \right] & (\bp_{i} \times \bl_i) 
    \end{bmatrix}
    \begin{bmatrix}
        \br \\ 1
    \end{bmatrix} = \textbf{0}_{3 \times 1}
\end{equation}
which is the same as
\begin{equation}
        \left[ \bl_i \times \right] \br + (\bp_{i} \times \bl_i) = \textbf{0}_{3 \times 1}
\end{equation}
\begin{equation}
        \left[ \bl_i \times \right] \br  = \left[ \bl_i \times \right] \bp_{i}
\end{equation}
which, in turn, is identical to the foundational DLT expression from Eq.~\eqref{eq:DLT1a}. Moreover, the fourth row in Eq.~\eqref{eq:PluckerSolution} is linearly dependent on the first three rows and simply says that $\br$ must lie in the plane spanned by the origin and the 3D LOS (i.e., the line connecting $\bp_i$ and $\br$). To arrive at this result, observe that the normal to this plane as expressed in the camera frame is given by $\bl_{i} \times \bp_{i}$. Since the camera location $\br$ lies along the LOS, it must lie in this plane. Hence $(\bl_{i} \times \bp_{i})^T \br = 0$, which is the fourth row in Eq.~\eqref{eq:PluckerSolution}.

\subsection{Law of Cosines: Explicit Range Estimation}
\label{Sec:ExplicitRange} 
The missing piece of information for each measurement is the unknown range $\rho_i = \| \bp_i - \br \|$, and so it is understandable that one might wish to explicitly estimate these unknown ranges. Following this line of reasoning, the unknown scale in Eq.~\eqref{eq:BasicPinholeNoFrame} may be made explicit by introducing the range,
\begin{equation}
    \label{eq:DefxiNoFrame}
    \rho_{i} \ba_{i} = \bp_{i} - \br \propto \bl_i
\end{equation}
which is true in any consistent frame. Solving the triangulation problem by finding the unknown ranges $\{\rho_i\}^n_{i=1}$ has been suggested within the context of spacecraft navigation by a number or authors \cite{Karimi:2015,Franzese:2020,Franzese:2022}. 

\subsubsection{Development by Vector Analysis}
As before, consider the case where one obtains $n \geq 2$ LOS measurements $\{ \bl_{i} \}^n_{i=1}$ to points with known 3-D locations $\{ \bp_{i} \}^n_{i=1}$. Choose to normalize the LOS measurements to be the unit vectors $\{ \ba_{i} \}^n_{i=1}$. Thus, by rearrangement of Eq.~\eqref{eq:DefxiNoFrame},
\begin{equation}
    \label{eq:PosAnyFrame}
    \br = \bp_{i} - \rho_i \ba_{i} 
\end{equation}
where the objective is to find $\br$ without any \emph{a priori} knowledge of $\rho_i$. 
Since $\br$ is the same for all measurements, the relation from Eq.~\eqref{eq:PosAnyFrame} may be used to equate any pair of measurements $i \neq j$, 
\begin{equation}
    \bp_{i} - \rho_i \ba_{i} = \br = \bp_{j} - \rho_j \ba_{j}
\end{equation}
which can be rearranged to find
\begin{equation}
     \rho_j \ba_{j} - \rho_i \ba_{i} = \bp_{j}  - \bp_{i} 
\end{equation}
Now, let $\bd_{ij}$ be the vector from $\bp_i$ to $\bp_j$,
\begin{equation}
    \label{eq:Defdij}
    \bd_{ij} = \bp_{j} - \bp_{i}
\end{equation}
Substituting for $\bd_{ij}$ and left-multiplying by either $\ba^T_{i}$ or $\ba^T_{j}$ leads to a pair of equations that are linear in the unknown ranges $\rho_i$ and $\rho_j$
\begin{equation}
        \label{eq:LawOfCosines1}
      (\ba^T_{i} \ba_{j}) \rho_j - \rho_i  = \ba^T_{i} \bd_{ij}
\end{equation}
\begin{equation}
        \label{eq:LawOfCosines2}
     \rho_j - (\ba^T_{j} \ba_{i}) \rho_i  = \ba^T_{j} \bd_{ij}
\end{equation}
It is possible to construct such a pair of equations for every $\{i,j\}$ satisfying $i < j \leq n$, ultimately producing a total of $2\binom{n}{2}$ relationships with $n$ unknown range values. The resulting equations may be stacked to create a linear least squares problem that may be solved for the $n$ unknown ranges. For example, the case of $n=3$ leads to the following linear system
\begin{equation}
      \begin{bmatrix}
      -1 & \ba^T_{1} \ba_{2} & 0 \\
      -\ba^T_{2} \ba_{1} & 1 & 0 \\
      -1 & 0 & \ba^T_{1} \ba_{3} \\
      -\ba^T_{3} \ba_{1} & 0 & 1 \\
      0 & -1 & \ba^T_{2} \ba_{3} \\
      0 & -\ba^T_{3} \ba_{2} & 1 
      \end{bmatrix}
      \begin{bmatrix}
        \rho_1\\ \rho_2 \\ \rho_3
      \end{bmatrix}
      = 
      \begin{bmatrix}
        \ba^T_{1} \bd_{{12}} \\
        \ba^T_{2} \bd_{{12}} \\
        \ba^T_{1} \bd_{{13}} \\
        \ba^T_{3} \bd_{{13}} \\
        \ba^T_{2} \bd_{{23}} \\
        \ba^T_{3} \bd_{{23}} \\
      \end{bmatrix}
      \label{eq:RangeLinSys}
\end{equation}
which may be solved in the least squares sense for the estimated ranges $\{ \hat{\rho}_i\}^n_{i=1}$. Of note is that Eq.~\eqref{eq:RangeLinSys} is true in any consistent frame.

With the ranges known, it is possible to solve for $\br$ directly using Eq.~\eqref{eq:PosAnyFrame} for any one of the LOS measurements. In general, however, it may be more desirable to solve for the position as the average of all the LOS measurements. Since the position is typically desired in the localization frame, 
\begin{equation}
    \hat{\br}_I = \frac{1}{n} \sum_{i=1}^{n}  \left( \bp_{I_i} - \hat{\rho}_i \ba_{I_i} \right)
        \label{eq:ExpRangeFinalSoln}
\end{equation}
An analytic expression for the covariance of this position estimate is given in the Appendix.

\subsubsection{Relationship to Law of Cosines}
It may be shown that Eq.~\eqref{eq:LawOfCosines1} and Eq.~\eqref{eq:LawOfCosines2} [and, hence, the rows of the linear system in Eq.~\eqref{eq:RangeLinSys}] are nothing more than the Law of Cosines in disguise. Therefore, begin by writing the Law of Cosines as for each of the interior angles belonging to the triangle described by edges $\rho_i$, $\rho_j$, and $\|\bd_{ij}\|$:
\begin{equation}
    \label{eq:LoC1}
    \| \bd_{ij} \|^2 = \rho_i^2 + \rho_j^2 - 2\rho_i \rho_j \ba^T_i \ba_j
\end{equation}
\begin{equation}
    \label{eq:LoC2}
    \rho_i^2 = \| \bd_{ij} \|^2 + \rho_j^2 - 2  \rho_j \bd^T_{ij} \ba_j
\end{equation}
\begin{equation}
    \label{eq:LoC3}
    \rho_j^2 = \rho_i^2 + \| \bd_{ij} \|^2 + 2  \rho_i \bd^T_{ij} \ba_i
\end{equation}
Use Eq.~\eqref{eq:LoC1} to substitute for $\| \bd_{ij} \|$ in Eq.~\eqref{eq:LoC2} to find
\begin{equation}
    \rho_i^2 = \left( \rho_i^2 + \rho_j^2 - 2\rho_i \rho_j \ba^T_i \ba_j \right) + \rho_j^2 - 2  \rho_j \bd^T_{ij} \ba_j
\end{equation}
Next, after eliminating $\rho^2_i$,
\begin{equation}
    2 \rho_j^2 - 2\rho_i \rho_j \ba^T_i \ba_j = 2  \rho_j \bd^T_{ij} \ba_j
\end{equation}
This simplifies to
\begin{equation}
    \rho_j - \rho_i (\ba^T_i \ba_j) = \bd^T_{ij} \ba_j
\end{equation}
and is identical to Eq.~\eqref{eq:LawOfCosines2}. Likewise, substitution of Eq.~\eqref{eq:LoC1} into Eq.~\eqref{eq:LoC3} will reproduce Eq.~\eqref{eq:LawOfCosines1}. Hence, the explicit range solution shown here is a vector representation of the Law of Cosines.

\subsubsection{Difficulty with Scalability}
Compared with other triangulation methods, the explicit range method does not scale well with a growing number of LOS measurements. 
In this method, the length of the state vector is dependent on the number of LOS measurements since the unknown range must be estimated for each measurement.  
Moreover, because the explicit range method depends on measurement pairs, the number of equations grows as $2\binom{n}{2}$---which is nearly exponential for $n \gg 2$---when every pairwise LOS combination is considered. For example, the consideration of only 20 LOS measurements would require the solution to a least squares problem with a matrix of size $380 \times 20$. To consider 192 measurements (which is the model point with the maximum number of measurements in the 3-D reconstruction example in Section~\ref{Sec:NotreDame}) leads to a matrix of size $36,672 \times 192$. At best, the explicit range method would require inversion of a $192 \times 192$ matrix in this example---as compared to inversion of only a $3 \times 3$ matrix for either the DLT (Section~\ref{Sec:DLT}) or the statistically optimal LOST algorithm (Section~\ref{Sec:LOST}). This rapid growth in problem size makes the explicit range estimation method comparatively undesirable for some applications. 

\subsection{Comparison of the Trigonometric Solutions}
\label{Sec:TrigSolnComp}
Both of the trigonometric solutions are exact when provided perfect measurements. Interestingly, when provided with noisy measurements, both the DLT method (Section~\ref{Sec:DLT}) and the explicit range method (Section~\ref{Sec:ExplicitRange}) produce an identical estimate of $\hat{\br}_I$ when a consistent LOS normalization is used with two LOS measurements ($n=2$). 
Specifically, this occurs when the LOS measurements in the DLT are scaled to be unit vectors [i.e., choose $\bl_{i} \propto \ba_i$ in Eq.~\eqref{eq:DLT_LS_NoFrame}]. This can be shown analytically for \text{two LOS} measurements, and a proof is provided in the Appendix. The results can also be observed in numerical experiments (see Section~\ref{Sec:ExampleTRN}, especially Fig.~\ref{fig:TRN_canted_unitaryDLT}). As compared to the DLT solution suggested in Eq.~\eqref{eq:DLT_Hxy}, this normalization weights points near the image center more and points near the image edges less (since $\| \bar{\bx}_i \|^{-1}$ is larger near the image center). While point localization performance may indeed be better near the image center and worse near the image edges (e.g., empirical focal plane model from Ref.~\cite{Shuster:1990}), the relative weighting of these observations should be derived from the measurement statistics and not from an arbitrary normalization scheme. Proper accounting of this fact tends to favor the DLT solution over the explicit range solution.

From a practical standpoint, there are a few reasons to prefer the DLT solution over the explicit range solution. The DLT is (1) much simpler, (2) scales more elegantly with many measurements, and (3) allows for easy control of the normalization scheme. The DLT also forms the foundation for statistically optimal triangulation (i.e., LOST algorithm in Section~\ref{Sec:LOST}) and dynamic triangulation (Section~\ref{Sec:DynamicTri}). For these reasons, the DLT method (Law of Sines method) is almost always preferable to the explicit range method (Law of Cosines method).

\section{Optimal Triangulation}
\label{Sec:OptimalTri}
The trigonometric solutions from Section~\ref{Sec:TrigSolutions} do not usually provide a statistically optimal estimate of the unknown point's location. Though both of these solutions do solve a least squares optimization problem, they do not minimize the correct cost function. Proper formulation of the cost function requires consideration how LOS measurement errors are related to state estimation errors and a realization that some measurements provide better information than others. One appropriate way to proceed is with a maximum likelihood estimate (MLE) framework \cite{Fisher:1922,Tapley:2004}. The two-measurement case admits some interesting optimal solutions, though the classical approach requires finding the roots of a polynomial of degree six \cite{Hartley:1997}. After first reviewing this special case, a more general approach---referred to here as the Linear Optimal Sine Triangulation (LOST) algorithm---is shown to provide an equivalent estimate as the solution to a linear system. The new LOST algorithm is generally preferable to the classical polynomial solutions in that (1) it provides an equivalent solution without the need to solve a degree six polynomial and (2) it provides the MLE for an arbitrary number ($n\geq2$) of LOS measurements as the solution to a linear system (with no iteration).

\subsection{Two LOS Measurements: The Polynomial Methods}
At least two absolute LOS measurements are necessary to triangulate the location of a camera and some interesting algorithms arise when exactly two LOS measurements exist. There are two different scenarios to consider here. The first scenario is when the two LOS measurements are collected from different cameras and extracted from different images, and so the two measured points generally lie in two different image planes. The second scenario is when both of the LOS measurements are collected by the same camera and extracted from the same image, and so the two measured points lie in the same image plane. These two scenarios lead to two different (but equivalent) optimal algorithms. The solution for two images involves solving a polynomial of degree six, while the solution for a single image involves solving only a quadratic equation. A small numerical example is provided in the supplemental material for this article.

Consider a collection of LOS measurements to $n$ objects that were produced by optical instruments.
Applying the measurement model from Section~\ref{eq:MeasModels},
\begin{equation}
    \label{eq:FPMeasModel}
   \bu_i =  \bS \bar{\bu}_i = h_i( \br) =
    \frac{\bS \bK \bT^I_{C_i} (\bp_{I_i} - \br_I) }{ \bk^T \bT^I_{C_i}  (\bp_{I_i} - \br_I) }
\end{equation}
and a noisy measurement of
\begin{equation}
   \tilde{\bu}_i = \bu_i + \bnu_i = h_{i}(\br) + \bnu_i
\end{equation}
As before, assuming $\bnu_i \sim \mathcal{N}(\textbf{0}_{2 \times 1},\bR_{\bu_i})$ and that the additive noise on measurements $i$ and $j$ are uncorrelated (i.e., $E[\bnu_i^T \bnu_j] = \textbf{0}_{2 \times 2}$), the MLE statement is
\begin{equation}
   \max \; J(\br) = C \exp \left\{ - \sum_{i=1}^n \left[\tilde{\bu}_i - h_i(\br) \right]^T \bR^{-1}_{\bu_i} \left[\tilde{\bu}_i - h_i(\br) \right] \right\}
\end{equation}
which is the same as the minimization problem
\begin{equation}
   \min \; J(\br) = \sum_{i=1}^n \left[\tilde{\bu}_i - h_i(\br) \right]^T \bR^{-1}_{\bu_i} \left[\tilde{\bu}_i - h_i(\br) \right]
\end{equation}
Thus, the MLE solution seeks to find the location $\br$ that minimizes the reprojection errors as weighted by the measurement covariance. 
In many cases, the image point errors are isotropic and the measurement covariance is $\bR_{\bu_i} \approx \sigma^2_{\bu_i} \bI_{2 \times 2}$. In this case,
\begin{equation}
   \min \; J(\br) = \sum_{i=1}^n \frac{1}{\sigma_{\bu_i}^2} \left[\tilde{\bu}_i - h_i(\br) \right]^T \left[\tilde{\bu}_i - h_i(\br) \right]
\end{equation}
Moreover, when there are only two LOS measurements ($n=2$) and assuming similar uncertainty in each measurement (i.e., $\sigma_{\bu_1} = \sigma_{\bu_2}$), the cost function simplifies further to
\begin{equation}
   \min_{\br \in \RR^3} \; J(\br) =  \|\tilde{\bu}_1 - h_1(\br) \|^2 + \|\tilde{\bu}_2 - h_2(\br) \|^2
\end{equation}
which is simply the sum of the squared reprojection errors. Since the two reprojected points $\bu_1 = h_1(\br)$ and $\bu_2 = h_2(\br)$ must satisfy the epipolar constraint, this is equivalent to
\begin{align}
   \min \; J(u_1,v_1,u_2,v_2 ) = & \left[ (\tilde{u}_1 - u_1)^2 + (\tilde{v}_1 - v_1)^2 \right] + \left[ (\tilde{u}_2 - u_2)^2 + (\tilde{v}_2 - v_2)^2 \right] \\
   & \text{s.t. } \bar{\bu}^T_2 \bF \bar{\bu}_1 = 0 \nonumber
\end{align}
where $\bF$ is the fundamental matrix. This is the cost function suggested in the majority of the contemporary literature when seeking an ``optimal'' triangulation with two LOS measurements \cite{Hartley:1997}.

Further, for a calibrated camera with a known calibration matrix $\bK$, the cost function may be rewritten in terms of the image plane coordinates. Assuming square pixels ($d_x = d_y$) and no detector skewness ($\alpha = 0$) the matrix $\bK$ simplifies and
\begin{align}
\label{eq:cost_function}
   \min \; J(x_1,y_1,x_2,y_2 )  & = w_1 \left[ (\tilde{x}_1 - x_1)^2 + (\tilde{y}_1 - y_1)^2 \right] + w_2 \left[ (\tilde{x}_2 - x_2)^2 + (\tilde{y}_2 - y_2)^2 \right] \\
   & \text{s.t. } \bar{\bx}^T_2 \bE \bar{\bx}_1 = 0 \nonumber
\end{align}
where $w_i =1 /\sigma^2_{\bx_i} = d^2_{x_i}/\sigma^2_{\bu_i}$ and where $\bE$ is the essential matrix. Note that this form has reintroduced the possibility of a different uncertainty for the two different LOS measurements. This cost function can be used to produce optimal triangulation algorithms for special case of two LOS measurements.

\subsubsection{Measurements from Different Images}
\label{Sec:TriTwoPointsHS}
Consider the situation where each LOS measurement comes from a different image. This is a necessary condition for the intersection form of the triangulation problem. The situation is less common for resection, but does still occur for a few important problems (e.g., triangulation with two objects whose angular separation is larger than a single optical instrument's field of view). The optimal solution for the intersection problem was developed by Hartley and Sturm in 1997 \cite{Hartley:1997} and involves finding the roots of a polynomial of degree six. It is possible to show that the resection problem is identical to the intersection problem, and that the optimal solution from Ref.~\cite{Hartley:1997} may be used for both forms.

\begin{figure}[b!]
\centering
\includegraphics[width=0.45\columnwidth,trim=0in 0in 0in 0in,clip]{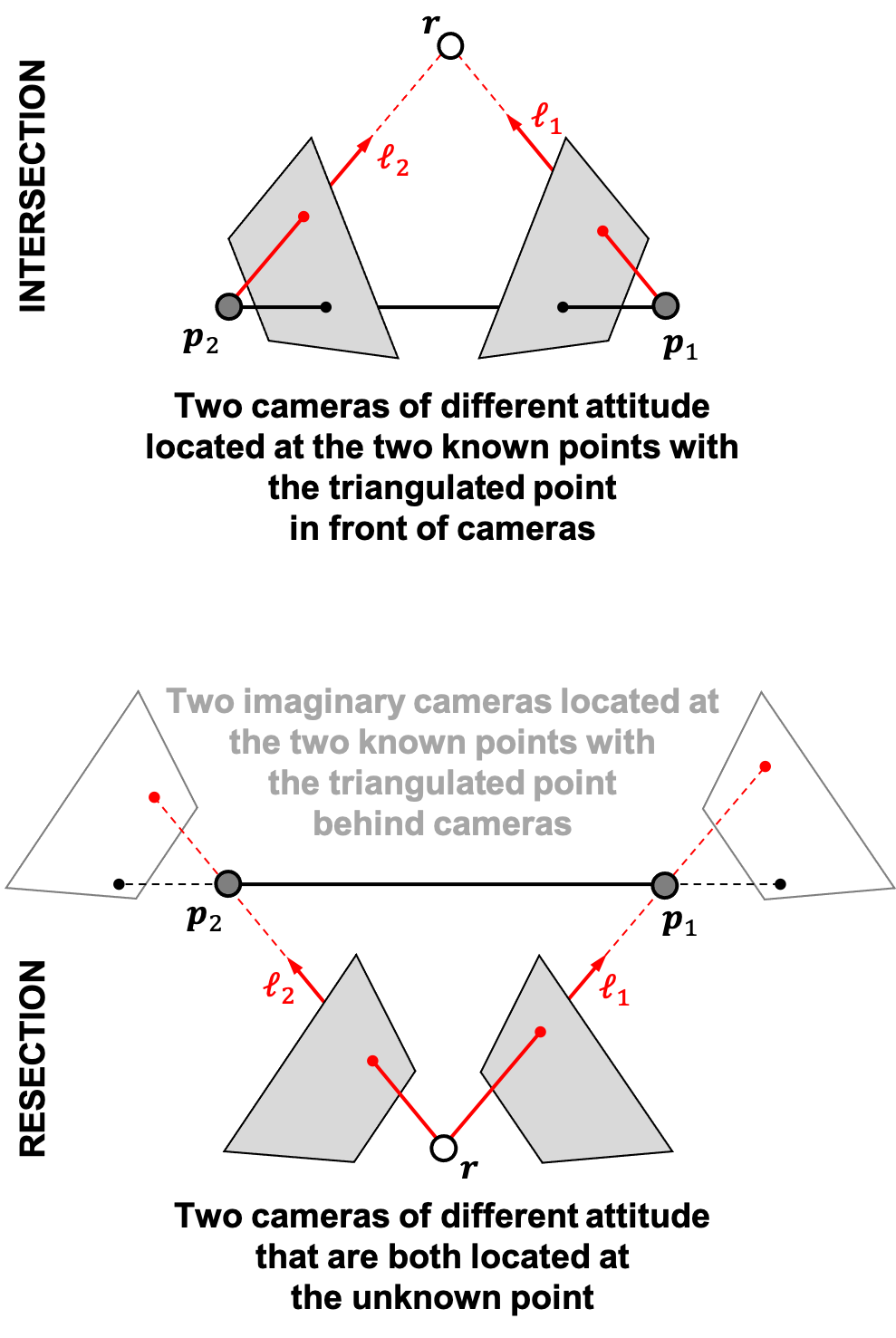}
	\caption{Illustration of the equivalence between intersection and resection geometry for two LOS measurements.}
	\label{fig:OptimalTwoImageGeom}
\end{figure}

The first task is to establish the equivalence between the intersection and resection form of the two-image triangulation problem. This may be accomplished by considering the geometry illustrated in Fig.~\ref{fig:OptimalTwoImageGeom}. The geometry of the intersection form is shown in the top frame of Fig.~\ref{fig:OptimalTwoImageGeom} and it is this configuration that is considered in Ref.~\cite{Hartley:1997}. The resection problem may seem very different at first, but the bottom frame of Fig.~\ref{fig:OptimalTwoImageGeom} illustrates how resection with two images is equivalent to intersection with two images. Specifically, consider a camera located at the unknown point $\br$ viewing the known point $\bp_i$. If one creates an imaginary camera of the same attitude but located at $\bp_i$ the LOS direction $\bl_i$ intersects the new image plane at exactly the same place---but the point $\br$ is \emph{behind} the imaginary camera. Since the triangulation algorithms do not depend on the observed point's chirality \cite{Hartley:1998}, it does not matter that the triangulated point lies behind these imaginary camera locations. Moreover, when using the two imaginary cameras in the bottom frame of Fig.~\ref{fig:OptimalTwoImageGeom}, the triangulation of the unknown point $\br$ is now an intersection (and not a resection) problem---thus the solution from Ref.~\cite{Hartley:1997} may be used directly.

The solution procedure will now be briefly summarized with only limited derivation (the reader interested in the details is directed to Ref.~\cite{Hartley:1997}) using a notation consistent with the present work. Begin by computing the image plane coordinates of the two epipoles (black dots on the image planes in Fig.~\ref{fig:OptimalTwoImageGeom}),
\begin{equation}
    \be_1 \propto \bT^I_{C_1} (\bp_{I_2} - \bp_{I_1})
    \quad \quad \text{and} \quad \quad
    \be_2 \propto \bT^I_{C_2} (\bp_{I_1} - \bp_{I_2})
\end{equation}
Given the epipoles $\be_1$ and $\be_2$, compute the rotations $\alpha_1$ and $\alpha_2$ as
\begin{equation}
    \begin{bmatrix} \sin \alpha_i \\ \cos \alpha_i \\ \cdot \end{bmatrix}
    \propto \be_i \times \tilde{\bar{\bx}}_i 
\end{equation}
where the left-hand side scale may be found by recalling $\sin^2 \alpha_i + \cos^2 \alpha_i = 1$. From here, one may directly compute the mapping $\bM_i$
\begin{equation}
    \bM_i = \begin{bmatrix} 
                \cos \alpha_i & -\sin \alpha_i &  y_i \sin \alpha_i - x_i \cos \alpha_i \\
                \sin \alpha_i &  \cos \alpha_i & -y_i \cos \alpha_i - x_i \sin \alpha_i \\
                0 & 0 & 1
            \end{bmatrix}
\end{equation}
that results in
\begin{equation}
    \begin{bmatrix} 1 \\ 0 \\ f_i \end{bmatrix} \propto \be'_i \propto \bM_i \be_i
\end{equation}
Consider now the epipolar constraint written in terms of the essential matrix
\begin{equation}
    \bar{\bx}_2^T \bE \bar{\bx}_1 = 0
\end{equation}
where $\bE$ is
\begin{equation}
    \bE \propto \bT^I_{C_2} \bT^{C_1}_I \left[ \be_1 \times \right]
\end{equation}
Note that the original work of Hartley and Sturm writes this in terms of the fundamental matrix, but this difference is not important here. Using the mappings $\bM_1$ and $\bM_2$, one computes the simplified form of $\bE'$
\begin{equation}
    \bE' \propto \bM_2^{-T} \bE \bM_1^{-1} \propto
    \begin{bmatrix}
        f_1 f_2 d & -f_2 c & -f_2 d \\
        -f_1 b & a & b \\
        -f_1 d & c & d
    \end{bmatrix}
\end{equation}
The elements $a,b,c,d,f_1,f_2$ from $\bE'$ may be used to form epipolar lines and to directly write the constrained image plane reprojection error from Eq.~\eqref{eq:cost_function} in terms of the single parameter $t$. Making the appropriate substitutions (see Ref.~\cite{Hartley:1997}) the cost function becomes
\begin{equation}
    \label{eq:CostFcnHS}
    \min_{t\in \RR} \; J(t) = w_1 \frac{t^2}{1+(tf_1)^2} + w_2
    \frac{(ct+d)^2}{(at+b)^2+f_2^2(ct+d)^2}
\end{equation}
where the solution presented here departs slightly from Hartley and Sturm by the introduction of different weightings of the two measurements (the solution from Ref.~\cite{Hartley:1997} weights the measurements equally). Applying the first differential condition $\partial J / \partial t = 0$ yields a polynomial of degree six in $t$,
\begin{equation}
    w_1 \frac{2t}{[1+(tf_1)^2]^2} - w_2\frac{2(ad-bc)(at+b)(ct+d)}{[(at+b)^2+f^2_2(ct+d)^2]^2} = 0
\end{equation}
which is also
\begin{equation}
    w_1 t [(at+b)^2+f^2_2(ct+d)^2]^2 - w_2 [1+(tf_1)^2]^2 (ad-bc)(at+b)(ct+d) = 0
\end{equation}
or, equivalently,
\begin{equation}
    \label{eq:HSPolySum}
    \sum_{k=0}^6 g_k t^k = 0
\end{equation}
where this polynomial's seven coefficients $\{ g_k \}_{k=0}^6$ may be readily computed. 
Solve for the six roots of Eq.~\eqref{eq:HSPolySum} and evaluate the real roots (there are often more than one) using Eq.~\eqref{eq:CostFcnHS} to find which one minimizes the cost function. Checking the roots is necessary since the degree six polynomial may have multiple local minima. Denote the optimal root as $\hat{t}$.

With $\hat{t}$ known, the optimal epipolar lines are
\begin{equation}
    \hat{\bm}'_1 \propto 
    \begin{bmatrix}
        \hat{t}f \\ 1 \\ -\hat{t}
    \end{bmatrix}
    \quad \quad \quad
    \hat{\bm}'_2 \propto 
    \begin{bmatrix}
        -f_2 (c\hat{t} + d) \\ a\hat{t} + b \\ c\hat{t} + d
    \end{bmatrix}
\end{equation}
Since the points were shifted to the origin by $\bM_i$, the goal is to find the point on these lines closest to the origin. It follows that
\begin{equation}
    \begin{bmatrix}
        \hat{x}'_1 \\ \hat{y}'_1 \\ 1
    \end{bmatrix} \propto
    \begin{bmatrix}
        -(\hat{t}f) (-\hat{t}) \\ -1(-\hat{t}) \\ (\hat{t}f)^2 + 1^2
    \end{bmatrix} = 
    \begin{bmatrix}
        f \hat{t}^2  \\ \hat{t} \\ f^2 \hat{t}^2 + 1
    \end{bmatrix}
\end{equation}
\begin{equation}
    \begin{bmatrix}
        \hat{x}'_2 \\ \hat{y}'_2 \\ 1
    \end{bmatrix} \propto
    \begin{bmatrix}
        -(-f_2 (c\hat{t} + d)) (c\hat{t} + d) \\ -(a\hat{t} + b)(c\hat{t} + d) \\ (-f_2 (c\hat{t} + d))^2 + (a\hat{t} + b)^2
    \end{bmatrix} = 
    \begin{bmatrix}
        f_2 (c\hat{t} + d)^2 \\ -(a\hat{t} + b)(c\hat{t} + d) \\ f^2_2 (c\hat{t} + d)^2 + (a\hat{t} + b)^2
    \end{bmatrix}
\end{equation}
which may each be mapped back to the final point according to
\begin{equation}
    \hat{\bar{\bx}}_i = 
    \begin{bmatrix}
        \hat{x}_i \\ \hat{y}_i \\ 1
    \end{bmatrix} \propto 
    \bM^{-1}_i
    \begin{bmatrix}
        \hat{x}'_i \\ \hat{y}'_i \\ 1
    \end{bmatrix} 
\end{equation}
Once the estimated image points $\{ \hat{\bar{\bx}}_i \}_{i=1}^2$ are known, the location of $\hat{\br}$ may be found with any of the trigonometric solutions from Section~\ref{Sec:TrigSolutions}. Since $\{ \hat{\bar{\bx}}_i \}_{i=1}^2$ are constrained to exactly satisfy the epipolar constraint the resulting triangulation problem is now exact.

\subsubsection{Measurements from the Same Image}
\label{Sec:QUAT}

A simpler solution is possible in the case when measurements come from the same camera (for resection) or from two cameras of the same attitude (for intersection). Under these conditions the optimal two-LOS triangulation problem may be solved by the solution to a quadratic equation (polynomial of degree two), instead of the usual polynomial of degree six from Section~\ref{Sec:TriTwoPointsHS}. This reduction to a quadratic equation has been known since at least the mid-1990s, when it was discussed in early work by both Armstrong \cite{Armstrong:1996} and Sturm \cite{Sturm:1997}. What follows is a different---but ultimately equivalent---derivation of this fact. 

To simplify the discussion, and without loss of generality, focus on the resection form of the triangulation problem. To find the triangulation constraint, first recognize that $\bl_{C}$, $\bl_{2}$, and $\bd_{{12}}$ are linearly dependent and all lie in a common plane. Thus, since the vector $\bd_{{12}} \times \bl_{2}$ is perpendicular to the this plane, it follows that
\begin{equation}
   \bl^T_{1} \left[ \bd_{12} \times \right] \bl_{2} = 0
\end{equation}
Recalling that in the camera frame $\bl_{C_i} \propto \bar{\bx}_i$, one may also write
\begin{equation}
   \bar{\bx}_1^T \left[ \bd_{C_{12}} \times \right] \bar{\bx}_2 = 0
\end{equation}
which is the same thing as the epipolar constraint. Recalling $\bar{\bx}^T_i = [x_i,y_i,1]$ and defining $\bd^T = [ d, e, f]$, the constraint develops as
\begin{equation}
\label{eq:constraint_1C}
\bar{\bx}_1^T \left[ \bd_{C_{12}} \times \right] \bar{\bx}_2 = x_1 \left( e - f y_2\right) + y_1  \left( f x_2 - d \right) + \left( d y_2 - e x_2 \right) = 0
\end{equation}
Thus, the triangulation (epipolar) constraint may be adjoined to the cost function from Eq.~\eqref{eq:cost_function} with a Lagrange multiplier, $\lambda$
\begin{equation}
   \min\; J(x_1,y_1,x_2,y_2,\lambda) = \sum_{i=1}^2 w_i \left[ (\tilde{x}_i - x_i)^2 + (\tilde{y}_i - y_i)^2 \right] + \lambda \left( \bar{\bx}_1^T \left[ \bd_{C_{12}} \times \right] \bar{\bx}_2 \right)
\end{equation}
Applying the first differential condition yields
\begin{subequations}
\begin{align}
\label{eq:parx1}
    \frac{\partial J}{\partial x_1} = 2 w_1 \left(x_1 - \tilde{x}_1 \right) - \lambda \left( f y_2 - e\right) = 0\\
\label{eq:pary1}
    \frac{\partial J}{\partial y_1} = 2 w_1 \left( y_1-\tilde{y}_1 \right) + \lambda \left( f x_2 - d \right) = 0\\
\label{eq:parx2}
    \frac{\partial J}{\partial x_2} = 2 w_2 \left( x_2 - \tilde{x}_2 \right) + \lambda \left( f y_1 - e \right) = 0\\
\label{eq:pary2}
    \frac{\partial J}{\partial y_2} = 2 w_2 \left( y_2 - \tilde{y}_2 \right) - \lambda \left( f x_1 - d \right) = 0
\end{align}
\end{subequations}
Additionally, ${\partial J}/{\partial \lambda} = 0$ gives the constraint at Eq.~\eqref{eq:constraint_1C} as a fifth equation to solve. 

Solving Eq.~\eqref{eq:parx1} to \eqref{eq:pary2} yields an expression of each image plane coordinate as a function of $\lambda$. Given the symmetry between each coordinate, all of their solutions take the form of a rational function where both the numerator and denominator are polynomials of degree two:
\begin{equation}
\label{eq:QUAT_coordinate_fraction}
    ( \cdot ) = \frac{N_{(\cdot)}}{D} = \frac{\gamma_{2}\lambda^2 + \gamma_{1} \lambda + \gamma_{0}}{\gamma_{2}\lambda^2 + \gamma_{1}\lambda + \gamma_{0}}
\end{equation}
with coefficients from Table~\ref{tab:coefficients_single}. Note that the denominator is the same for each coordinate. For example, the fraction for the optimal value of $x_1$ takes the form
\begin{equation}
\label{eq:QUATx1}
    \hat{x}_1 = \frac{N_{x_1}}{D} = \frac{[df]\lambda^2 + [2 w_2 (e - f \tilde{y_2})] \lambda + [-4 w_1 w_2 \tilde{x}_1]}{[f^2]\lambda^2 + [0]\lambda + [-4 w_1 w_2]}
\end{equation}
\begin{table}[bt!]
\caption{Coefficients of the polynomial expressions of the state variable with respect to $\lambda$.}
\centering
\begin{tabular}{lccccc}
\hline
\hline
      & $D$          & $N_{x_1}$                   & $N_{y_1}$                   & $N_{x_2}$                   & $N_{y_2}$                                     \\ \hline
$\gamma_2$ & $f^2$        & $d f$                       & $e f$                       & $d f$                       & $e f$                                         \\ 
$\gamma_1$ & $0$          & $2 w_2 (e - f \tilde{y_2})$ & $2 w_2 (f \tilde{x}_2 - d)$ & $2 w_1 (f \tilde{y}_1 - e)$ & $2 w_1 (d - f \tilde{x}_1)$                   \\ 
$\gamma_0$ & $-4 w_1 w_2$ & $-4 w_1 w_2 \tilde{x}_1$    & $-4 w_1 w_2 \tilde{y}_1$    & $-4 w_1 w_2 \tilde{x}_2$    & $-4 w_1 w_2 \tilde{y}_2$ \\ 
\hline
\hline
\end{tabular}
\label{tab:coefficients_single}
\end{table}

The constraint from Eq. \eqref{eq:constraint_1C} may now be rewritten in terms of $\lambda$ only as
\begin{equation}
    f \frac{N_{y_1} N_{x_2} - N_{x_1} N_{y_2}}{D^2} + \frac{e N_{x_1} - d N_{y_1} + d N_{y_2} - e N_{x_2}}{D} = 0
\end{equation}
or equivalently
\begin{equation}
    \label{eq:Lam4poly}
    f N_{y_1} N_{x_2} - N_{x_1} N_{y_2} + e N_{x_1}D - d N_{y_1}D + d N_{y_2}D - e N_{x_2}D = 0
\end{equation}
The terms within the coefficients of $\lambda^4$ and $\lambda^3$ cancel out, such that Eq.~\eqref{eq:Lam4poly} simplifies into a polynomial of degree two
\begin{equation}
\label{eq:second_order_polyn}
h_2 \lambda^2 + h_1 \lambda + h_0 = 0
\end{equation}
with coefficients
\begin{equation}
    h_2 = f^2 w_1 w_2 \left[f  (\tilde{x}_2 \tilde{y}_1 - \tilde{x}_1 \tilde{y}_2) + e (\tilde{x}_1 - \tilde{x}_2) + d (\tilde{y}_2 - \tilde{y}_1)  \right]
\end{equation}
\begin{equation}
     h_1 = - 2 w_1 w_2 \left[ f^2 (w_1 \tilde{x}_1^2 + w_1 \tilde{y}_1^2 + w_2 \tilde{x}_2^2 + w_2 \tilde{y}_2^2) -2 f (d w_1 \tilde{x}_1 + d w_2 \tilde{x}_2 + e w_1 \tilde{y}_1 + e w_2 \tilde{y}_2)
     + (d^2+e^2) (w_1+w_2) \right]
\end{equation}
\begin{equation}
    h_0 = 4 w_1 w_2 \frac{h_2}{f^2}
\end{equation}
The solution in the general case is given by
\begin{equation}
    \label{eq:quadroot_general}
    \lambda = \frac{-h_1 \pm \sqrt{h_1^2 - 4 h_2 h_0}}{2 h_2}
\end{equation}
In practice, the algorithm proposed first solves the quadratic equation to find the two roots. It then computes the image plane coordinates for each root with Eq.~\eqref{eq:QUAT_coordinate_fraction}. The set of coordinates that minimizes Eq.~\eqref{eq:cost_function} is selected as the solution. 

It is important to consider the case where $f=0$, because $h_2=0$ yields an undefined solution if it is solved with Eq.~\eqref{eq:quadroot_general}. In that case, the solution is much simpler to compute from the Lagrangian, or the unique root can be computed from the existing coefficients as
\begin{equation}
    \label{eq:LambdaLimit}
\begin{aligned}
    \lambda_{f\rightarrow 0} &= \lim_{f \rightarrow 0} \left(-\frac{h_0}{h_1} \right)\\
            &= \frac{2 w_1 w_2 \left[ e (\tilde{x}_1 - \tilde{x}_2) + d (\tilde{y}_2 - \tilde{y}_1)  \right]}{(d^2+e^2) (w_1+w_2)}
\end{aligned}
\end{equation}
For a numerical implementation, it is important to switch to the solution from Eq.~\eqref{eq:LambdaLimit} whenever $h_2$ is very small compared to the other coefficients.

\subsection{More than Two LOS Measurements: The Linear Optimal Sine Triangulation (LOST) Method}
\label{Sec:LOST}
Solving the MLE triangulation problem by explicit minimization of the reprojection errors does not easily scale to more than two LOS measurements. This was attempted in Ref.~\cite{Stewenius:2005} for three LOS measurements ($n=3$) and was found to require the solution to a system of polynomials possessing 47 roots. To attempt this for $n \geq 4$ is completely unreasonable. In light of these challenges, it is commonly argued that to obtain the MLE solution for $n$ views requires one to iteratively solve a non-linear least squares problem (so-called \emph{bundle adjustment}). This, however, is not true. It is possible to create a MLE solution to $\br$ as the solution to a linear system by a double applications of the Law of Sines. The resulting method is referred to here as the Linear Optimal Sine Triangulation (LOST) algorithm.

Begin with the first application of the Law of Sines as written in the form of the DLT. Rewrite Eq.~\eqref{eq:DLT2} as
\begin{equation}
    \left[ \bar{\bx}_i \times \right] \bT^I_{C_i} (\br_I - \bp_{I_i} ) = \textbf{0}_{3 \times 1}
\end{equation}
This condition is only satisfied with perfect measurements. When only the noisy measurements $\tilde{\bar{\bx}}_i = \bK^{-1} \tilde{\bar{\bu}}$ are available, the right-hand side is no longer exactly zero. Instead, one finds that
\begin{equation}
    \label{eq:NoisyDLTforLOST}
    \left[ \tilde{\bar{\bx}}_i \times \right] \bT^I_{C_i} (\br_I - \bp_{I_i} ) = \beps_i
\end{equation}
Recalling from Eq.~\eqref{eq:xbarnoisy} that $\tilde{\bar{\bx}}_i = \bar{\bx}_i + \bw_i$, the residual vector $\beps_i$ is found to be
\begin{equation}
    \beps_i = \left[  \bT^I_{C_i} (\bp_{I_i}  - \br_I ) \times \right] \bw_i 
\end{equation}
with covariance 
\begin{equation}
    \label{eq:RepsTpr}
    \bR_{\beps_i} = E[\beps_i \beps_i^T] = - \left[  \bT^I_{C_i} (\bp_{I_i}  - \br_I ) \times \right] \bR_{\bar{\bx}_i} \left[  \bT^I_{C_i} (\bp_{I_i}  - \br_I ) \times \right]
\end{equation}
where $\bR_{\bar{\bx}_i} = E[\bw_i \bw_i^T]$ is from Eq.~\eqref{eq:DefRxbxb}. The leading negative sign in Eq.~\eqref{eq:RepsTpr} is a result of the identity $[ \, \cdot \, \times]^T = - [ \, \cdot \, \times]$.
It is possible to rewrite $\bR_{\bar{\bx}_i}$ explicitly in terms of the corresponding measurement $\bar{\bu}_i$ and unknown range $\rho_i$ by substitution from Eq.~\eqref{eq:PixCoordToLOS} and Eq.~\eqref{eq:DefxiNoFrame}
\begin{equation}
    \label{eq:RepsKu}
    \bR_{\beps_i} = - \frac{\rho^2_i}{\| \bK^{-1}_i \bar{\bu}_i \|^2} \left[  \bK^{-1}_i \bar{\bu}_i \times \right] \bR_{\bar{\bx}_i} \left[  \bK^{-1}_i \bar{\bu}_i \times \right]
\end{equation}
Therefore, one may write the MLE problem as
\begin{equation}
    \min J(\br_I) = \sum_{i=1}^n \beps^T_i \bR^{-1}_{\beps_i} \beps_i
\end{equation}
which, after substituting for $\beps_i$ using Eq.~\eqref{eq:NoisyDLTforLOST}, becomes
\begin{equation}
    \label{eq:LOSTCostFcn1}
    \min J(\br_I) =  \sum_{i=1}^n -(\br_I - \bp_{I_i} )^T \bT_I^{C_i}  \left[ \tilde{\bar{\bx}}_i \times \right]  \bR^{-1}_{\beps_i} \left[ \tilde{\bar{\bx}}_i \times \right] \bT^I_{C_i} (\br_I - \bp_{I_i} )
\end{equation}
There are three apparent issues here---all of which relate to the practical computation of $\bR^{-1}_{\beps_i}$. First is that $\bR_{\beps_i}$ depends on the ideal pixel coordinates $\bar{\bu}_i$ when only the measured pixel coordinates $\tilde{\bar{\bu}}_i$ are available in practice. Second is that the $3 \times 3$ matrix $\bR_{\beps_i}$ is only rank two and, therefore, not invertable. Third is that $\bR_{\beps_i}$ depends on the unknown distance $\rho_i$ between points $\br$ and $\bp_i$. All three of these issues may be avoided.

Begin with addressing the use of the noisy measurements $\tilde{\bar{\bu}}_i$ in place of the perfect $\bar{\bu}_i$ in Eq.~\eqref{eq:RepsKu}. This is a common problem in navigation. Indeed, a measurement's error covariance is dependent on the measurement itself for all but the simplest sensor models. When approximating the error covariance of $\beps_i$ with the measurement $\tilde{\bar{\bu}}_i$, denote the noisy covariance as $\tilde{\bR}_{\beps_i}$, where
\begin{equation}
    \label{eq:RepsKuNoisy}
    \tilde{\bR}_{\beps_i} = - \frac{\tilde{\rho}^2_i}{\| \bK^{-1}_i \tilde{\bar{\bu}}_i \|^2} \left[  \bK^{-1}_i \tilde{\bar{\bu}}_i \times \right] \bR_{\bar{\bx}_i} \left[  \bK^{-1}_i \tilde{\bar{\bu}}_i \times \right]
\end{equation}
Using the measurement in place of the truth when computing the covariance is correct to first order and has been studied before for image-based LOS measurements \cite{Cheng:2006,Shuster:1990}. Moreover, using measurements or estimates is the only choice since the true values will never be known. This approximation is acceptable in practice since the covariance $\bR_{\beps_i}$ is only used to weight the different residual vectors $\beps_i$. The numerical results in Section~\ref{Sec:Applications} confirm this assertion. Thus, proceed using $\tilde{\bR}_{\beps_i}$ in place of $\bR_{\beps_i}$.

Next, address the computation of $\tilde{\bR}^{-1}_{\beps_i}$. Observe from Eq.~\eqref{eq:RepsKuNoisy} that the null space of $\tilde{\bR}_{\beps_i}$ is in the direction of $\bK^{-1}_i \tilde{\bar{\bu}}_i = \tilde{\bar{\bx}}_i$. Observe also that $\left[ \tilde{\bar{\bx}}_i \times \right]$ has a null space in the same direction. Thus, because of the alignment of the null spaces, the central term in the cost function of Eq.~\eqref{eq:LOSTCostFcn1} may be replaced with the identity 
\begin{equation}
\left[ \tilde{\bar{\bx}}_i \times \right] \tilde{\bR}^{-1}_{\beps_i} \left[ \tilde{\bar{\bx}}_i \times \right] = \left[ \tilde{\bar{\bx}}_i \times \right]  \tilde{\bR}^{\dagger}_{\beps_i} \left[ \tilde{\bar{\bx}}_i \times \right]
\end{equation}
where the superscript $\dagger$ indicates the pseudoinverse. It is always possible to find pseudoinverse numerically with the SVD of the $3 \times 3$ matrix $\tilde{\bR}_{\beps_i}$. In most cases, however, this is not the preferred approach. Instead, the pseudoinverse may usually be computed analytically as discussed in the Appendix.

Finally, address the issue of the unknown range $\rho_i$. Dealing with this unknown range was a central issue for the trigonometric solutions presented in Section~\ref{Sec:TrigSolutions}, where it was either avoided [e.g., DLT: see Eq.~\eqref{eq:DLT1}] or directly estimated [e.g., explicit range: see Eq.~\eqref{eq:RangeLinSys}].
Thus, the unmitigated dependence on $\rho_i$ (or on $\br$) would seem to suggest an iterative algorithm, where the unknown range is derived from the previous iteration. Another option is to attempt some kind of approximation in special situations \cite{Kobylka:2021}. However, the unknown range $\rho_i$ may be constructed directly from the measurements themselves.
Interestingly, the solution for how to obtain the unknown range was suggested by Frisius in 1533 in the same booklet \cite{Frisius:1533} that (re)introduced the idea of triangulation to the modern world. Making use of Haasbroek’s translation of the original Latin, Frisius observes ``that the distances can also be computed `with the tables of sine, but I omitted this intentionally as it is too difficult for the common man' '' \cite{Haasbroek:1968}. To the modern (and \emph{un}common!) [wo]man, Frisius's suggestion is most straightforwardly mechanized by employing the Law of Sines,
\begin{equation}
    \label{eq:LawOfSinesLOST1}
        \frac{ \| \bd_{ij} \| }{\sin \theta_{ij}} = \frac{\rho_i}{\sin \psi_{ij}}
\end{equation}
with angles as defined in Fig.~\ref{fig:LawOfSinesGeom}. Recall that $\bd_{ij}$ may be computed from Eq.~\eqref{eq:Defdij} and is a known quantity. Thus, the sine functions in Eq.~\eqref{eq:LawOfSinesLOST1} may be rewritten in terms of the cross-product of known vectors, and it follows that the distance $\rho_i$ is given by
\begin{equation}
       \rho_i =  \frac{ \| \bd_{ij} \| \sin \psi_{ij} }{\sin \theta_{ij}} = \frac{\| \bd_{ij} \times \ba_{j} \|}{ \| \ba_i \times \ba_j \|}
\end{equation}
where everything on the right-hand side is known. This expression for $\rho_i$ is true in any consistent frame. Since LOS unit vectors $\ba_i$ and $\ba_j$ may come from different cameras having different frames ($C_i$ and $C_j$), one may perform the cross products in the localization frame
\begin{equation}
       \tilde{\rho}_i = \frac{\| \bT^{C_i}_I \bK^{-1}_i \tilde{\bar{\bu}}_i \| \, \| \bT^{C_j}_I \bK^{-1}_j \tilde{\bar{\bu}}_j \| \, \| \bd_{I_{ij}} \times \bT^{C_j}_I \bK^{-1}_j \tilde{\bar{\bu}}_j \|}{ \| \bT^{C_j}_I \bK^{-1}_j \tilde{\bar{\bu}}_j \| \, \| \bT^{C_i}_I \bK^{-1}_i \tilde{\bar{\bu}}_i  \times \bT^{C_j}_I \bK^{-1}_j \tilde{\bar{\bu}}_j  \|} 
       = \frac{\| \bK^{-1}_i \tilde{\bar{\bu}}_i \| \, \| \bd_{I_{ij}} \times \bT^{C_j}_I \bK^{-1}_j \tilde{\bar{\bu}}_j \|}{ \| \bT^{C_i}_I \bK^{-1}_i \tilde{\bar{\bu}}_i  \times \bT^{C_j}_I \bK^{-1}_j \tilde{\bar{\bu}}_j  \|} 
\end{equation}
where, once again, the measured $\tilde{\bar{\bu}}_i$ is used in place of the true $\bar{\bu}_i$ (this is acceptable when the approximation is \emph{only} used for approximating the covariance $\bR_{\beps_i}$). Thus, applying these observations to Eq.~\eqref{eq:RepsKu} yields an expression for $\tilde{\bR}_{\beps_i}$ that is written entirely in terms of known (or measured) quantities
\begin{equation}
    \label{eq:RepsKu2}
    \tilde{\bR}_{\beps_i} = - 
    \gamma^2_i \left[  \bK^{-1}_i \tilde{\bar{\bu}}_i \times \right] \bR_{\bar{\bx}_i} \left[  \bK^{-1}_i \tilde{\bar{\bu}}_i \times \right]
\end{equation}
where
\begin{equation}
    \label{eq:Defgamma}
    \gamma_i = \frac{\tilde{\rho}_i}{\| \bK^{-1}_i \tilde{\bar{\bu}}_i  \|} = \frac{\| \bd_{I_{ij}} \times \bT^{C_j}_I \bK^{-1}_j \tilde{\bar{\bu}}_j \|}{ \| \bT^{C_i}_I \bK^{-1}_i \tilde{\bar{\bu}}_i  \times \bT^{C_j}_I \bK^{-1}_j \tilde{\bar{\bu}}_j  \|}
\end{equation}
Finally, for any given any LOS measurement $\tilde{\bl}_i \propto \bK_i^{-1} \tilde{\bar{\bu}}_i$, there is some ambiguity in which additional LOS measurement $\tilde{\bl}_j \propto \bK_j^{-1} \tilde{\bar{\bu}}_j$ (for $i \neq j$) to use for computing $\gamma_i$ with Eq.~\eqref{eq:Defgamma}. Experience has shown that any reasonable scheme (even a random one) provides essentially the same triangulation performance in practice (i.e., small variations in $\gamma_i$ do not change the relative measurement weightings in the MLE enough to matter). Efforts to find the ``optimal'' choice of a companion measurement $j$ do not appear to be worth the additional computational effort.

Now, to find the optimal solution, return to Eq.~\eqref{eq:LOSTCostFcn1} and apply the first differential condition
\begin{equation}
     \sum_{i=1}^n -2(\hat{\br}_I - \bp_{I_i} )^T \bT_I^{C_i}\left[ \tilde{\bar{\bx}}_i \times \right] \tilde{\bR}^{\dagger}_{\beps_i}  \left[ \tilde{\bar{\bx}}_i \times \right] \bT^I_{C_i} = 0
\end{equation}
Substituting for $\tilde{\bar{\bx}}_i = \bK^{-1}_i \tilde{\bar{\bu}}$ and rearranging gives
\begin{equation}
     \left( \sum_{i=1}^n \bT_I^{C_i}\left[ \bK^{-1}_i \tilde{\bar{\bu}}_i \times \right] \tilde{\bR}^{\dagger}_{\beps_i}  \left[ \bK^{-1}_i \tilde{\bar{\bu}}_i \times \right] \bT^I_{C_i} \right) \hat{\br}_I = \sum_{i=1}^n \bT_I^{C_i}\left[ \bK^{-1}_i \tilde{\bar{\bu}}_i \times \right] \tilde{\bR}^{\dagger}_{\beps_i}  \left[ \bK^{-1}_i \tilde{\bar{\bu}}_i \times \right] \bT^I_{C_i} \bp_{I_i}
     \label{eq:NormEqnsLOST}
\end{equation}
As expected, this final result is nothing more than the normal equations for an optimally weighted DLT (with weighting $\bW_i = \bR^{\dagger}_{\beps_i} \sim \bR^{-1}_{\beps_i}$). Following the usual approach for solving any least squares system \cite{Goulb:2013}, the numerical stability may be improved by avoiding explicit formation of the normal equations and instead solving (e.g., with QR factorization)
\begin{equation}
    \begin{bmatrix}
        \bB_1 \left[ \bK_1^{-1} \tilde{\bar{\bu}}_1 \times \right] \bT^I_{C_1}  \\
        \bB_2 \left[ \bK_2^{-1} \tilde{\bar{\bu}}_2 \times \right] \bT^I_{C_2}  \\
        \vdots \\
        \bB_n \left[ \bK_n^{-1} \tilde{\bar{\bu}}_n \times \right] \bT^I_{C_n} 
    \end{bmatrix}
    \br_I = 
    \begin{bmatrix}
        \bB_1 \left[ \bK_1^{-1} \tilde{\bar{\bu}}_1 \times \right] \bT^I_{C_1} \bp_{I_1} \\
        \bB_2 \left[ \bK_2^{-1} \tilde{\bar{\bu}}_2 \times \right] \bT^I_{C_2} \bp_{I_2} \\
        \vdots \\
        \bB_n \left[ \bK_n^{-1} \tilde{\bar{\bu}}_n \times \right] \bT^I_{C_n} \bp_{I_n}
    \end{bmatrix}
\end{equation}
which makes use of the factorization $\tilde{\bR}^{\dagger}_{\beps_i} = \bB_i^T \bB_i$. In the case that $\bR_{\bx_i} = \sigma^2_{\bx_i} \bI_{2 \times 2}$, the results of the Appendix show that
\begin{equation}
     \bB_i = \frac{q_i}{\| \bK^{-1}_i \tilde{\bar{\bu}}_i\|^2 } \bS \left[ \bK^{-1}_i \tilde{\bar{\bu}}_i \times \right]^2
\end{equation}
where
\begin{equation}
    \label{eq:Defqi}
     q_i = \frac{1}{\sigma_{\bx_i} \gamma_i} = \frac{\| \bK^{-1}_i \tilde{\bar{\bu}}_i \| }{\sigma_{\bx_i} \rho_i} = \frac{ \| \bT^{C_i}_I \bK^{-1}_i \tilde{\bar{\bu}}_i  \times \bT^{C_i}_I \bK^{-1}_j \tilde{\bar{\bu}}_j  \|}{\sigma_{\bx_i} \| \bd_{I_{ij}} \times \bT^{C_j}_I \bK^{-1}_j \tilde{\bar{\bu}}_j \|}
\end{equation}
Recalling the identity $[\bb \times ]^3 = -\| \bb \|^2 [\bb \times ]$, it follows that the above simplifies to
\begin{equation}
    \begin{bmatrix}
        q_1 \bS \left[ \bK_1^{-1} \tilde{\bar{\bu}}_1 \times \right] \bT^I_{C_1}  \\
        q_2 \bS  \left[ \bK_2^{-1} \tilde{\bar{\bu}}_2 \times \right] \bT^I_{C_2}  \\
        \vdots \\
        q_n \bS \left[ \bK_n^{-1} \tilde{\bar{\bu}}_n \times \right] \bT^I_{C_n} 
    \end{bmatrix}
    \br_I = 
    \begin{bmatrix}
        q_1 \bS \left[ \bK_1^{-1} \tilde{\bar{\bu}}_1 \times \right] \bT^I_{C_1} \bp_{I_1} \\
        q_2 \bS \left[ \bK_2^{-1} \tilde{\bar{\bu}}_2 \times \right] \bT^I_{C_2} \bp_{I_2} \\
        \vdots \\
        q_n \bS \left[ \bK_n^{-1} \tilde{\bar{\bu}}_n \times \right] \bT^I_{C_n} \bp_{I_n}
    \end{bmatrix}
    \label{eq:LOSTforQR}
\end{equation}
Note here that the matrix $\bS = [\bI_{2 \times 2}, \textbf{0}_{2 \times 1}]$ removes the redundant third row for each of the measurements. 
The reader is reminded that the two remaining rows were shown in Section~\ref{Sec:CollinearityEqns} to be equivalent to the collinearity equations. Moreover, substituting from Eq.~\eqref{eq:Defqi}, one finds that
\begin{equation}
     q_i \bS \left[ \bK_i^{-1} \tilde{\bar{\bu}}_i \times \right] \bT^I_{C_i} = \frac{\| \bK_i^{-1} \tilde{\bar{\bu}}_i \|^2}{\sigma_{\bx_i} \rho_i } \bS \left[ \tilde{\ba}_{C_i} \times \right] \bT^I_{C_i}
\end{equation}
It is observed that a particular LOS measurement is weighted less as either the measurement error ($\sigma_i$) or the range ($\rho_i$) increase. 

Whether it is ultimately best to solve for $\hat{\br}_I$ via the normal equations [Eq.~\eqref{eq:NormEqnsLOST}] or by QR factorization [Eq.~\eqref{eq:LOSTforQR}] depends on the problem at hand and the priorities of the analyst. The normal equations solution requires less memory and fewer computations (when $n$ is large), while the QR factorization tends to have slightly better numerical stability. The numerical stability of the QR factorization approach becomes more important as the angle(s) between the LOS measurements become smaller. It is also straightforward to write Eq.~\eqref{eq:LOSTforQR} as a total least squares problem, which may be solved via singular value decomposition.

Since the weighting is statistically optimal, the covariance may be computed as
\begin{equation}
\label{eq:optDLT_cov}
     \bP_{\br} = - \left( \sum_{i=1}^n \bT_I^{C_i}\left[ \bK^{-1}_i \tilde{\bar{\bu}} \times \right]  \tilde{\bR}^{\dagger}_{\beps_i}  \left[ \bK^{-1}_i \tilde{\bar{\bu}} \times \right] \bT^I_{C_i} \right)^{-1}
\end{equation}
where, again, the leading negative sign is a result of the identity $[ \, \cdot \, \times]^T = - [ \, \cdot \, \times]$. In the case of $\bR_{\bx_i} = \sigma^2_{\bx_i} \bI_{2 \times 2}$ computing the pseudoinverse can be avoided and
\begin{equation}
    \label{eq:optDLT_cov2}
     \bP_{\br} = - \left( \sum_{i=1}^n 
     q^2_i \bT^{C_i}_I \left[ \bK^{-1}_i \tilde{\bar{\bu}}_i  \times \right] \bS^T \bS  \left[ \bK^{-1}_i \tilde{\bar{\bu}}_i \times \right] \bT^I_{C_i} \right)^{-1}
\end{equation}
where $q_i$ is from Eq.~\eqref{eq:Defqi}. Since $\bK^{-1}_i$ may be written directly (see Ref.~\cite{Christian:2021}), computing the covariance $\bP_r$ only requires a single matrix inverse. This is fast to compute since the inverse is of a symmetric $3 \times 3$ matrix. 

Finally, though the authors are unaware of prior weighted DLT methods that are both optimal and direct (non-iterative), the LOST method proposed here will likely seem obvious to the contemporary practitioner of estimation theory. What becomes clear, however, is that the widespread claims of suboptimality of the DLT are often misplaced. The suboptimality is not a shortcoming of the DLT itself, but a shortcoming of the choice to solve the resulting equations as an unweighted least squares problem. Accounting for the different uncertainty in different measurements is essential to any estimation problem---be that triangulation with the DLT, navigating a spacecraft, or estimating anything else. Indeed, handling the the relative weighting of observations was one of Fisher's original complaints against the method of ordinary least squares that led to the development of the MLE approach between 1912 and 1922 \cite{Aldrich:1997}. Thus, the authors make no claim that the concepts behind the LOST method are especially novel. The LOST algorithm is simply the natural consequence of applying the MLE framework to the DLT algorithm, while recognizing the Law of Sines may be used to efficiently acquire the measurement covariance.

\section{Dynamic Triangulation}
\label{Sec:DynamicTri}
The concepts from triangulation are not confined to concurrent measurements (or to static systems). Triangulation with moving objects was pioneered for maritime navigation and tracking after widespread deployment of radar systems in the 1940s---largely under the name bearings-only target motion analysis (BO-TMA) \cite{Nardone:1984}. Similar concepts were considered for applications in spacecraft orbit determination by Kaplan \cite{Kaplan:2011}, though these historical presentations often obscure elegant relations to triangulation (especially the DLT from Section~\ref{Sec:DLT}). As with other triangulation methods, there is no meaningful difference between the algorithms to address the intersection (e.g., orbit determination) and resection (e.g., navigation) forms of the dynamic triangulation problem. 

Consider a camera aboard a vehicle whose state $\bxi$ evolves according the linear dynamics
\begin{equation}
   \dot{\bxi}(t) = \bA(t) \, \bxi(t)
\end{equation}
Such a dynamical system exhibits a solution of the form \cite{Gelb:1974}
\begin{equation}
    \label{eq:LinearEOM}
  \bxi_i = \bPhi(t_i,t_0) \bxi_0
\end{equation}
where $\bPhi(t_i,t_0)$ is the state transition matrix (STM) that advances the state from $t_0$ to $t_i$. Suppose that one of the dynamical states within $\bxi$ is the vehicle's position $\br$. Let the rows of the STM $\bPhi$ corresponding to $\br$ be denoted by $\bPhi_\br$ such that
\begin{equation}
  \br_i = \bPhi_{\br}(t_i,t_0) \bxi_0
\end{equation}

Thus the LOS measurement at any point in time is given by
\begin{equation}
  \bl_i \propto \bp_i - \br_i =  \bp_i - \bPhi_{\br}(t_i,t_0) \bxi_0
\end{equation}
Following the same approach as for the DLT in Section~\ref{Sec:DLT}, take the cross product with the measured LOS direction,
\begin{equation}
  \left[ \bl_i \times \right] \bp_i - \left[ \bl_i \times \right] \bPhi_{\br}(t_i,t_0) \bxi_0 = \textbf{0}_{3 \times 1}
\end{equation}
such that 
\begin{equation}
  \left[ \bl_i \times \right] \bPhi_{\br}(t_i,t_0) \bxi_0 = \left[ \bl_i \times \right] \bp_i 
\end{equation}
Stacking these equations for many measurements leads to a modified version of the DLT,
\begin{equation}
    \begin{bmatrix}
        \left[ \bl_1 \times \right] \bPhi_{\br}(t_1,t_0) \\
        \vdots \\
        \left[ \bl_n \times \right] \bPhi_{\br}(t_n,t_0) 
    \end{bmatrix} \bxi_0 = 
    \begin{bmatrix}
        \left[ \bl_1 \times \right] \bp_1  \\
        \vdots \\
        \left[ \bl_n \times \right] \bp_n 
    \end{bmatrix} 
    \label{eq:DynamicDLT}
\end{equation}
which is true in any consistent frame. This formulation is nothing more than a specialized form of classical batch estimation, which is widely used in astrodynamics applications \cite{Tapley:2004}. 

For a STM $\bPhi_{\br}(t_i,t_0)$ of general form and LOS measurements to a variety of different points $\bp_i \neq 0$, one may usually solve for the full state vector $\bxi_0$. However, under certain degeneracies in dynamics, sampling times, or point configurations, it is sometimes not possible to solve for $\bxi_0$ without ambiguity. The variations of degeneracies are many and their full enumeration is not practical. However, the most common degeneracy occurs when all LOS measurements are between a moving point and a single reference point at the origin [or a reference point that can be shifted to the origin while maintaining the linear dynamics of Eq.~\eqref{eq:LinearEOM}], which leads to a family of homothetic solutions characterized by an unobservable scale (e.g., Fig.~\ref{fig:HomotheticTraj}).  As an example of this, the common case of spacecraft relative navigation is discussed in Section~\ref{Sec:ExampleRelNav}.

\section{Example Applications}
\label{Sec:Applications}

\subsection{Terrain Relative Navigation (TRN)}
\label{Sec:ExampleTRN}
Imaging and triangulation play a crucial role in Terrain Relative Navigation (TRN). While TRN does have a variety of terrestrial applications \cite{Carr:1980,Golden:1980,Kim:2004}, it is particularly important in interplanetary space exploration where few other navigation observables are available. Specific examples include orbit determination from landmarks \cite{Miller:2003,Cheng:2003,Christian:2019} and precision landing \cite{Cheng:2005,Adams:2008,Steffes:2019,Olds:2022}. TRN often takes the form of absolute triangulation since the camera-to-terrain attitude may be obtained using knowledge of the spacecraft's inertial attitude (e.g., from star trackers \cite{Liebe:1995,Liebe:2002,Christian:2021star}) and the celestial body's inertial attitude (e.g., from ephemeris files \cite{Acton:1996,Acton:2018}).

As an example, consider a simulated lander descending from an altitude of 2,000 m to an altitude of 200 m.
During this time, a camera onboard the lander observes two points (either natural or artificial) on the surface of the celestial body. Suppose the lander's camera is similar to the Mars 2020 LCAM \cite{johnson:2017,Maki:2020}, having a FOV of $90 \times 90$ deg and a detector with $1024\times1024$ pixels. 
At regular intervals along a simulated descent trajectory, a Monte Carlo simulation is performed with 1,000 sample points generated by Eq.~\eqref{eq:LOSnoisy} with standard deviation of $\sigma_{\bu_i} = 0.1$ pixel. Here, a comparison is made between the two trigonometric solutions (DLT and explicit range) and the optimal polynomial solution from Section~\ref{Sec:QUAT}.  
For each of these three methods, the analytic covariance (see Appendix) is compared with the Monte Carlo sample covariance. 
These comparisons are made for two different scenarios.

\begin{figure}[b!]
    \centering
    \includegraphics[width=0.6\linewidth]{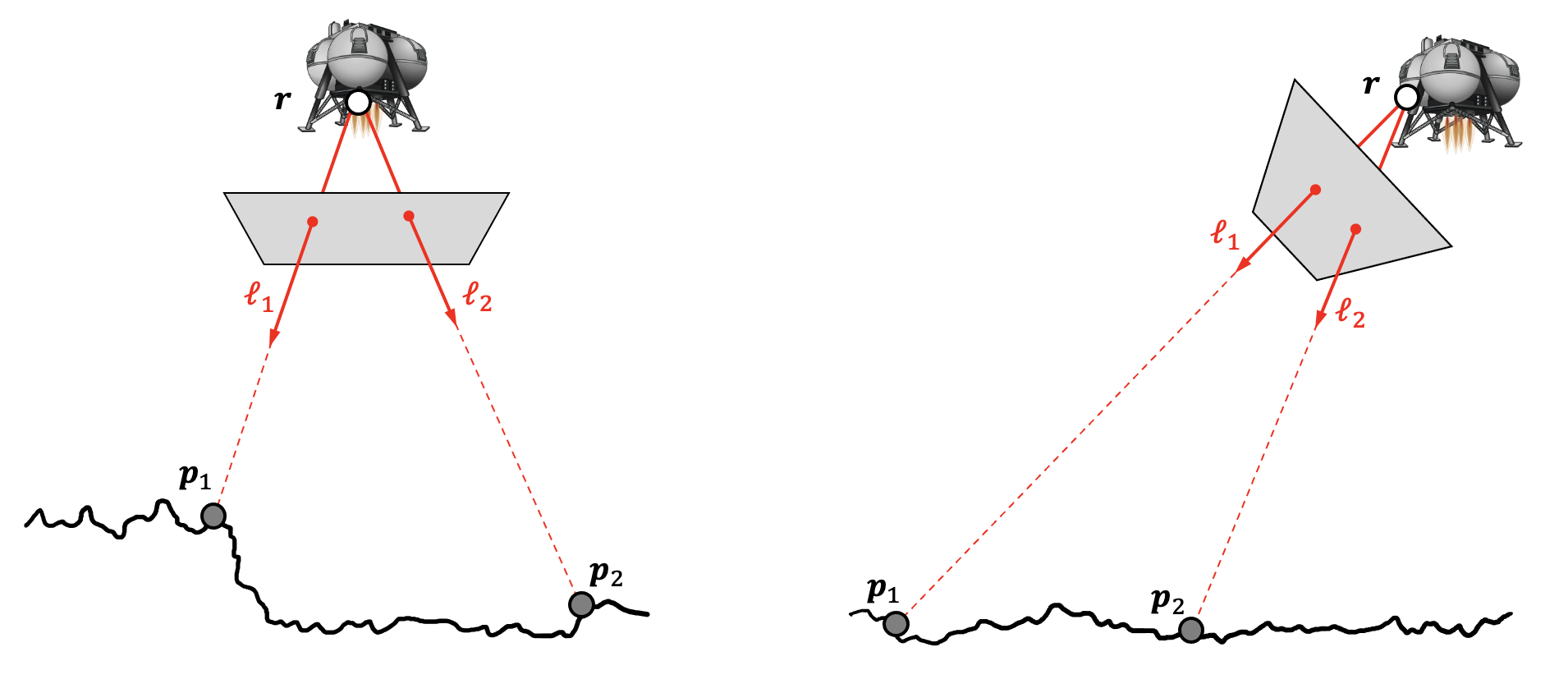}
    \caption{Illustration of a lander using two surface points for TRN. The first scenario is for a nadir pointed camera (left) and the second scenario is for a camera pointed 45 deg off-nadir (right).} 
    \label{fig:lander_drawing}
\end{figure}

\begin{figure}[b!]
\centering
\includegraphics[width=1\columnwidth,trim=0in 0in 0in 0in,clip]{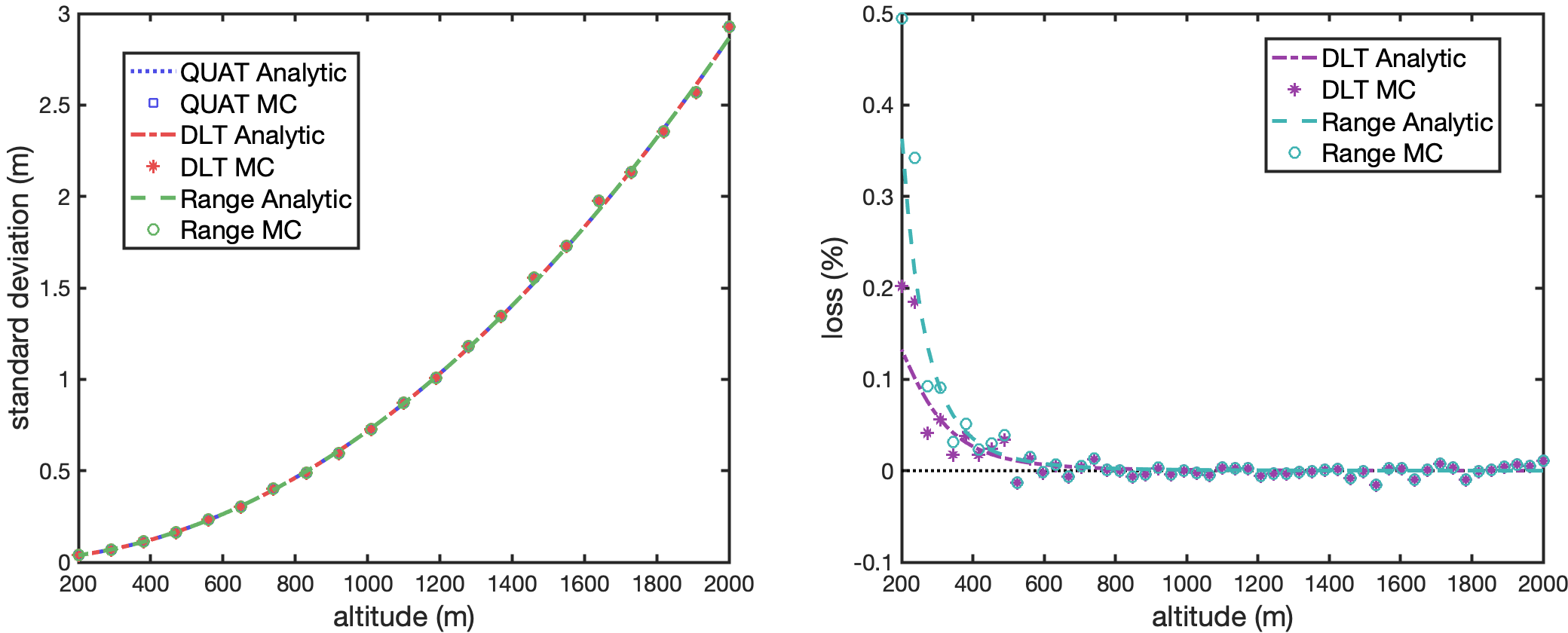}
	\caption{Analytical (lines) and Monte Carlo (dots) total standard deviation of the triangulated position for the first TRN scenario (nadir camera). Left frame shows the standard deviation for all methods. Right frame shows the percentage of increase in standard deviation (loss in precision) of the trigonometric solutions as compared to the quadratic polynomial (QUAT) method.}
	\label{fig:TRN_straight}
\end{figure}

In a first scenario, consider a downward (nadir) pointed camera that observes two points on the ground. This is illustrated in the left-hand frame of Fig.~\ref{fig:lander_drawing}. The two points are placed 300 m apart and centered beneath the lander. One point ($\bp_1$) has an elevation of 30 m above the other point ($\bp_2$).
As the lander descends, the apparent angular separation of the two objects increases---and so their distance to one another on the image plane also increases. Now, define the total standard deviation of the position estimate $\sigma_{\br}$ as
\begin{equation}
    \label{eq:TotalSTD}
   \sigma_{\br} = \sqrt{\text{Tr}\left[ \bP_{\br} \right]}
\end{equation}
where $\text{Tr}[\, \cdot \,]$ is the trace operator and $\bP_{\br}$ is the error covariance of the triangulated position $\hat{\br}$. This answer is the same in any frame since $\text{Tr}\left[ \bP_{\br} \right] = \text{Tr}\left[ \bT \bP_{\br} \bT^T \right]$. Figure~\ref{fig:TRN_straight} shows $\sigma_{\br}$ for each of the triangulation methods (left-hand frame) and the relative increase in standard deviation (or loss in precision) of the two trigonometric methods against the optimal polynomial method (right-hand frame).  In this particular case, there is nearly no difference between the optimal solution (labeled QUAT), the DLT, and the explicit range. This follows the intuition gained from Section~\ref{Sec:LOST} which explains how the optimal weighting of the DLT is inversely proportional to the measurement uncertainty and the range. Since both the measurement uncertainty and the range are about the same for both measurements (due to the symmetry of this example scenario) the DLT and explicit range methods happen to have a nearly optimal weighting (by luck rather than design). As the lander continues to descend, the small 30 m altitude difference between $\bp_1$ and $\bp_2$ accounts for a larger percentage of the range, and so small differences begin to arise at very low altitudes (e.g., below about 600 m in the right-hand frame of Fig.~\ref{fig:TRN_straight}). This motivates a second scenario with larger range differences.

\begin{figure}[b!]
\centering
\includegraphics[width=1\columnwidth,trim=0in 0in 0in 0in,clip]{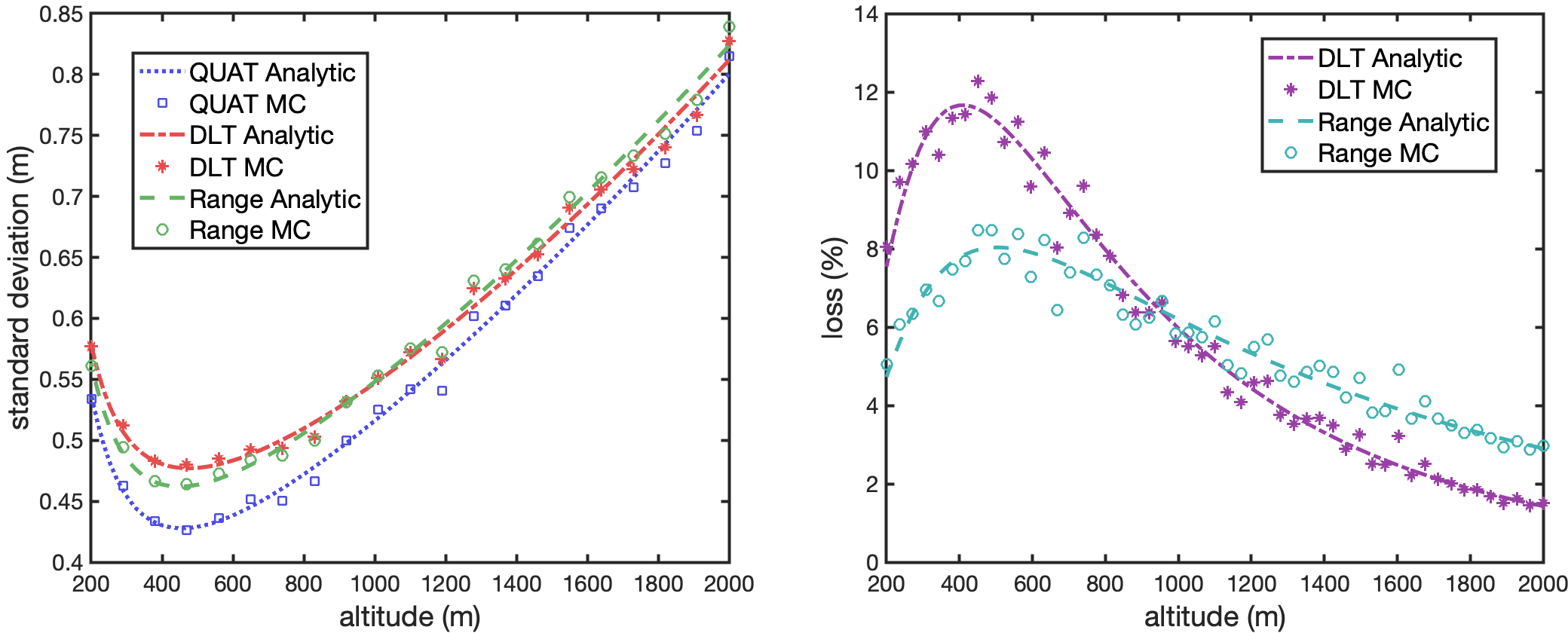}
	\caption{Analytical (lines) and Monte Carlo (dots) total standard deviation of the triangulated position for the second TRN scenario (45 deg off-nadir camera). Left frame shows the standard deviation for all methods. Right frame shows the percentage of increase in standard deviation (loss in precision) of the trigonometric solutions as compared to the quadratic polynomial (QUAT) method.}
	\label{fig:TRN_canted}
\end{figure}

\begin{figure}[t!]
\centering
\includegraphics[width=0.45\columnwidth,trim=0in 0in 0in 0in,clip]{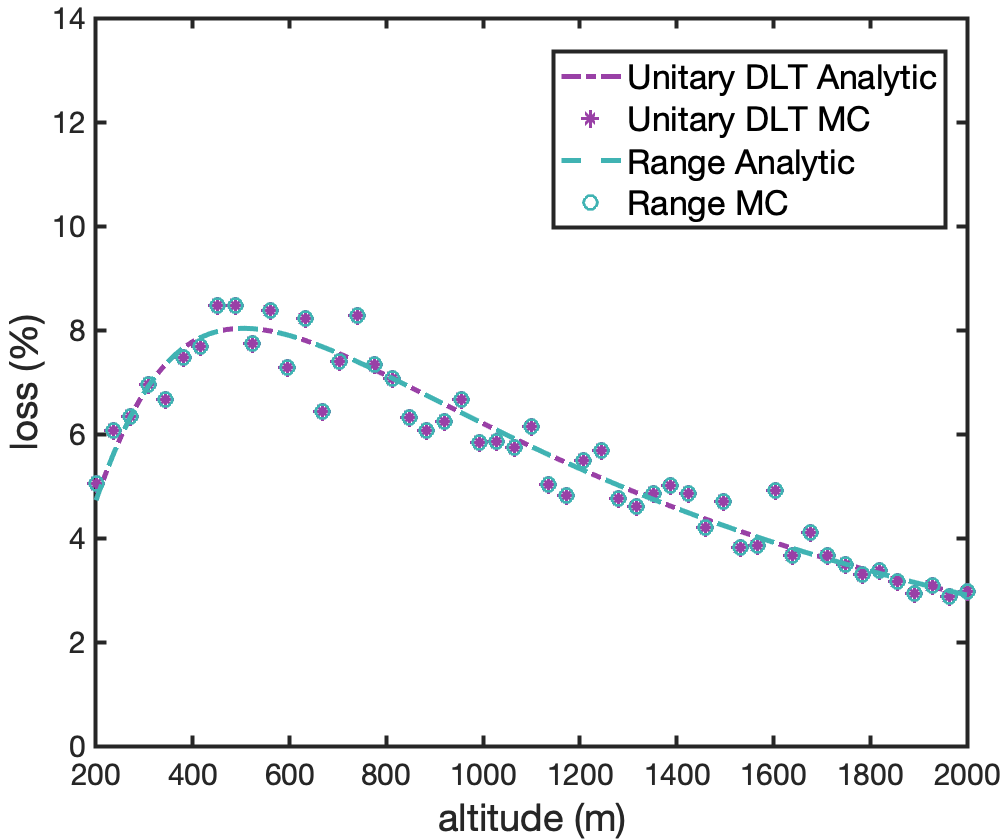}
\caption{The DLT with unit vector LOS measurements provides the exact same result as the explicit range method. Results are for the second TRN scenario (45 deg off-nadir camera).}
	\label{fig:TRN_canted_unitaryDLT}
\end{figure}

In a second scenario, illustrated in Fig.~\ref{fig:lander_drawing}, the camera is pointed 45 deg off nadir and observes two objects on the ground. The two objects are located at 3,000 m and 300 m from the point on the ground beneath the lander.  
As with the first scenario, triangulation performance is evaluated for the two triangulation methods (DLT and explicit range) and for the optimal polynomial method and results are shown in Fig.~\ref{fig:TRN_canted}. 
The improvement of the optimal solution triangulation is more pronounced in this case, with nearly 12\% of gain in relative precision over the DLT at 400 m. The small difference between the DLT and explicit range method is a result of the normalization chosen. If all the LOS measurements in the DLT are converted to unit vectors, then the DLT and explicit range methods produce identical estimates as shown in Fig.~\ref{fig:TRN_canted_unitaryDLT}. That the DLT and exlicit range methods are identical under such a normalization was discussed in Section~\ref{Sec:TrigSolnComp} and proven analytically in the Appendix. This provides a numerical validation of this fact.

\begin{figure}[t!]
\centering
\includegraphics[width=0.45\columnwidth,trim=0in 0in 0in 0in,clip]{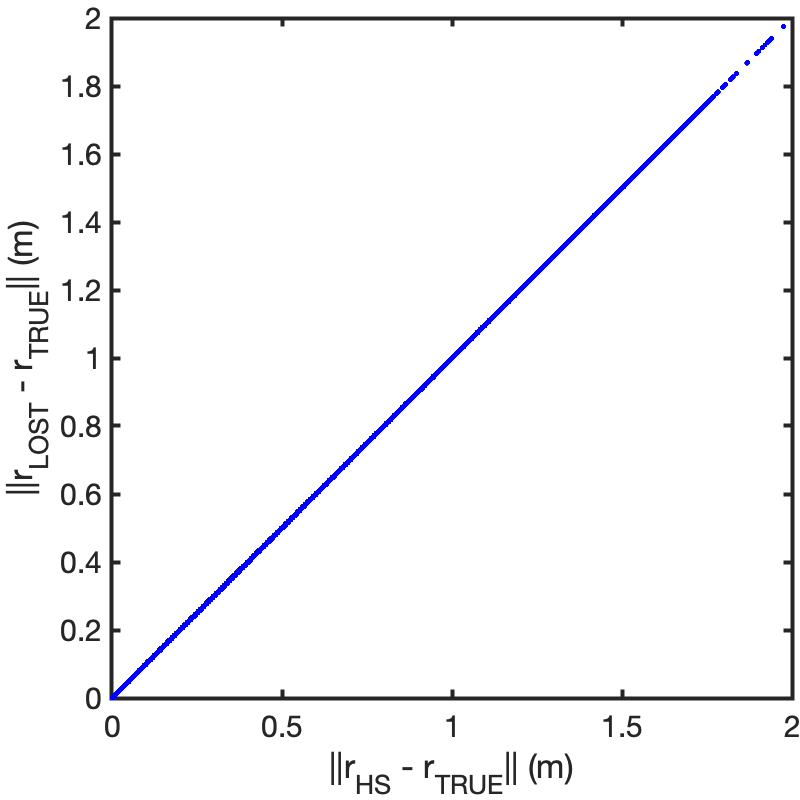}
\caption{The position errors are essentially the same for the the Hartley \& Sturm method and the LOST method. Results are for TRN scenario two (45 deg off-nadir camera) at an altitude of 1,000 m.}
	\label{fig:TRN_scatter_HS_LOST}
\end{figure}

The second scenario may also be used to illustrate the practical equivalence between the Hartley and Sturm method (labeled HS, see Ref.~\cite{Hartley:1997} and Section~\ref{Sec:TriTwoPointsHS}), the quadratic polynomial method (labeled QUAT, see Section~\ref{Sec:QUAT}), and the new LOST method (see Section~\ref{Sec:LOST}). At an altitude of 1 km, a one-million sample Monte Carlo analysis was performed and the results are summarized in Table~\ref{tab:TRN_opt_comparison}. As compared to the truth, the standard deviation of all three methods (HS, QUAT, LOST) are identical to the reported precision. All three are functionally zero mean, having mean errors about three orders of magnitude smaller than the standard deviation. When compared to one another, the HS and QUAT algorithms are identical to about 10 digits (standard deviation of difference on the order of $10^{-7}$ m on a position vector of length $10^3$ m). This is considered numerically equivalent when accounting for the fact that the HS solution relies on numerically finding the roots of a polynomial of degree six. The difference between the LOST and HS methods (standard deviation of $1.25\times10^{-4}$ m) is about three orders of magnitude smaller than the difference of either of these with respect to the truth (standard deviation of $4.38\times10^{-1}$ m). Such a difference is practically inconsequential. This can be made explicit by considering a scatter plot of the errors for each (see Fig.~\ref{fig:TRN_scatter_HS_LOST}), which clearly shows that one method is no better than the other. In this Monte Carlo analysis the LOST method was closer to the truth for 50.05\% of the cases, while the HS method was closer to the truth for 49.95\% of the cases.

\begin{table}[bt!]
\caption{TRN triangulation residual statistics for scenario two (45 deg off-nadir camera) at an altitude of 1,000 m.}
\label{tab:TRN_opt_comparison}
\centering
\begin{tabular}{lccccc}
\hline 
\hline
                      & HS vs TRUE   & QUAT vs TRUE & LOST vs TRUE & QUAT vs HS   & LOST vs HS   \\ \hline
mean x (m) & $-5.2118\times 10^{-4}$ & $-5.2118\times10^{-4}$ & $-4.6149\times10^{-4}$ & $4.6997\times10^{-11}$  & $5.9695\times10^{-5}$  \\
mean y (m) & $-4.2484\times10^{-5}$ & $-4.2484\times10^{-5}$ & $-4.2484\times10^{-5}$ & $-3.6710\times10^{-11}$ & $2.4340\times10^{-12}$  \\
mean z (m) & $7.3771\times10^{-4}$  & $7.3771\times10^{-4}$  & $6.7233\times10^{-4}$  & $-5.1492\times10^{-11}$ & $-6.5378\times10^{-5}$ \\
standard dev (m)      & $4.3804\times10^{-1}$  & $4.3804\times10^{-1}$  & $4.3804\times10^{-1}$  & $1.0414\times10^{-7}$  & $1.2507\times10^{-4}$ \\ 
\hline 
\hline
\end{tabular}
\end{table}

\subsection{Interplanetary Exploration}
\begin{figure}[b!]
    \centering
    \includegraphics[width=0.4\linewidth]{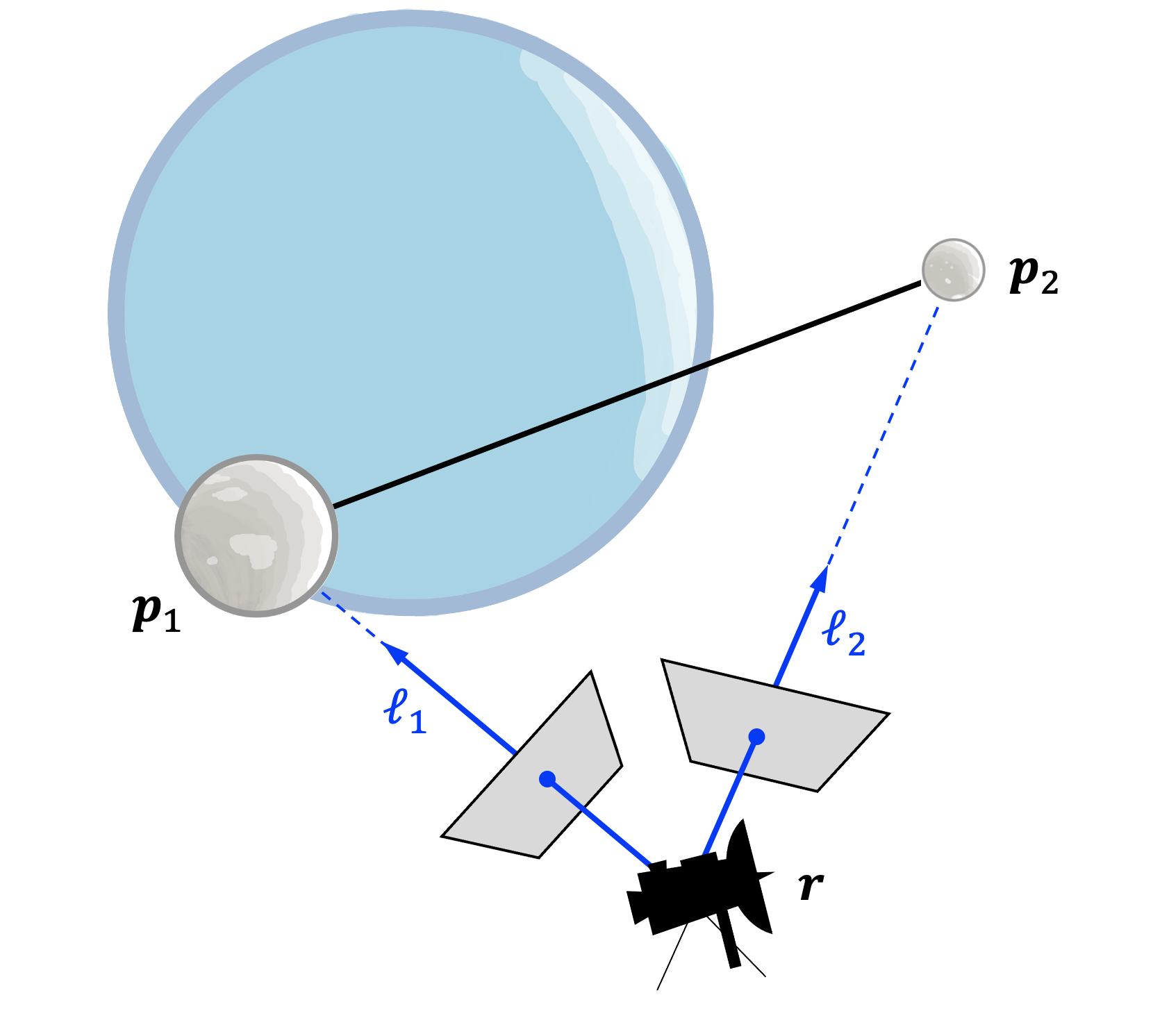}
    \caption{Spacecraft determining its position by triangulation using two moons around Uranus.}
    \label{fig:uranus_drawing}
\end{figure}

Triangulation is expected to play an important role in future exploration missions to the outer planets. Both the gas giants (Jupiter and Saturn) and the ice giants (Uranus and Neptune) have a large number of moons, and it is possible to use LOS measurements to these moons to triangulate a spacecraft's location. 
The U.S. National Academies have identified the Uranus Orbiter and Probe (UOP) as ``the highest-priority new Flagship mission for initiation in the decade 2023-2032'' \cite{PlanetaryDecadalSurvey:2022}, which makes Uranus a prime candidate for further study. A simulated scenario is created where a spacecraft observes the moons Titania and Oberon, with a different camera pointing in the direction of each of the moons. An illustration of this is found in Fig.~\ref{fig:uranus_drawing}. 

Supposing an epoch of 1 January 2050 00:00:0.0 UTC, the known locations of Titania ($\bp_1$) and Oberon ($\bp_2$) relative to Uranus are 
\begin{equation}
    \bp_1 = 
    \begin{bmatrix}
        2.8607\times 10^5 \\ -3.2961\times 10^5 \\ -3.3944\times 10^2
    \end{bmatrix} \text{ km}
    \quad \quad
    \bp_2 = 
    \begin{bmatrix}
        5.0811 \times 10^5 \\ -2.8608 \times 10^5 \\ -9.0978\times 10^2
    \end{bmatrix} \text{ km} \nonumber
\end{equation}
These positions were obtained from the \texttt{URA111} ephemeris files \cite{Jacobson:2014} and processed using the SPICE toolkit \cite{Acton:1996,Acton:2018}. These two moons are then viewed by simulated cameras aboard a simulated spacecraft residing somewhere in the equatorial plane of Uranus. Although the true spacecraft location is in the equatorial plane, the moons Titania and Oberon are not in this plane---and, thus, the triangulation problem considered here is fully 3-D. Moreover, this specific arrangement and choice of epoch allows for qualitative comparison of these results with the moon-based localization study provided in Ref.~\cite{Lubey:2020}.

It is not possible to practically image a specific moon from everywhere in the Uranian system. Thus, following a similar approach as Ref.~\cite{Lubey:2020}, a number of visibility constraints are adopted. For a moon to be visible, it cannot be blocked from view in any way by Uranus, and needs to have at least 5\% of FOV clearance from Uranus. Additionally, the moon is not visible if it occupies more than 80\% of the FOV. Finally, the spacecraft is not able to see the moon if the LOS to the sun is within 30 deg of the LOS to the moon. The triangulation is only performed in the case where both moons are visible. To facilitate easy comparison, this analysis uses the mid-resolution camera from Ref.~\cite{Lubey:2020}, with a FOV of $7$ degrees and an IFOV of $60$ $\mu$rad.

This simulated scenario (of practical relevance to future interplanetary exploration) may be used to illustrate a number of important observations about triangulation algorithms. These observations are facilitated by the evaluation of results from a Monte Carlo analysis. Specifically, consider a $(3\times10^6)\times (3\times10^6)$ km region centered around Uranus. At each point within this region, a Monte Carlo analysis was preformed with LOS errors generated according to Eq.~\eqref{eq:LOSnoisy}. These noisy LOS data are used to triangulate the spacecraft position using the DLT method from Section~\ref{Sec:DLT}, Hartley and Sturm (HS) polynomial method from Ref.~\cite{Hartley:1997} (and Section~\ref{Sec:TriTwoPointsHS}), and the new LOST method from Section~\ref{Sec:LOST}. 

\begin{figure}[tb!]
    \centering
    \includegraphics[width=1\linewidth]{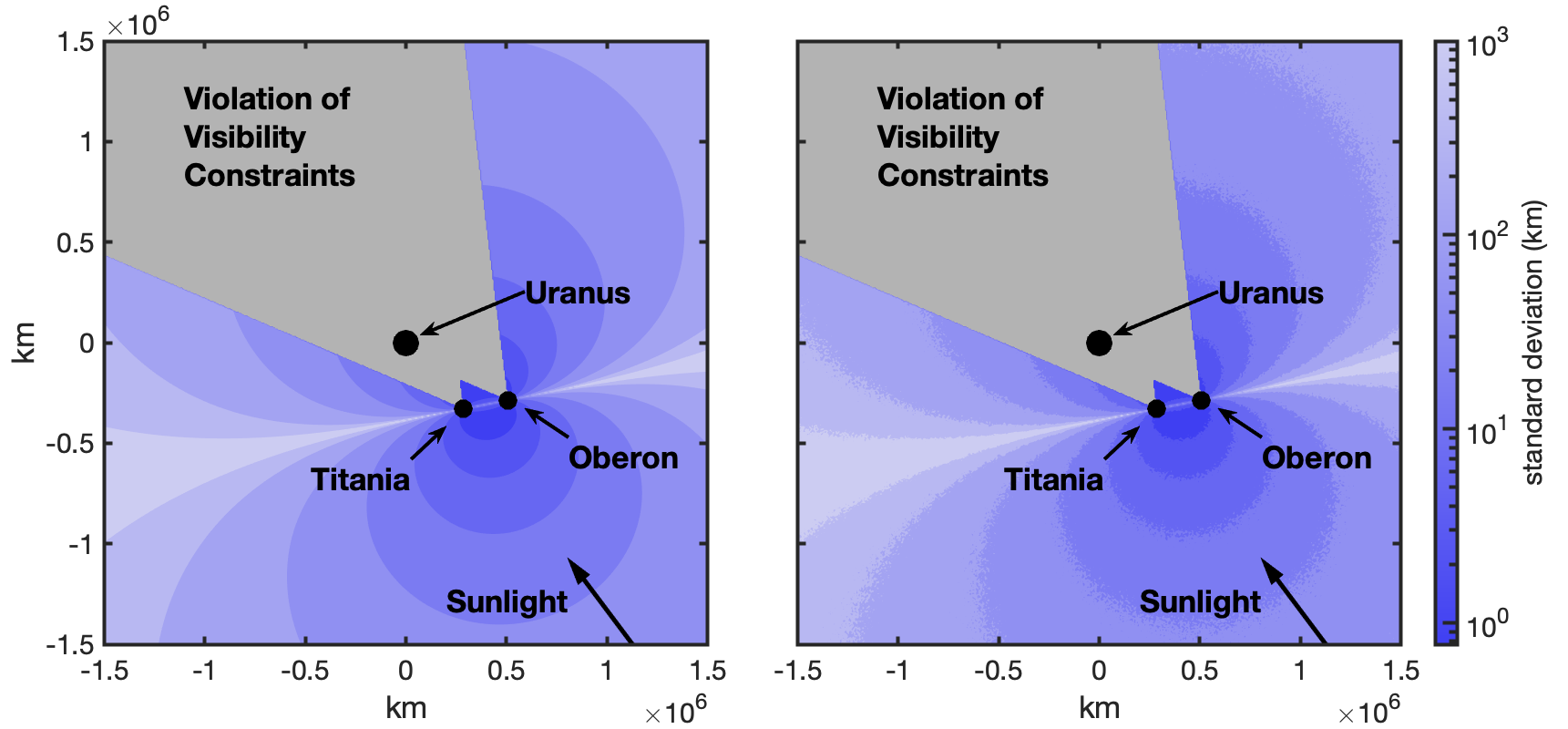}
    \caption{Analytical (left) and Monte Carlo (right) standard deviations of the spacecraft position with the LOST method.}
    \label{fig:uranus_std_OPT}
\end{figure}

\begin{figure}[tb!]
    \centering
    \includegraphics[width=1\linewidth]{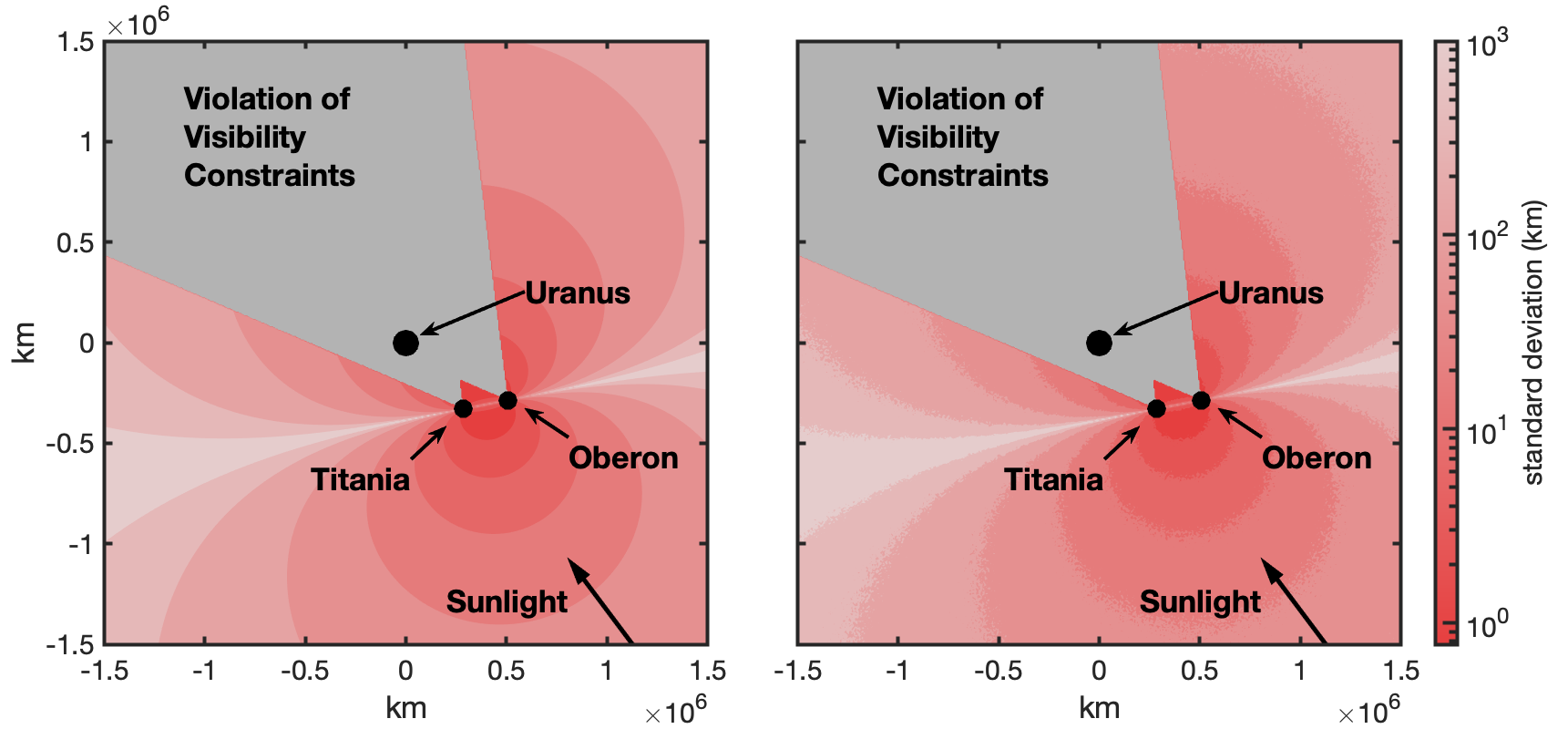}
    \caption{Analytical (left) and Monte Carlo (right) standard deviations of the spacecraft position with the DLT method.}
    \label{fig:uranus_std_DLT}
\end{figure}

The first use of the Monte Carlo analysis is to validate the analytic covariance models. Exceptional agreement between the analytic and sample covariances were seen in all cases. To summarize these results, contours of the total standard deviation $\sigma_{\br}$ [see Eq.~\eqref{eq:TotalSTD}] are shown for the LOST method in Fig.~\ref{fig:uranus_std_OPT} and for the DLT method in Fig.~\ref{fig:uranus_std_DLT}. The contours for the HS method are visually indistinguishable from the LOST method, and so a separate figure is not necessary.  Indeed, the covariance for the HS and LOST methods are identical to machine precision for the analytic case and the total standard deviation agrees to better than 0.1\% everywhere for the numerical case (which is consistent with the expected uncertainty in the numerically computed covariance with a finite number of Monte Carlo samples). 

\begin{figure}[tb!]
    \centering
    \includegraphics[width=1\linewidth]{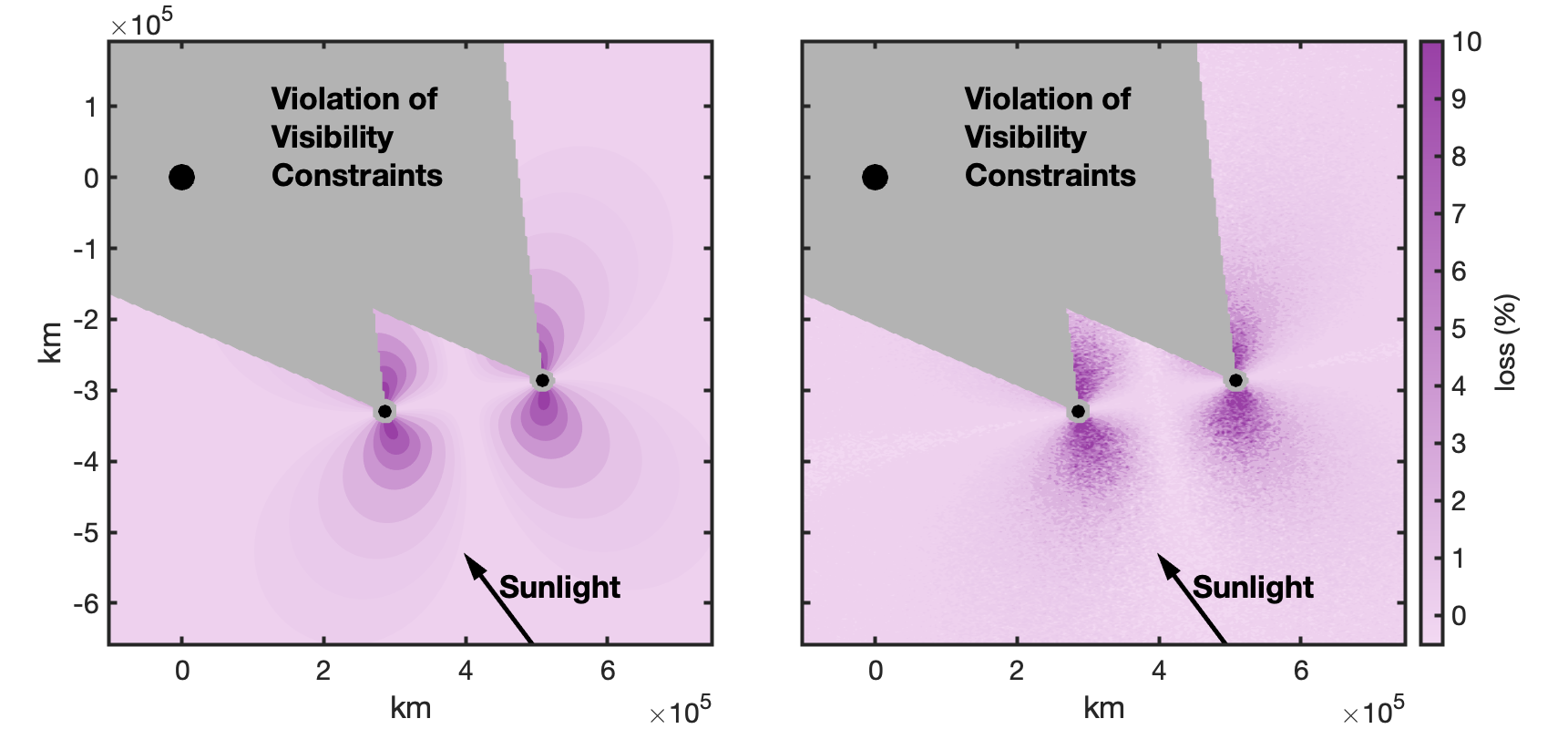}
    \caption{Analytical (left) and Monte Carlo (right) percentage increase in standard deviation (loss in precision) of DLT compared to LOST. The analytical loss never goes under zero.} 
    \label{fig:uranus_rel_std}
\end{figure}

A few observations are immediately apparent from Figs.~\ref{fig:uranus_std_OPT} and \ref{fig:uranus_std_DLT}, and agree with simple intuition. First, triangulation performance is best (localization error is smallest) when the angle between the LOS directions and the baseline $\bd_{12}$ is large. Conversely, performance is worst (localization error is largest) when this angle is small. Second, triangulation performance is best when close to the moons and this performance deteriorates as the distance to the observed moons increases. Third, the analytic covariance for the LOST (and HS) method is always smaller than the DLT method, as can be seen in Fig.~\ref{fig:uranus_rel_std}. The improved performance obtained with LOST as compared to the DLT is only significant when the distances between the observed points are very different.

\begin{figure}[t!]
    \centering
    \includegraphics[width=1\linewidth]{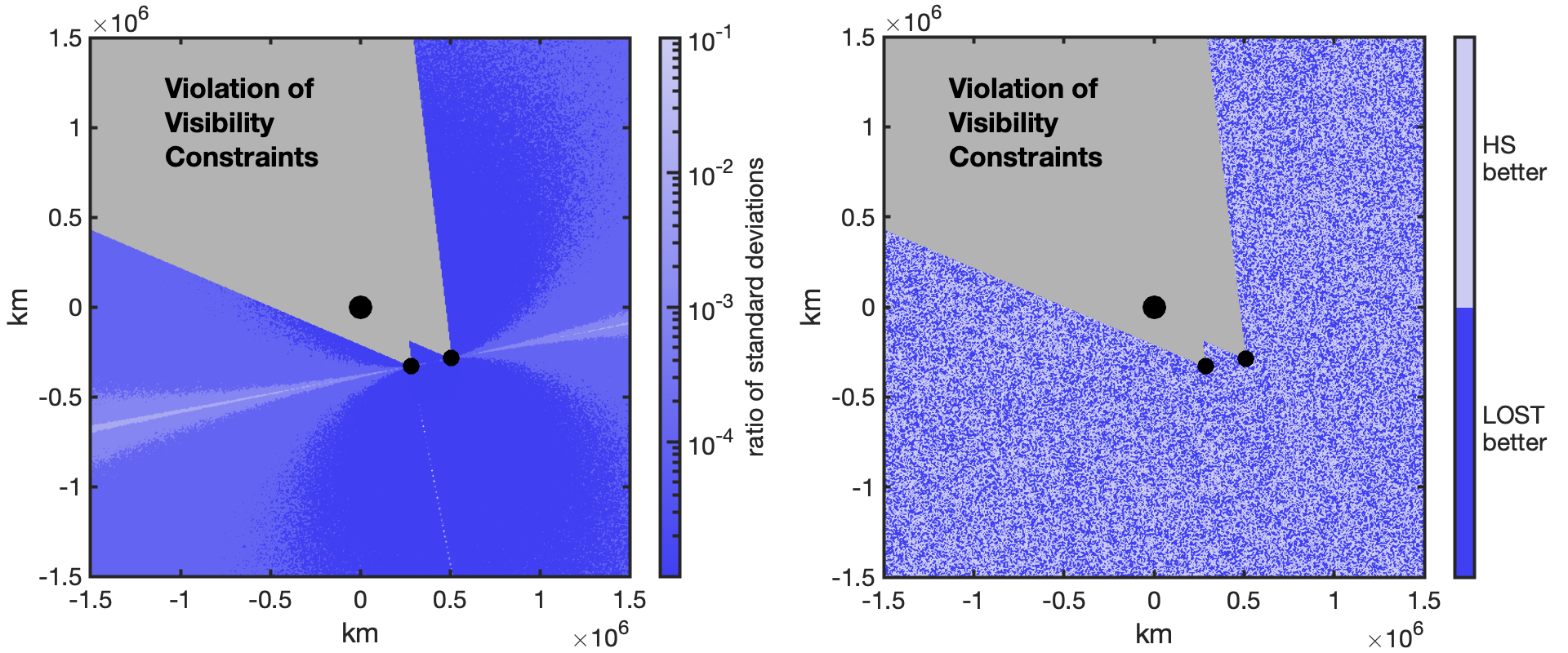}
    \caption{The Hartley and Sturm (HS) and LOST algorithms have equivalent performance. Monte Carlo results show the ratio of $\sigma_{\Delta \br}/\sigma_{LOST}$ (left) and the pattern of where each algorithm performed best (right).}
    \label{fig:uranus_LOST_OPT_comparison}
\end{figure}

The second use of the Monte Carlo analysis is further investigate the statistical equivalence of the HS and LOST methods. The covariance of these two methods has already been shown to be equivalent---to within machine precision for the analytic result and to within the expected error for the Monte Carlo result. Having an equivalent covariance, however, does not imply that the HS and LOST methods provide the exact same estimate for any particular realization of measurement noise. Two additional analyses are preformed to address this question. First, the total standard deviation of the residual $\Delta \br = \br_{LOST} - \br_{HS}$ is computed and then contours of the ratio $\sigma_{\Delta \br} / \sigma_{LOST}$ shown in the left-hand frame of Fig.~\ref{fig:uranus_LOST_OPT_comparison} (for comparison, contours of $\sigma_{LOST}$ are as shown in Fig.~\ref{fig:uranus_std_OPT}). Second, at every point where a Monte Carlo analysis is performed, the number of cases is recorded where each method (LOST or HS) is best. The results are summarized in the right-hand frame of Fig.~\ref{fig:uranus_LOST_OPT_comparison}, where a point is colored dark blue if the LOST algorithm was closer to the truth for more than half of the cases (or light blue if HS was closer to the truth for more than half). Two things are now clear. First, the standard deviation of the residual $\sigma_{\Delta \br}$ is usually 3--4 orders of magnitude smaller than the total standard deviation $\sigma_{\br}$ (and at least one order of magnitude smaller everywhere). Second, there is no apparent structure in which method is closer to the truth, suggesting that there is no viewing geometry that systematically makes the LOST method better (or worse) than HS method. Moreover, each algorithm was found to be ``best'' with about the same frequency (the LOST algorithm was better for about 50.1\% of the points evaluated). Thus, not only do the LOST and HS methods have exactly the same covariance, but neither of the methods is more likely to produce a better estimate than the other. 

While the HS and LOST algorithms have equivalent localization performance, the LOST algorithm has a few notable advantages. First, for the two-measurement case, the solution to LOST takes the form of a small linear system, which is easier (and faster) to solve than the polynomial of degree six in the HS method. Second, the LOST method retains its linear form for $n \geq 3$ LOS measurements---thus allowing for statistically optimal triangulation for a very large number of measurements as the direct solution to a linear system (and without the need for iteratively solving a nonlinear least squares problem). By comparison, the optimal HS method does not scale to more than two LOS measurements. 

\subsection{Three-Dimensional Reconstruction}
\label{Sec:NotreDame}
Shape modeling and 3-D reconstruction is an important problem for the exploration of small bodies in the Solar System \cite{Gaskell:2008,Palmer:2022}. Such a reconstruction may be accomplished by triangulation. Rather than demonstrate the capability on a space object, an example is presented using a standard benchmark dataset. 

\begin{figure}[p!]
    \centering
    \includegraphics[width=1\linewidth]{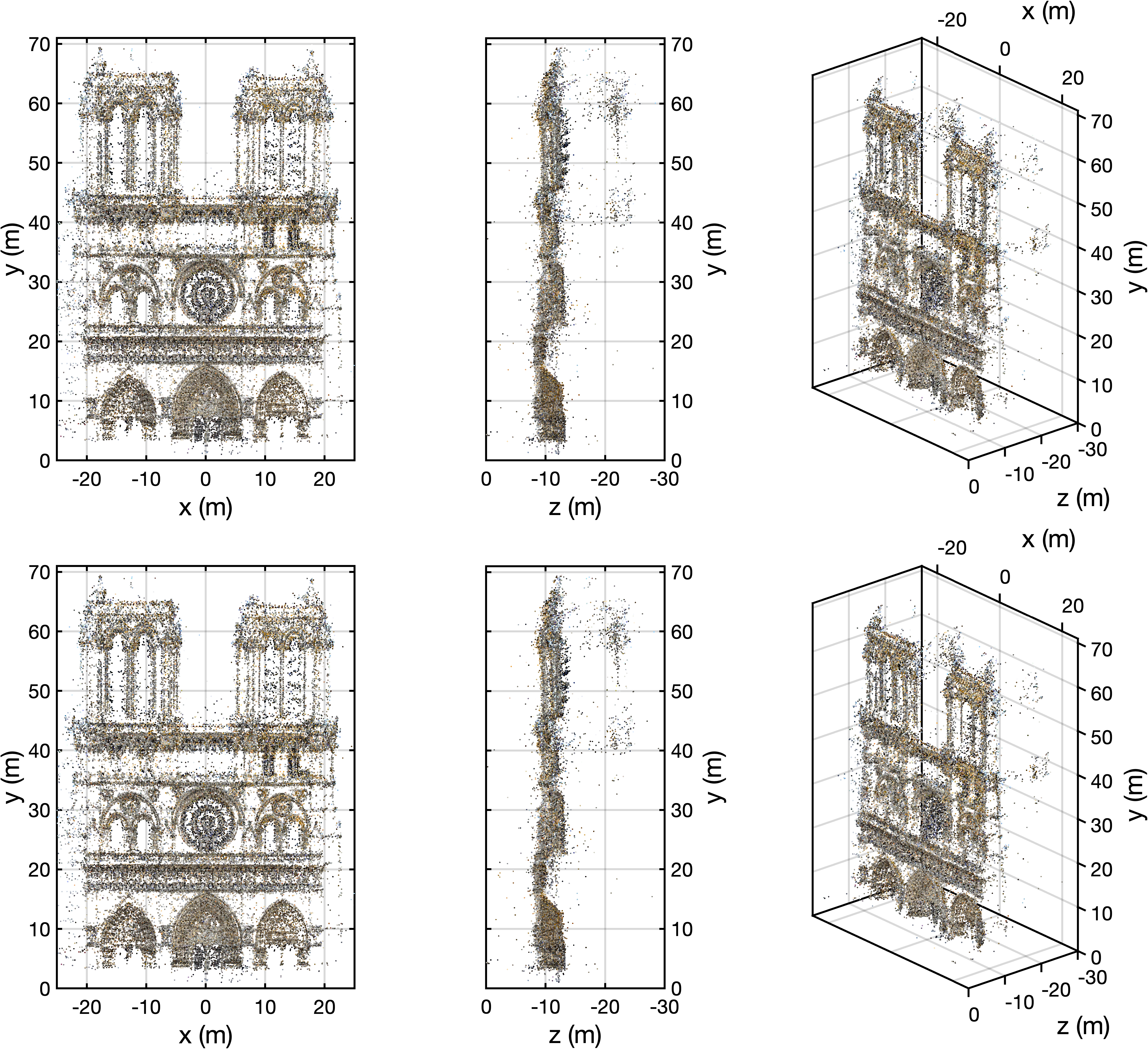}
    \caption{Point cloud reconstruction of the west facade of Notre-Dame de Paris. The top row shows the reference model from \cite{Snavely:2006} while the bottom row is the solution found with the LOST method.}
    \label{fig:notre_dame}
\end{figure}

\begin{figure}[h!]
    \centering
    \includegraphics[width=0.45\linewidth]{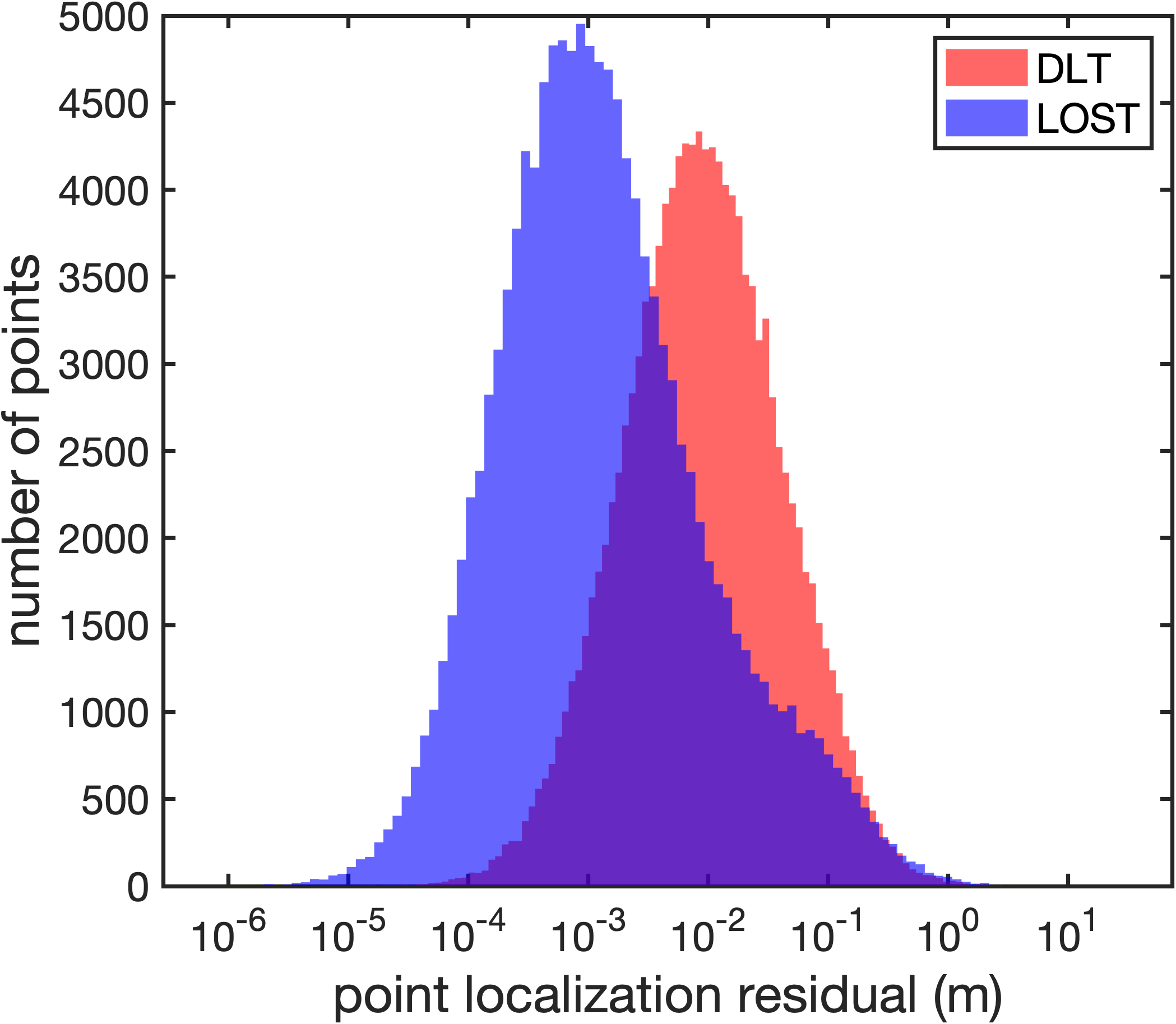}
    \caption{Position residuals for the 3-D reconstruction of Notre-Dame de Pairs using the DLT and LOST methods as compared to the reference model.} 
    \label{fig:notre_dame_histogram}
\end{figure}

The Notre-Dame de Paris cathedral lies on the \^{I}le de la Cit\'{e} in Paris and the main body of the cathedral was built between about 1160 and 1245 \cite{Bruzelius:1987}. Notre-Dame is widely regarded as the preeminent surviving example of French Gothic architecture and is amongst the most visited (and photographed) structures in France. As a result of its popularity, there are a large number of public-domain photographs---making Notre-Dame a natural choice (amongst others) for demonstration of the ground-breaking image-based 3-D reconstruction work by Snavely, Seitz, and Szeliski \cite{Snavely:2006,Snavely:2008}. A 3-D cloud of 127,431 points originating from 715 pictures (and as many cameras) obtained from Ref.~\cite{Snavely:2006}\footnote{Supplementary data for Ref.~\cite{Snavely:2006} vailable from: \url{http://phototour.cs.washington.edu/datasets/}.} may be used to reconstruct the west facade of the cathedral. This is an example of the intersection form of the triangulation problem. The solutions of the DLT and the LOST methods are compared to the reference model included in the benchmark dataset. As illustrated in Fig.~\ref{fig:notre_dame}, the distinction between the reference model and the LOST reconstruction is hardly distinguishable to the naked eye. The histogram presented in Fig.~\ref{fig:notre_dame_histogram} confirms what the eye can (or cannot) see: the majority of the point positions found by LOST are within a centimeter or less from the reference model. Figure~\ref{fig:notre_dame_histogram} also highlights that the residuals for most points are about one order of magnitude lower for the LOST method as compared to the DLT method.

\subsection{Rendezvous, Proximity Operations, and Docking (RPOD)}
\label{Sec:ExampleRelNav}

Rendezvous, Proximity Operations, and Docking (RPOD) is a critical capability for a variety of spaceflight applications \cite{Naasz:2012}, including human exploration \cite{Houbolt:1961,Young:1970,Goodman:2006}, interplanetary exploration \cite{Mattingly:2005}, and satellite servicing \cite{Reed:2016}. When two spacecraft in nearly circular orbits are close to one another, their relative dynamics are nearly linear and may be described by the Clohessy-Wiltshire (CW) equations \cite{Clohessy:1960}. It has been known for some time that the relative state between two particles is unobservable under CW dynamics when only LOS measurements are available \cite{Woffinden:2007,Woffinden:2009}. There are two ways to reintroduce observability during coasting flight (i.e., with no maneuvers). The first is to consider the nonlinear dynamics \cite{Sullivan:2017}. The second is to alter the triangulation geometry. This brief example focuses only on the latter of these these solutions.

Consider the situation where the origin of an LVLH frame is fixed to a chief spacecraft in a circular orbit. The relative motion of a nearby deputy spacecraft in this same LVLH frame is governed by the CW equations having a solution of the form:
\begin{equation}
    \bxi_i = \begin{bmatrix} \br_i \\ \bv_i \end{bmatrix} =
     \begin{bmatrix} \bPhi_{\br \br}(t_i,t_0) & \bPhi_{\br \bv}(t_i,t_0) \\ 
     \bPhi_{\bv \br}(t_i,t_0) & \bPhi_{\bv \bv}(t_i,t_0)\end{bmatrix}
      \begin{bmatrix} \br_0 \\ \bv_0 \end{bmatrix} =
    \bPhi(t_i,t_0) \bxi_0
\end{equation}
where $\br_i$ and $\bv_i$ are LVLH position and velocity of the deputy relative to the chief at time $t_i$. The equations for $\bPhi(t_i,t_0)$ may be found in Ref.~\cite{Clohessy:1960} or a contemporary astrodynamics textbook (e.g., \cite{Vallado:2007}).

Using the notation from Section~\ref{Sec:DynamicTri}, let $\bPhi_\br$ denote the rows of the STM that produce the position vector,
\begin{equation}
    \bPhi_\br(t_i,t_0) =
     \begin{bmatrix} \bPhi_{\br \br}(t_i,t_0) & \bPhi_{\br \bv}(t_i,t_0) \end{bmatrix}
\end{equation}
which is a $3 \times 6$ matrix in this case.

Suppose for a moment that LOS measurements are collected between the chief and deputy spacecraft (the same equations result regardless of the vehicle on which the camera resides). Since the chief spacecraft is at the origin, this implies that $\bp_{I_i} = 0, \; \forall t_i$. Hence, substituting this scenario into Eq.~\eqref{eq:DynamicDLT}, one arrives at
\begin{equation}
    \begin{bmatrix} 
        \left[ \bl_1 \times \right] \bPhi_\br(t_1,t_0) \\
        \vdots \\
        \left[ \bl_n \times \right] \bPhi_\br(t_n,t_0)
    \end{bmatrix}
    \begin{bmatrix} 
        \br_0 \\
        \bv_0
    \end{bmatrix} = \textbf{0}_{3n \times 1}
    \label{eq:RelNavDLTzero}
\end{equation}
Since the result is a homogeneous linear system, it follows that both $\bxi_0$ and $\alpha \bxi_0$ (for $\alpha \neq 0$) provide equally acceptable solutions to the linear system. There are infinitely many homothetic solutions $\alpha \bxi_0$ that represent an expansion or contraction of the figure (i.e., true deputy's trajectory) about the homothetic center (i.e., the chief's location at the origin of the LVLH frame). This is nothing more than a triangulation interpretation of the well-known scale ambiguity associated with angles-only RelNav that is usually shown by other means (e.g.,~Ref.~\cite{Woffinden:2007}).

It is obvious now that scale observability may be introduced by making the right-hand side of Eq.~\eqref{eq:RelNavDLTzero} non-zero. This may be practically achieved by introducing a known offset in the location of either the camera (on the deputy) or the observed point (on the chief). Such a situation may occur when either of the spacecraft are not small when compared to the scale of the relative motion.

To see this, suppose there is a camera aboard the deputy spacecraft with a known offset of $\delta \br_{i}$ at time $t_i$. In this case, the LOS measurement is
\begin{equation}
    \bl_i \propto \bp_i - (\br_i + \delta \br_i)
\end{equation}
and so the DLT equation becomes
\begin{equation}
    \left[ \bl_i \times \right] \br_{i} = \left[ \bl_i \times \right] \left( \bp_{i} - \delta \br_i \right)
\end{equation}
Therefore, after mapping $\br_i$ to time $t_0$,
\begin{equation}
    \left[ \bl_i \times \right] \bPhi(t_i,t_0) \bxi_{0}  = \left[ \bl_i \times \right]\left( \bp_{i} - \delta \br_i \right)
\end{equation}
Letting the localization frame be the LVLH frame (denoted by $L$) and using the normalization $\bl_{C_i} \propto \bK^{-1} \bar{\bu}_i$,
\begin{equation}
    \left[ \bK^{-1} \bar{\bu}_i \times \right] \bT^{L}_{C_i} \bPhi(t_i,t_0) \bxi_{0}  = \left[ \bK^{-1} \bar{\bu}_i \times \right] \bT^{L}_{C_i}\left( \bp_{L_i} - \delta \br_{L_i} \right)
\end{equation}
When more than one measurement is available at more than one time, this expression may be stacked in the usual way to form a linear system
\begin{equation}
    \begin{bmatrix}
    \left[ \bK^{-1} \bar{\bu}_1 \times \right] \bT^{L}_{C_1} \bPhi(t_1,t_0) \\
    \vdots \\
    \left[ \bK^{-1} \bar{\bu}_i \times \right] \bT^{L}_{C_i} \bPhi(t_i,t_0) 
    \end{bmatrix} \bxi_{0} = 
     \begin{bmatrix}
     \left[ \bK^{-1} \bar{\bu}_1 \times \right] \bT^{L}_{C_1}\left( \bp_{L_1} - \delta \br_{L_1} \right) \\
     \vdots \\
     \left[ \bK^{-1} \bar{\bu}_i \times \right] \bT^{L}_{C_i}\left( \bp_{L_i} - \delta \br_{L_i} \right)
     \end{bmatrix} 
\end{equation}
So long as the offsets (and measurement times) are chosen to avoid $\left[ \bK^{-1} \bar{\bu}_i \times \right] \bT^{L}_{C_i}\left( \bp_{C_i} - \delta \br_{C_i} \right) = 0$ for all of the observation times then the problem becomes observable. In most cases this condition is the same as ensuring $\bp_{C_i} \neq \delta \br_{C_i}$. The idea of introducing observability by an offset in just the point aboard the deputy (i.e., $\bp_{C_i} = 0$ and $\delta \br_{C_i} \neq 0$) was explored in Ref.~\cite{Klein:2015}. The result above demonstrates that an equivalent solution could be achieved by introducing an offset on the chief (i.e., $\bp_{C_i} \neq 0$ and $\delta \br_{C_i} = 0$).

\section{Conclusions}
Absolute line of sight (LOS) measurements from images may be used to solve both spacecraft navigation and three-dimensional (3-D) reconstruction problems via triangulation. The mathematical foundations necessary for triangulation have been understood since antiquity, and knowledge of this history enhances the versatility of the modern practitioner. This manuscript reviews the historical evolution of triangulation algorithms that ultimately led to the embodiments of the trigonometric solutions that are in widespread use today. Many variations on these trigonometric solutions [e.g, with the Direct Linear Transform (DLT), collinearity equations, or Pl\"{u}cker coordinates] are shown to be equivalent to one another. In the special case of two LOS observations, the statistically optimal solution of Hartley and Sturm \cite{Hartley:1997} is summarized and the resulting polynomial of degree six is considered. When the two LOS observations come from the same camera (e.g., with monocular navigation), the derivation can be modified to produce a much simpler polynomial of degree two. Moreover, in the general case of many measurements (from one or many cameras), it is shown that a Maximum Likelihood Estimate (MLE) may be developed as the solution to a small linear system (of size $2n \times 3$ for $n$ LOS measurements)---resulting in the so-called Linear Optimal Sine Triangulation (LOST) method. The non-iterative LOST method is shown to provide an equivalent solution to the classical Hartley and Sturm method, but has the comparative advantages of (1) being scalable to many measurements and (2) not requiring the solution to a polynomial of degree six. Finally, many of the triangulation results are also shown to be valid for dynamic systems, especially when the dynamics are linear. The various algorithms and analytic insights are validated through a variety of examples, including planetary terrain relative navigation (TRN), interplanetary optical navigation (OPNAV), 3-D reconstruction of Notre-Dame de Paris, and angles-only relative navigation (RelNav).

\section*{Appendix}
\subsection{Analytic Covariance Expressions}
\subsubsection{Direct Linear Transform}
The DLT is formulated in Eq.~\eqref{eq:DLT_Hxy} as an ordinary (unweighted) least squares problem. Given a problem $\bH \bx = \by$, recall that the covariance for a generic weighting is,
\begin{equation}
    \bP = (\bH^T \bW \bH)^{-1} (\bH^T \bW  \bR \bW^T \bH )  (\bH^T \bW \bH)^{-1} 
\end{equation}
where $\bR$ is the measurement covariance and $\bW$ is the measurement weighting matrix. When $\bW = \bR^{-1}$ the result is the MLE solution. However, when $\bW = \bI_{n \times n}$, the result is the ordinary least squares solution having covariance
\begin{equation}
    \bP = (\bH^T \bH)^{-1} (\bH^T \bR \bH )  (\bH^T \bH)^{-1} 
\end{equation}
Substituting into this notation, it follows from Eq.~\eqref{eq:DLT_Hxy}, that 
\begin{equation}
    \bH = \begin{bmatrix}
        \left[ \bK_1^{-1} \bar{\bu}_1 \times \right] \bT^I_{C_1}  \\
        \vdots \\
        \left[ \bK_n^{-1} \bar{\bu}_n \times \right] \bT^I_{C_n} 
    \end{bmatrix}
\end{equation}
and, therefore,
\begin{equation}
    \bH^T \bH = - \sum_{i=1}^n \bT_I^{C_i} \left[ \bK_i^{-1} \bar{\bu}_1 \times \right]^2 \bT^I_{C_i} 
\end{equation}
Moreover, obtaining $\bR_{\beps_i}$ from Eq.~\eqref{eq:RepsKu2} one finds that
\begin{equation}
\label{eq:DLT_cov}
        \bP_{DLT} = -(\bH^T \bH)^{-1} \left( \sum_{i=1}^n \bT_I^{C_i} \left[ \bK_i^{-1} \bar{\bu}_i \times \right] \tilde{\bR}_{\beps_i} \left[ \bK_i^{-1} \bar{\bu}_i \times \right] \bT^I_{C_i}  \right) (\bH^T \bH)^{-1}
\end{equation}
where everything here is known. The leading negative sign is a result of the identity $[ \, \cdot \, \times]^T = - [ \, \cdot \, \times]$.

\subsubsection{Explicit Range}
Writing an analytic expression for the covariance of the explicit range method is difficult in general due to combinatorial nature of the problem. It is not generally clear which combinations of measurements should be used. Thus, the discussion that follows focuses only on the two-measurement case. Additional attention is not warranted since this method is rarely the preferred approach for triangulation.

To develop the covariance, begin with the expression from Eq.~\eqref{eq:ExpRangeFinalSoln} 
\begin{equation}
    \br_I = \frac{1}{n}\sum_{i=1}^n \bp_{I_i} - \rho_i \ba_{I_i}
\end{equation}
and then compute the differential
\begin{equation}
    \delta \br_I = -\frac{1}{n}\sum_{i=1}^n \delta \rho_i \ba_{I_i} + \rho_i \delta \ba_{I_i}
\end{equation}
For the case of $n=2$, this may be written as
\begin{equation}
    \label{eq:RngCovTerm1}
   -2 \delta \br = 
   \bA_{12}
   \begin{bmatrix}
        \delta \rho_1\\ \delta \rho_2 
      \end{bmatrix} + 
      \bB_{12}
      \begin{bmatrix}
        \delta \ba_{I_1} \\ \delta \ba_{I_2}
      \end{bmatrix} 
\end{equation}
\begin{equation}
   \bA_{12} = 
   \begin{bmatrix}
        \ba_{I_1} & \ba_{I_2}
      \end{bmatrix} 
   \quad \text{and} \quad
    \bB_{12} = 
      \begin{bmatrix}
        \rho_1 \bI_{3 \times 3} &  \rho_2 \bI_{3 \times 3}
      \end{bmatrix}
\end{equation}
Similarly, taking the differential of Eq.~\eqref{eq:RangeLinSys} for $n=2$, one obtains
\begin{equation}
      \begin{bmatrix}
      -1 & \ba^T_{I_1} \ba_{I_2}  \\
      -\ba^T_{I_2} \ba_{I_1} & 1 
      \end{bmatrix}
      \begin{bmatrix}
        \delta \rho_1\\ \delta \rho_2 
      \end{bmatrix}
      = 
      \begin{bmatrix}
        \bd^T_{I_{12}} \delta \ba_{I_1} - \rho_2( \ba^T_{I_2} \delta \ba_{I_1} + \ba^T_{I_1} \delta \ba_{I_2} )\\
        \bd_{I_{12}} \delta \ba_{I_2}  + \rho_1( \ba^T_{I_2} \delta \ba_{I_1} + \ba^T_{I_1} \delta \ba_{I_2} )
      \end{bmatrix}
\end{equation}
which may be rewritten as
\begin{equation}
\begin{bmatrix}
        \label{eq:RngCovTerm2}
        \delta \rho_1\\ \delta \rho_2 
      \end{bmatrix} = \bC_{12}
      \begin{bmatrix}
        \delta \ba_{I_1}\\ \delta \ba_{I_2} 
      \end{bmatrix}
\end{equation}
\begin{equation}
      \bC_{12} = \frac{1}{ (\ba^T_{I_1} \ba_{I_2})^2 - 1 }
      \begin{bmatrix}
      1 & -\ba^T_{I_1}\ba_{I_2}  \\
      \ba^T_{I_2} \ba_{I_1} & -1 
      \end{bmatrix}
      \begin{bmatrix}
        \bd^T_{I_{12}} - \rho_2 \ba^T_{I_2} & -\rho_2 \ba^T_{I_1} \\
        \rho_1 \ba^T_{C_2} &
        \bd_{I_{12}}  + \rho_1 \ba_{I_1}
      \end{bmatrix}
\end{equation}
Substituting Eq.~\eqref{eq:RngCovTerm2} into Eq.~\eqref{eq:RngCovTerm1}, it follows that
\begin{equation}
    \label{eq:RngCovTerm3}
   \delta \br = -\frac{1}{2}
   ( \bA_{12} \bC_{12} + \bB_{12})
      \begin{bmatrix}
        \delta \ba_{C_1}\\ \delta \ba_{C_2} 
      \end{bmatrix}
\end{equation}
Also, computing $\partial \ba_{C_i} / \partial \bar{\bx}_i$, observe that
\begin{equation}
        \label{eq:RngCovTerm4}
      \begin{bmatrix}
        \delta \ba_{C_1}\\ \delta \ba_{C_2} 
      \end{bmatrix} = 
      -\bD_{12}
      \begin{bmatrix}
        \delta \bar{\bx}_{1}\\ \delta \bar{\bx}_{2} 
      \end{bmatrix}
\end{equation}
\begin{equation}
      \bD_{12} = 
      \begin{bmatrix}
      \| \bar{\bx}^T_1 \|^{-1}\bT^{C_1}_I \left[ \ba_{C_1} \times \right]^2 & \textbf{0}_{3 \times 3} \\
      \textbf{0}_{3 \times 3} & \| \bar{\bx}^T_2 \|^{-1}\bT^{C_2}_I \left[ \ba_{C_2} \times \right]^2 
      \end{bmatrix}
\end{equation}
Which, after substitution of Eq.~\eqref{eq:RngCovTerm4} into Eq.~\eqref{eq:RngCovTerm3}, yields
\begin{equation}
   \delta \br = \frac{1}{2}
   ( \bA_{12} \bC_{12} + \bB_{12}) \bD_{12}
      \begin{bmatrix}
        \delta \bar{\bx}_{1}\\ \delta \bar{\bx}_{2} 
      \end{bmatrix}
\end{equation}
And, finally,
\begin{equation}
   \bP_{RNG} = E[\delta \br \, \delta \br^T] = \frac{1}{4}
   ( \bA_{12} \bC_{12} + \bB_{12}) \bD_{12}
      \bR_{\bar{\bx}} \bD^T_{12} ( \bA_{12} \bC_{12} + \bB_{12})^T
\end{equation}

\subsubsection{Hartley and Sturm Polynomial}
To develop the analytic covariance for the Hartley and Sturm method, begin with the measurement model from Eq.~\eqref{eq:FPMeasModel}
\begin{equation}
   \bu_i =  \frac{\bS \bK_i \bT^I_{C_i} (\bp_{I_i} - \br_I) }{ \bk^T \bK_i \bT^I_{C_i}  (\bp_{I_i} - \br_I) }
\end{equation}
Since this method (as modified here) explicitly minimizes the covariance-weighted reprojection error, one obtains a covariance
\begin{equation}
   \bP_{\br} = \left[ \sum_{i=1}^2 \bA^T_i \bR^{-1}_{\bu_i} \bA_i  \right]^{-1}
\end{equation}
where
\begin{equation}
   \bA_i = \frac{\partial \bu_i}{\partial \br_I} = \frac{-\bS \bK_i \bT^I_{C_i}}{\bk^T \bK_i  \bT^I_{C_i}  (\bp_{I_i} - \br_I)} \left( \bI_{3 \times 3} - \frac{ (\bp_{I_i} - \br_{I}) \bk^T \bK_i  \bT^I_{C_i}}{\bk^T \bK_i \bT^I_{C_i}  (\bp_{I_i} - \br_I)}\right)
\end{equation}
This may be simplified somewhat by considering the structure of $\bA_i$ and $\bR_{\bu_i}$. First, recognize that $\bk^T \bK_i = \bk^T$. Also, making use of the identity $(\ba^T \bb) \bI_{3\times3} - \bb \ba^T = [\ba \times][\bb \times]$, one observes that $\bA_i$ becomes
\begin{equation}
     \bA_i = \frac{1}{ \gamma^2_i } \bS \bK_i \bT^I_{C_i} \left[ \bT^{C_i}_I \bk \times \right] \left[ (\bp_{I_i}-\br_I) \times \right] 
\end{equation}
where $\gamma_i$ is from Eq.~\eqref{eq:Defgamma} and may also be written as $\gamma_i = \bk^T \bT^I_{C_i}  (\bp_{I_i} - \br_I)$.
Further, recognizing that $\bar{\bx}_i = \bT^I_{C_i}(\bp_{I_i} - \br_I)/\gamma_i$,
\begin{equation}
     \bA_i = \frac{1}{ \gamma_i } \bS \bK_i \bT^I_{C_i} \left[ \bT^{C_i}_I \bk \times \right] \left[ \bT^{C_i}_i \bar{\bx}_i \times \right] 
\end{equation}
Factoring out the rotation $\bT^I_{C_i}$
\begin{equation}
     \bA_i = \frac{1}{ \gamma_i } \bS \bK_i \left[ \bk \times \right] \left[ \bar{\bx}_i \times \right] \bT^I_{C_i}
\end{equation}
If $\bR^{-1}_{\bu_i} = \sigma^{-2}_{\bu_i} \bI_{2 \times 2}$, substituting into
\begin{equation}
     \bP^{-1}_{\br} = \sum_{i=1}^n \frac{1}{ \sigma^2_{\bx_i} \gamma^2_i } \bT^{C_i}_I \left[ \bar{\bx}_i \times \right] \left[ \bk \times \right] \bK^T_i \bS^T \bS  \bK_i \left[ \bk \times \right] \left[ \bar{\bx}_i \times \right] \bT^I_{C_i}
\end{equation}
Suppose that pixels are square ($d_x = d_y$) and that there is no detector skewness ($\alpha=0$) when forming the camera calibration matrix $\bK_i$. Under this condition, note that $\left[ \bk \times \right] \bK_i^T \bS^T \bS \bK_i \left[ \bk \times \right] = - d^2_{x_i} \bS^T \bS$. Moreover, since $\bar{\bx}_i = \bK^{-1}_i \bar{\bu}_i$,
\begin{equation}
     \bP^{-1}_{\br} = -\sum_{i=1}^n \frac{d^2_{x_i}}{ \sigma^2_{\bu_i} \gamma^2_i } \bT^{C_i}_I \left[ \bK^{-1}_i \bar{\bu}_i  \times \right] \bS^T \bS  \left[ \bK^{-1}_i \bar{\bu}_i \times \right] \bT^I_{C_i}
\end{equation}
Recognizing that $\sigma_{\bx_i} = \sigma_{\bu_i}/d_x$,
this becomes
\begin{equation}
     \bP^{-1}_{\br} = -\sum_{i=1}^n \frac{1}{ \sigma^2_{\bx_i} \gamma^2_i } \bT^{C_i}_I \left[ \bK^{-1}_i \bar{\bu}_i  \times \right] \bS^T \bS  \left[ \bK^{-1}_i \bar{\bu}_i \times \right] \bT^I_{C_i}
\end{equation}
which is identical to Eq.~\eqref{eq:FinalCovLOSApp}. Hence, the analytic covariance for the HS and LOST methods is the same when $\bR_{\bu_i} = \sigma^{2}_{\bu_i} \bI_{2 \times 2} = d_{x_i}^2 \sigma_{\bx_i}^2 \bI_{2 \times 2} = d_{x_i}^2 \bR_{\bx_i}$.

\subsubsection{LOST}
The analytic LOST covariance is given from Eq.~\eqref{eq:optDLT_cov}, repeated here as
\begin{equation}
\label{eq:CovLOST}
     \bP_{\br} = - \left( \sum_{i=1}^n \bT_I^{C_i}\left[ \bK^{-1}_i \bar{\bu} \times \right]   \bR^{\dagger}_{\beps_i}  \left[ \bK^{-1}_i \bar{\bu} \times \right] \bT^I_{C_i} \right)^{-1}
\end{equation}
The complicating term for analytic analysis is $\bR^{\dagger}_{\beps_i}$. This is addressed first. Beginning with the definition of $\bR_{\beps_i}$ from Eq.~\eqref{eq:RepsKu2}, which is written compactly as
\begin{equation}
    \bR_{\beps_i} = -\gamma^2_i \left[  \bK^{-1}_i \bar{\bu}_i \times \right] \bR_{\bar{\bx}_i} \left[  \bK^{-1}_i \bar{\bu}_i \times \right]
\end{equation}
where $\gamma_i$ is from Eq.~\eqref{eq:Defgamma}. Under the case when $\bR_{\bar{\bx}} =\sigma^2_{\bar{\bx}_i} \bS^T \bS$, observe that
\begin{equation}
    \bR^{\dagger}_{\beps_i} = \frac{1}{\sigma^2_{\bar{\bx}_i} \gamma^2_i \| \bK^{-1}_i \bar{\bu}_i \|^4  } \left[  \bK^{-1}_i \bar{\bu}_i \times \right]^2 \bS^T \bS \left[  \bK^{-1}_i \bar{\bu}_i \times \right]^2
\end{equation}
Substituting this into the Eq.~\eqref{eq:CovLOST} leads to
\begin{equation}
     \bP^{-1}_{\br} = -\sum_{i=1}^n \frac{1}{\sigma^2_{\bar{\bx}_i} \gamma^2_i \| \bK^{-1}_i \bar{\bu}_i \|^4  } \bT_I^{C_i} \left[  \bK^{-1}_i \bar{\bu}_i \times \right]^3 \bS^T \bS \left[  \bK^{-1}_i \bar{\bu}_i \times \right]^3 \bT^I_{C_i} 
\end{equation}
which simplifies to 
\begin{equation}
    \label{eq:FinalCovLOSApp}
     \bP^{-1}_{\br} = -\sum_{i=1}^n \frac{1}{\sigma^2_{\bar{\bx}_i} \gamma^2_i  } \bT_I^{C_i} \left[  \bK^{-1}_i \bar{\bu}_i \times \right] \bS^T \bS \left[  \bK^{-1}_i \bar{\bu}_i \times \right] \bT^I_{C_i} 
\end{equation}

\subsection{Equivalence of the Trigonometric Solutions}
When there are two LOS measurements ($n=2$) the DLT and explicit range methods are equivalent. To begin, one can analytically find the position estimate by the explicit range as
\begin{equation}
    \br_I = \frac{1}{2}\sum^2_{i=1}(\bp_i-\rho_i \ba_i)
\end{equation}
\begin{equation}
    \br_I = 
    \frac{1}{2}(\bp_1 + \bp_2) -
    \frac{1}{2}
    \begin{bmatrix}
      \ba_{I_1} & \ba_{I_2}
    \end{bmatrix}
    \begin{bmatrix}
      \rho_1 \\ \rho_2
    \end{bmatrix}
\end{equation}
\begin{equation}
\label{eq:pos_range_2}
\begin{aligned}
    \hat{\br}_I &= \frac{1}{2}(\bp_1 + \bp_2) -
    \frac{1}{2}
    \begin{bmatrix}
      \ba_{I_1} & \ba_{I_2}
    \end{bmatrix}
    \left( 
        \frac{1}{ (\ba^T_{I_1} \ba_{I_2})^2 - 1 }
      \begin{bmatrix}
      1 & -\ba^T_{I_1}\ba_{I_2}  \\
      \ba^T_{I_2} \ba_{I_1} & -1 
      \end{bmatrix}
      \begin{bmatrix}
        \ba^T_{I_1} \bd_{I_{12}} \\
        \ba^T_{_I2} \bd_{I_{12}}
      \end{bmatrix}
    \right)\\
    &= \frac{\left((\ba^T_{I_1} \ba_{I_2})^2 - 1 \right)(\bp_1 + \bp_2) 
    + \ba_{I_1} \left( \ba_{I_1}^T\ba_{I_2}\ba_{I_2}^T(\bp_2-\bp_1) - \ba_{I_1}^T (\bp_2-\bp_1)\right)
    + \ba_{I_2}\left( \ba_{I_2}^T (\bp_2-\bp_1)      -\ba_{I_2}^T\ba_{I_1}\ba_{I_1}^T(\bp_2-\bp_1)\right)}{2 \left((\ba^T_{I_1} \ba_{I_2})^2 - 1 \right)} \\
    &= \frac{\left((\ba^T_{I_1} \ba_{I_2})^2 - 1 \right)(\bp_1 + \bp_2) 
    - \ba_{I_1} \ba_{I_1}^T (\bp_2-\bp_1) + \ba_{I_2} \ba_{I_2}^T (\bp_2-\bp_1) 
    + \ba_{I_1} \ba_{I_1}^T\ba_{I_2}\ba_{I_2}^T(\bp_2-\bp_1) 
    - \ba_{I_2} \ba_{I_2}^T\ba_{I_1}\ba_{I_1}^T(\bp_2-\bp_1)}{2 \left((\ba^T_{I_1} \ba_{I_2})^2 - 1 \right)}
    \end{aligned}
\end{equation}

Now, consider the two-LOS version of the DLT from Eq.~\eqref{eq:DLT_LS_NoFrame}. Choose to describe the LOS measurement by $\bl_i = \alpha_i \ba_i$, where $\alpha_i \neq 0$ is the scaling associated with the unit vector LOS $\ba_i$, such that
\begin{equation}
    \begin{bmatrix}
        \alpha_1 \left[ \ba_1  \times \right] \\
        \alpha_2 \left[ \ba_2  \times \right] 
    \end{bmatrix}
    \br = 
    \begin{bmatrix}
        \alpha_1 \left[ \ba_1 \times \right]\bp_{1} \\
        \alpha_2 \left[ \ba_2 \times \right]\bp_{2}
    \end{bmatrix}
\end{equation}
Solving this linear system in the least squares sense for $\br$ yields
\begin{equation}
    \label{eq:TwoMeasDLT1}
    \hat{\br} =  \left(\sum_{i=1}^2 \alpha_i^2 \left[ \ba_i  \times \right]^2 \right)^{-1} 
    \sum_{i=1}^2 \alpha_i^2 \left[ \ba_i  \times \right]^2 \bp_i
\end{equation}
Recalling that that if $\ba$ is a $3 \times 1$ unit vector then $-\left[\ba \times \right]^2 = \bI_{3 \times 3} - \ba \ba^T$, then one may write
\begin{equation}
\begin{aligned}
    \frac{\left(\sum_{i=1}^2 \alpha_i^2 \left[ \ba_i  \times \right]^2 \right)}{-(\alpha_1^2 + \alpha_2^2)}
    &= \bI_{3 \times 3} \underbrace{- \frac{\alpha_1^2}{\alpha_1^2 + \alpha_2^2}}_{\beta_1} \ba^T_{I_1} \ba_{I_1} \underbrace{- \frac{\alpha_2^2}{\alpha_1^2 + \alpha_2^2}}_{\beta_2}  \ba^T_{I_2} \ba_{I_2}
\end{aligned}
\end{equation}

The matrices $\ba^T_{i} \ba_{i}$ are both rank one, and it is possible to analytically invert this sum of matrices with the Sherman-Morrison formula (applied twice). It is also useful to consider that $1+\beta_1=-\beta_2$, so that the inverse finally becomes
\begin{equation}
\label{eq:inverseofsum}
\begin{aligned}
\left(\frac{\left(\sum_{i=1}^2 \alpha_i^2 \left[ \ba_i  \times \right]^2 \right)}{-(\alpha_1^2 + \alpha_2^2)}\right)^{-1} &= 
\bI_{3 \times 3} - \beta_1\frac{\ba_{I_1} \ba^T_{I_1}}{1+\beta_1} - \frac{\beta_2 \left( \bI_{3 \times 3} - \beta_1\frac{\ba_{I_1} \ba^T_{I_1}}{1+\beta_1} \right) \ba_{I_2} \ba^T_{I_2} \left( \bI_{3 \times 3} - \beta_1\frac{\ba_{I_1} \ba^T_{I_1}}{1+\beta_1} \right)}{1+\beta_2 \ba^T_{I_2} \left( \bI_{3 \times 3} - \beta_1\frac{\ba_{I_1} \ba^T_{I_1}}{1+\beta_1} \right)\ba_{I_2}}
\end{aligned}
\end{equation}
When the two measurements are equally weighted (i.e., when $\alpha_1 = \alpha_2$), one finds that $\beta_1=\beta_2=-1/2$, so the expression in Eq.~\eqref{eq:inverseofsum} becomes
\begin{equation}
    \label{eq:TwoMeasDLTSimpleInverse}
    \left(\frac{\left(\sum_{i=1}^2 \alpha_i^2 \left[ \ba_i  \times \right]^2 \right)}{-(\alpha_1^2 + \alpha_2^2)}\right)^{-1} = 
    - \frac{
    -\left((\ba^T_{I_1} \ba_{I_2})^2-1\right)(\ba_{I_1} \ba^T_{I_1}+\bI_{3 \times 3})+(\ba_{I_1} \ba^T_{I_1}+\bI_{3 \times 3}) \ba_{I_2} \ba^T_{I_2} (\ba_{I_1} \ba^T_{I_1}+\bI_{3 \times 3})
    }{
    (\ba^T_{I_1} \ba_{I_2})^2-1
    }
\end{equation}
Expanding Eq.~\eqref{eq:TwoMeasDLT1}, the position given by the DLT is therefore
\begin{equation}
    \hat{\br} = 
     \frac{1}{-(\alpha_1^2 + \alpha_2^2)}\left(\frac{\left(\sum_{i=1}^2 \alpha_i^2 \left[ \ba_i  \times \right]^2 \right)}{-(\alpha_1^2 + \alpha_2^2)}\right)^{-1}
    \left(\alpha_1^2 \left[ \ba_{I_1}  \times \right]^2 \bp_1 + \alpha_2^2 \left[ \ba_{I_2}  \times \right]^2 \bp_2 \right)
\end{equation}
which, under the assumption of $\alpha_1 = \alpha_2$ (i.e., $\beta_1 = \beta_2 = -1/2$) and after substitution from Eq.~\eqref{eq:TwoMeasDLTSimpleInverse}, becomes
\begin{equation}
\label{eq:pos_DLT_2}
    \hat{\br} = 
     \frac{\left((\ba^T_{I_1} \ba_{I_2})^2-1\right) (\bp_2+\bp_1) - \ba_{I_1} \ba_{I_1}^T (\bp_2-\bp_1) + \ba_{I_2} \ba_{I_2}^T (\bp_2-\bp_1)
    + \ba_{I_1} \ba_{I_1}^T \ba_{I_2} \ba_{I_2}^T (\bp_2-\bp_1) - \ba_{I_2} \ba_{I_2}^T \ba_{I_1} \ba_{I_1}^T (\bp_2-\bp_1)}{2 \left((\ba^T_{I_1} \ba_{I_2})^2-1\right)}
\end{equation}
Thus, the equally-weighted unit vector DLT [Eq.~\eqref{eq:pos_DLT_2}] and the explicit range [Eq.~\eqref{eq:pos_range_2}] give the same position for two LOS observations.


\section*{Acknowledgments}
The authors thank Peter Sturm for insightful conversations about triangulation and his feedback on this manuscript.

\bibliography{main}

\begin{thebibliography}{122}
\newcommand{\enquote}[1]{``#1''}
\providecommand{\natexlab}[1]{#1}
\providecommand{\url}[1]{\texttt{#1}}
\providecommand{\urlprefix}{URL }
\expandafter\ifx\csname urlstyle\endcsname\relax
  \providecommand{\doi}[1]{\discretionary{}{}{}https://doi.org/#1}\else
  \providecommand{\doi}[1]{\discretionary{}{}{}\urlstyle{rm}\url{https://doi.org/#1}}\fi

\bibitem[{Christian et~al.(2019)Christian, {McMahon}, {DellaGuistina}, Ernst,
  Russell, Golish, {McCabe}, S, Lauretta, Gay, and Schoch}]{Christian:2019b}
Christian, J.~A., {McMahon}, J., {DellaGuistina}, D.~N., Ernst, C.~M., Russell,
  R.~P., Golish, D.~R., {McCabe}, J.~S., S, K., Lauretta, D.~S., Gay, P., and
  Schoch, P., \enquote{{Resolved Imagery as a Tool for Space Science and
  Exploration},} \emph{{NASA Exploration Science Forum}}, 2019.

\bibitem[{Gaskell et~al.(2008)Gaskell, {Barnouin-Jha}, Scheers, and {et
  al.}}]{Gaskell:2008}
Gaskell, R.~W., {Barnouin-Jha}, O.~S., Scheers, D.~J., and {et al.},
  \enquote{{Characterizing and Navigating Small Bodies with Imaging Data},}
  \emph{Meteoritics \& Planetary Science}, Vol.~43, No.~6, 2008, pp.
  1049--1061.
\newblock \doi{10.1111/j.1945-5100.2008.tb00692.x}.

\bibitem[{Palmer and {et al.}(2022)}]{Palmer:2022}
Palmer, E.~E., and {et al.}, \enquote{Practical Stereophotoclinometry for
  Modeling Shape and Topography on Planetary Missions,} \emph{The Planetary
  Science Journal}, Vol.~3, 2022.
\newblock \doi{10.3847/PSJ/ac460f}.

\bibitem[{Owen et~al.(2008)Owen, Duxbury, Acton, Synnott, Riedel, and
  Bhaskaran}]{Owen:2008}
Owen, W., Duxbury, T., Acton, C., Synnott, S., Riedel, J., and Bhaskaran, S.,
  \enquote{{A Brief History of Optical Navigation at JPL},} \emph{AAS Guidance
  and Control Conference}, 2008.

\bibitem[{Christian(2021)}]{Christian:2021}
Christian, J.~A., \enquote{{A Tutorial on Horizon-Based Optical Navigation and
  Attitude Determination with Space Imaging Systems},} \emph{{IEEE Access}},
  2021, pp. 19,819--19,853.
\newblock \doi{10.1109/ACCESS.2021.3051914}.

\bibitem[{Driedger and Ferguson(2021)}]{Driedger:2021}
Driedger, M., and Ferguson, P., \enquote{Feasibility Study of an Orbital
  Navigation Filter Using Resident Space Object Observations,} \emph{Journal of
  Guidance, Control, and Dynamics}, Vol.~44, No.~3, 2021, pp. 622--628.
\newblock \doi{10.2514/1.G005210}.

\bibitem[{Bradley et~al.(2020)Bradley, Olikara, Bhsakaran, and
  Young}]{Bradley:2020}
Bradley, N., Olikara, Z., Bhsakaran, S., and Young, B., \enquote{{Cislunar
  Navigation Accuracy Using Optical Observations of Natural and Artificial
  Targets},} \emph{{Journal of Spacecraft and Rockets}}, Vol.~57, No.~4, 2020,
  pp. 777--792.
\newblock \doi{10.2514/1.A34694}.

\bibitem[{Karimi and Mortari(2015)}]{Karimi:2015}
Karimi, R.~R., and Mortari, D., \enquote{{Interplanetary Autonomous Navigation
  Using Visible Planets},} \emph{{Journal of Guidance, Control, and Dynamics}},
  Vol.~38, No.~6, 2015, pp. 1151--1156.
\newblock \doi{10.2514/1.G000575}.

\bibitem[{Broschart et~al.(2019)Broschart, Bradley, and
  Bhsakaran}]{Broschart:2019}
Broschart, S.~B., Bradley, N., and Bhsakaran, S., \enquote{{Kinematic
  Approximation of Position Accuracy Achieved Using Optical Observations of
  Distant Asteroids},} \emph{{Journal of Spacecraft and Rockets}}, Vol.~56,
  No.~5, 2019, pp. 1383--1392.
\newblock \doi{10.2514/1.A34354}.

\bibitem[{Hinga(2020)}]{Hinga:2020}
Hinga, M.~B., \enquote{{Cis-Lunar Autonomous Navigation via Implementation of
  Optical Asteroid Angle-Only Measurements},} \emph{21st Advanced Maui Optical
  and Space Surveillance Technologies (AMOS) Conference}, {Maui, HI}, 2020.

\bibitem[{Franzese and Topputo(2022)}]{Franzese:2022}
Franzese, V., and Topputo, F., \enquote{Deep-Space Optical Navigation
  Exploiting Multiple Beacons,} \emph{The Journal of the Astronautical
  Sciences}, 2022.
\newblock \doi{10.1007/s40295-022-00303-5}.

\bibitem[{Cheng et~al.(2003)Cheng, Johnson, Matthies, and Olson}]{Cheng:2003}
Cheng, Y., Johnson, A., Matthies, L., and Olson, C., \enquote{Optical Landmark
  Detection for Spacecraft Navigation,} \emph{AAS/AIAA Astrodynamics Specialist
  Conference}, 2003.

\bibitem[{Woffinden and Geller(2009)}]{Woffinden:2009}
Woffinden, D.~C., and Geller, D.~K., \enquote{Observability Criteria for
  Angles-Only Navigation,} \emph{IEEE Transactions on Aerospace and Electronic
  Systems}, Vol.~45, No.~3, 2009, pp. 1194--1208.
\newblock \doi{10.1109/TAES.2009.5259193}.

\bibitem[{McKee et~al.(2022)McKee, Kowalski, and Christian}]{McKee:2022}
McKee, P., Kowalski, J., and Christian, J.~A., \enquote{Navigation and star
  identification for an interstellar mission,} \emph{Acta Astronautica}, Vol.
  192, 2022, pp. 390--401.
\newblock \doi{10.1016/j.actaastro.2021.12.007}.

\bibitem[{Hartley and Sturm(1997)}]{Hartley:1997}
Hartley, R.~I., and Sturm, P., \enquote{Triangulation,} \emph{Computer Vision
  and Image Understanding}, Vol.~68, No.~2, 1997, pp. 146--157.
\newblock \doi{10.1006/cviu.1997.0547}.

\bibitem[{Stew\'{e}nius et~al.(2005)Stew\'{e}nius, Schaffalitzky, and
  Nist\'{e}r}]{Stewenius:2005}
Stew\'{e}nius, H., Schaffalitzky, F., and Nist\'{e}r, D., \enquote{How Hard is
  3-View Triangulation Really?} \emph{IEEE International Conference on Computer
  Vision (ICCV’05)}, 2005.
\newblock \doi{10.1109/ICCV.2005.115}.

\bibitem[{Chace and Manning(1927)}]{Chace:1927}
Chace, A.~B., and Manning, H.~P., \emph{{The Rhind Mathematical Papyrus,
  British Museum 10057 and 10058, Vol. 1: Free Translation and Commentary}},
  Mathematical Association of America, Oberlin, OH, 1927, pp. 96--99.
\newblock \urlprefix\url{https://hdl.handle.net/2027/mdp.39015017389894}.

\bibitem[{Molinsky(2015)}]{Molinsky:2015}
Molinsky, M., \enquote{Some Original Sources for Modern Tales of Thales --- The
  Tale of the Pyramids,} \emph{Convergence}, 2015.

\bibitem[{Euclid et~al.(1570)Euclid, Billingsley, Dee, Candale, and
  Day}]{EuclidElements}
Euclid, Billingsley, H., Dee, J., Candale, F. D.~F., and Day, J., \emph{{The
  elements of geometrie of the most auncient philosopher Evclide of Megara}},
  Imprinted at London: By Iohn Daye, 1570.
\newblock \urlprefix\url{https://www.loc.gov/item/03020856/}.

\bibitem[{Archibald(1950)}]{Archibald:1950}
Archibald, R.~C., \enquote{The First Translation of Euclid's Elements into
  English and its Source,} \emph{The American Mathematical Monthly}, Vol.~57,
  No. 7P1, 1950, pp. 443--452.
\newblock \doi{10.1080/00029890.1950.11990256}.

\bibitem[{Toomer(1974{\natexlab{a}})}]{Toomer:1974b}
Toomer, G.~J., \enquote{The Chord Table of Hipparchus and the Early History of
  Greek Trigonometry,} \emph{Centarus}, Vol.~18, No.~1, 1974{\natexlab{a}}, pp.
  6--28.
\newblock \doi{10.1111/j.1600-0498.1974.tb00205.x}.

\bibitem[{{Ptolemy active 2nd century} and Toomer(1984)}]{Almagest}
{Ptolemy active 2nd century}, and Toomer, G.~J., \emph{{Ptolemy's Almagest,
  translated and annotated by G.J. Toomer}}, Springer-Verlag, New-York, 1984.

\bibitem[{Toomer(1974{\natexlab{b}})}]{Toomer:1974}
Toomer, G.~J., \enquote{Hipparchus on the Distances of the Sun and Moon,}
  \emph{Archive for History of Exact Sciences}, Vol.~14, No.~2,
  1974{\natexlab{b}}, pp. 126--142.

\bibitem[{Lorch(2002)}]{Lorch:2002}
Lorch, R., \enquote{Greek-Arabic-Latin: The Transmission of Mathematical Texts
  in the Middle Ages,} \emph{Science in Context}, Vol.~14, No. 1--2, 2002, pp.
  313--331.
\newblock \doi{10.1017/S0269889701000114}.

\bibitem[{Frisius(1533)}]{Frisius:1533}
Frisius, G., \emph{Libellus de locorum describendorum ratione}, 1533.

\bibitem[{Pogo(1935)}]{Pogo:1935}
Pogo, A., \enquote{Gemma Frisius, His Method of Determining Differences of
  Longitude by Transporting Timepieces (1530), and His Treatise on
  Triangulation (1533),} \emph{Isis}, Vol.~22, No.~2, 1935, pp. 469--506.

\bibitem[{Haasbroek(1968)}]{Haasbroek:1968}
Haasbroek, N.~D., \emph{Gemma Frisius, Tycho Brahe and Snellius and Their
  Triangulations}, Rijkscommissie voor Geodesie, Delft, Netherlands, 1968.

\bibitem[{{Snellius}(1617)}]{Snellius:1617}
{Snellius}, W., \emph{De Terrae ambitus vera quantitate}, 1617.
\newblock \urlprefix\url{https://mdz-nbn-resolving.de/details:bsb10806698}.

\bibitem[{Breitenberger(1984)}]{Breitenberger:1984}
Breitenberger, E., \enquote{Gauss's geodesy and the axiom of parallels,}
  \emph{Archive for History of Exact Sciences}, Vol.~31, No.~3, 1984, pp.
  273--289.
\newblock \doi{10.1007/BF00327704}.

\bibitem[{Gauss(1828)}]{Gauss:1828}
Gauss, C.~F., \emph{{Supplementum theoriae combinationis observationum
  erroribus minimis obnoxiae}}, typis Dieterichianis, G\"{o}ttingen, 1828.
\newblock \doi{10.3931/e-rara-2858}.

\bibitem[{{McCaw}(1918)}]{McCaw:1918}
{McCaw}, G.~T., \enquote{Resection in Survey,} \emph{The Geographical Journal},
  Vol.~52, No.~2, 1918, pp. 105--123.
\newblock \doi{10.2307/1779558}.

\bibitem[{Bernard et~al.(2018)Bernard, Nafi, and Geller}]{Bernard:2018}
Bernard, A.~J., Nafi, A.~M., and Geller, D.~K., \enquote{Using Triangulation in
  Optical Orbit Determination,} \emph{AAS/AIAA Astrodynamics Specialist
  Conference}, 2018.

\bibitem[{Chen et~al.(2021)Chen, Liu, Li, and Kang}]{Chen:2021}
Chen, L., Liu, C., Li, Z., and Kang, Z., \enquote{A New Triangulation Algorithm
  for Positioning Space Debris,} \emph{Remote Sensing}, Vol.~13, No.~23, 2021.
\newblock \doi{10.3390/rs13234878}.

\bibitem[{Pollefeys et~al.(2000)Pollefeys, Koch, Vergauwen, and {Van
  Gool}}]{Pollefeys:2000}
Pollefeys, M., Koch, R., Vergauwen, M., and {Van Gool}, L., \enquote{Automated
  reconstruction of 3D scenes from sequences of images,} \emph{{ISPRS} Journal
  of Photogrammetry and Remote Sensing}, Vol.~55, No.~4, 2000, pp. 251--267.
\newblock \doi{10.1016/S0924-2716(00)00023-X}.

\bibitem[{Snavely et~al.(2008)Snavely, Seitz, and Szeliski}]{Snavely:2008}
Snavely, N., Seitz, S.~M., and Szeliski, R., \enquote{{Modeling the World from
  Internet Photo Collections},} \emph{{International Journal of Computer
  Vision}}, Vol.~80, 2008, pp. 189--210.
\newblock \doi{10.1007/s11263-007-0107-3}.

\bibitem[{Poirot and {McWilliams}(1976)}]{Poirot:1976}
Poirot, J.~L., and {McWilliams}, G.~V., \enquote{{Navigation by Back
  Triangulation},} \emph{{IEEE Transactions on Aerospace and Electronic
  Systems}}, Vol.~12, No.~2, 1976, pp. 270--274.

\bibitem[{Feissel and Mignard(1998)}]{Feissel:1998}
Feissel, M., and Mignard, F., \enquote{The adoption of ICRS on 1 January 1998:
  meaning and consequences,} \emph{Astronomy and Astrophysics}, Vol. 331, 1998,
  pp. L33--L36.

\bibitem[{Ma et~al.(1998)Ma, Arias, Eubanks, Fey, Gontier, Jacobs, Sovers,
  Archinal, and Charlot}]{Ma:1998}
Ma, C., Arias, E., Eubanks, T., Fey, A., Gontier, A., Jacobs, C., Sovers, O.,
  Archinal, B., and Charlot, P., \enquote{The International Celestial Reference
  Frame as Realized by Very Long Baseline Interferometry,} \emph{The
  Astronomical Journal}, Vol. 116, 1998, pp. 516--546.
\newblock \doi{10.1086/300408}.

\bibitem[{Fey et~al.(2015)Fey, Gordon, and {et al.}}]{Fey:2015}
Fey, A., Gordon, D., and {et al.}, \enquote{The Second International Celestial
  Reference Frame by Very Long Baseline Interferometry,} \emph{The Astronomical
  Journal}, Vol. 150, 2015.
\newblock \doi{10.1088/0004-6256/150/2/58}.

\bibitem[{Kaplan(2011)}]{Kaplan:2011}
Kaplan, G.~H., \enquote{Angles-Only Navigation: Position and Velocity Solution
  from Absolute Triangulation,} \emph{Navigation: Journal of The Institute of
  Navigation}, Vol.~58, No.~3, 2011, pp. 187--201.
\newblock \doi{10.1007/s11263-008-0152-6}.

\bibitem[{Liebe(1995)}]{Liebe:1995}
Liebe, C.~C., \enquote{{Star Trackers for Attitude Determination},} \emph{IEEE
  Aerospace and Electronic Systems Magazine}, Vol.~10, No.~6, 1995, pp. 10--16.
\newblock \doi{10.1109/62.387971}.

\bibitem[{Liebe(2002)}]{Liebe:2002}
Liebe, C.~C., \enquote{{Accuracy Performance of Star Trackers --- A Tutorial},}
  \emph{IEEE Transactions on Aerospace and Electronic Systems}, Vol.~38, No.~2,
  2002, pp. 587--599.
\newblock \doi{10.1109/TAES.2002.1008988}.

\bibitem[{Christian and Crassidis(2021)}]{Christian:2021star}
Christian, J.~A., and Crassidis, J.~L., \enquote{Star Identification and
  Attitude Determination with Projective Cameras,} \emph{IEEE Access}, Vol.~9,
  2021, pp. 25,768--25,794.
\newblock \doi{10.1109/ACCESS.2021.3054836}.

\bibitem[{{El Hassan}(2002)}]{Hassan:2002}
{El Hassan}, I.~M., \enquote{Two-Dimensional Resection---A Survey Of Analytical
  Techniques,} \emph{Australian Surveyor}, Vol.~47, No.~1, 2002, pp. 14--23.
\newblock \doi{10.1080/00050356.2002.10558838}.

\bibitem[{Grunert(1841)}]{Grunert:1841}
Grunert, J.~A., \enquote{{Das Pothenot'sche Problem in erweiterter Gestalt;
  nebst Bemerkungen über seine Anwendungen in der Geodisie},} \emph{Archiv der
  Mathematik und Physik}, Vol.~1, 1841, pp. 238--248.

\bibitem[{Haralick et~al.(1994)Haralick, Lee, Ottenberg, and
  N\"{o}lle}]{Haralick:1994}
Haralick, R.~M., Lee, C., Ottenberg, K., and N\"{o}lle, M., \enquote{Review and
  Analysis of Solutions of the Three Point Perspective Pose Estimation
  Problem,} \emph{International Journal of Computer Vision}, Vol.~13, No.~3,
  1994, pp. 331--356.
\newblock \doi{10.1007/BF02028352}.

\bibitem[{Gao et~al.(2003)Gao, Hou, Tang, and Cheng}]{Gao:2003}
Gao, X., Hou, X., Tang, J., and Cheng, H., \enquote{Complete solution
  classification for the perspective-three-point problem,} \emph{{IEEE
  Transactions on Pattern Analysis and Machine Intelligence}}, Vol.~25, No.~8,
  2003, pp. 930--943.
\newblock \doi{10.1109/TPAMI.2003.1217599}.

\bibitem[{Tanygin(2014)}]{Tanygin:2014}
Tanygin, S., \enquote{Closed-Form Solution for Lost-In-Space Visual Navigation
  Problem,} \emph{Journal of Guidance, Control, and Dynamics}, Vol.~37, No.~6,
  2014, pp. 1754--1766.
\newblock \doi{10.2514/1.G000529}.

\bibitem[{{Pascual-Escudero} et~al.(2021){Pascual-Escudero}, Nayak, Briot,
  Kermorgant, Martinet, {Safey El Din}, and Chaumette}]{PascualEscudero:2021}
{Pascual-Escudero}, B., Nayak, A., Briot, S., Kermorgant, O., Martinet, P.,
  {Safey El Din}, M., and Chaumette, F., \enquote{Complete Singularity Analysis
  for the Perspective-Four-Point Problem,} \emph{International Journal of
  Computer Vision}, Vol. 129, 2021, pp. 1217--1237.
\newblock \doi{10.1007/s11263-020-01420-0}.

\bibitem[{Hartley and Zisserman(2003)}]{Hartley:2003}
Hartley, R., and Zisserman, A., \emph{Multiple View Geometry, 2nd Ed.},
  Cambridge University Press, Cambridge, UK, 2003, pp. 70--73,312--321.

\bibitem[{Quan and Lan(1999)}]{Quan:1999}
Quan, L., and Lan, Z., \enquote{Linear N-Point Camera Pose Estimation,}
  \emph{IEEE Transactions on Pattern Analysis and Machine Intelligence},
  Vol.~21, No.~8, 1999, pp. 774--780.
\newblock \doi{10.1109/34.784291}.

\bibitem[{Ansar and Daniilidis(2003)}]{Ansar:2003}
Ansar, A., and Daniilidis, K., \enquote{Linear Pose Estimation from Points or
  Lines,} \emph{{IEEE Transactions on Pattern Analysis and Machine
  Intelligence}}, Vol.~25, No.~5, 2003, pp. 578--589.
\newblock \doi{10.1109/TPAMI.2003.1195992}.

\bibitem[{Lepetit et~al.(2009)Lepetit, {Moreno-Noguer}, and Fua}]{Lepetit:2009}
Lepetit, V., {Moreno-Noguer}, F., and Fua, P., \enquote{EP$n$P: An Accurate
  $O(n)$ Solution to the P$n$P Problem,} \emph{International Journal of
  Computer Vision}, Vol.~81, 2009, pp. 155--166.
\newblock \doi{10.1007/s11263-008-0152-6}.

\bibitem[{Calhoun and Dabney(1995)}]{Calhoun:1995}
Calhoun, P.~C., and Dabney, R., \enquote{Solution to the problem of determining
  the relative 6 DOF state for spacecraft automated rendezvous and docking,}
  \emph{Proc. SPIE}, Vol. 2466, 1995, pp. 175--184.
\newblock \doi{10.1117/12.211505}.

\bibitem[{Crassidis et~al.(2000)Crassidis, Alonso, and
  Junkins}]{Crassidis:2000}
Crassidis, J.~L., Alonso, R., and Junkins, J.~L., \enquote{{Optimal Attitude
  and Position Determination from Line-of-Sight Measurements},} \emph{{The
  Journal of the Astronautical Sciences}}, Vol.~48, 2000, pp. 391--408.
\newblock \doi{10.1007/BF03546286}.

\bibitem[{Besl and {McKay}(1992)}]{Besl:1992}
Besl, P.~J., and {McKay}, N.~D., \enquote{A method for registration of 3-D
  shapes,} \emph{IEEE Transactions on Pattern Analysis and Machine
  Intelligence}, Vol.~14, No.~2, 1992, pp. 239--256.
\newblock \doi{10.1109/34.121791}.

\bibitem[{Yang et~al.(2021)Yang, Shi, and Carlone}]{Yang:2021}
Yang, H., Shi, J., and Carlone, L., \enquote{TEASER: Fast and Certifiable Point
  Cloud Registration,} \emph{IEEE Transactions on Robotics}, Vol.~37, No.~2,
  2021, pp. 314--333.
\newblock \doi{10.1109/TRO.2020.3033695}.

\bibitem[{Nardone et~al.(1984)Nardone, Lindgren, and Gong}]{Nardone:1984}
Nardone, S.~C., Lindgren, A.~G., and Gong, K.~F., \enquote{Fundamental
  properties and performance of conventional bearings-only target motion
  analysis,} \emph{IEEE Transactions on Automatic Control}, Vol.~29, No.~9,
  1984, pp. 775--787.
\newblock \doi{10.1109/TAC.1984.1103664}.

\bibitem[{Harper(1955)}]{Harper:1955}
Harper, R.~B., \enquote{Pilot Boat Identification, Harbor Radar Systems,}
  \emph{Navigation}, Vol.~4, No.~5, 1955, pp. 195--201.
\newblock \doi{10.1002/j.2161-4296.1955.tb00164.x}.

\bibitem[{Casey(1888)}]{Casey:1888}
Casey, J., \emph{A sequel to the first six books of the Elements of Euclid,
  containing an easy introduction to modern geometry, with numerous examples},
  Hodges, Figgis \& co., 1888, pp. 173--175.

\bibitem[{Lindgren and Gong(1977)}]{Lindgren:1977}
Lindgren, A.~G., and Gong, K.~F., \enquote{Position and Velocity Estimation Via
  Bearing Observations,} \emph{IEEE Transactions on Aerospace and Electronic
  Systems}, Vol.~14, No.~4, 1977, pp. 564--577.
\newblock \doi{10.1109/TAES.1978.308681}.

\bibitem[{Nardone and Aidala(1981)}]{Nardone:1981}
Nardone, S.~C., and Aidala, V.~J., \enquote{Observability Criteria for
  Bearings-Only Target Motion Analysis,} \emph{IEEE Transactions on Aerospace
  and Electronic Systems}, Vol.~17, No.~2, 1981, pp. 162--166.
\newblock \doi{10.1109/TAES.1981.309141}.

\bibitem[{Clohessy and Wiltshire(1960)}]{Clohessy:1960}
Clohessy, W.~H., and Wiltshire, R.~S., \enquote{Terminal guidance system for
  satellite rendezvous,} \emph{Journal of the Aerospace Sciences}, Vol.~27,
  No.~9, 1960, pp. 653--658.
\newblock \doi{10.2514/8.8704}.

\bibitem[{Woffinden and Geller(2007)}]{Woffinden:2007}
Woffinden, D.~C., and Geller, D.~K., \enquote{Navigating the Road to Autonomous
  Orbital Rendezvous,} \emph{Journal of Spacecraft and Rockets}, Vol.~44,
  No.~4, 2007, pp. 898--909.
\newblock \doi{10.2514/1.30734}.

\bibitem[{Semple and Kneebone(1952)}]{Semple:1952}
Semple, J.~G., and Kneebone, G.~T., \emph{Algebraic Projective Geometry},
  Oxford University Press, Oxford, UK, 1952.

\bibitem[{Gallier(2011)}]{Gallier:2011}
Gallier, J., \emph{Geometric Methods and Applications, 2nd Edition},
  Springer-Verlag, New York, 2011.

\bibitem[{Christian et~al.(2016)Christian, Benhacine, Hikes, and
  {D'Souza}}]{Christian:2016}
Christian, J., Benhacine, L., Hikes, J., and {D'Souza}, C., \enquote{Geometric
  Calibration of the Orion Optical Navigation Camera using Star Field Images,}
  \emph{{The Journal of the Astronautical Sciences}}, Vol.~63, No.~4, 2016, pp.
  335--353.
\newblock \doi{10.1007/s40295-016-0091-3}.

\bibitem[{Bos and {et al.}(2020)}]{Bos:2020}
Bos, B.~J., and {et al.}, \enquote{In-Flight Calibration and Performance of the
  OSIRIS-REx Touch And Go Camera System (TAGCAMS),} \emph{{Space Science
  Reviews}}, Vol.~71, 2020.
\newblock \doi{10.1007/s11214-020-00682-x}.

\bibitem[{Wahba(1965)}]{Wahba:1965}
Wahba, G., \enquote{A Least Square Estimate of Satellite Attitude,} \emph{SIAM
  Review}, Vol.~7, No.~3, 1965, p. 409.
\newblock \doi{10.1137/1007077}.

\bibitem[{Farrell et~al.(1966)Farrell, Stuelpnagel, Wessner, Velman, and
  Brook}]{Farrell:1966}
Farrell, J.~L., Stuelpnagel, J.~C., Wessner, R.~H., Velman, J.~R., and Brook,
  J.~E., \enquote{A Least Squares Estimate of Satellite Attitude (Grace
  Wahba),} \emph{SIAM Review}, Vol.~8, No.~3, 1966, pp. 384--386.
\newblock \doi{10.1137/1008080}.

\bibitem[{Shuster and Oh(1981)}]{Shuster:1981}
Shuster, M.~D., and Oh, S.~D., \enquote{Three-Axis Attitude Determination from
  Vector Observations,} \emph{Journal of Guidance and Control}, Vol.~4, No.~1,
  1981, pp. 70--77.
\newblock \doi{10.2514/3.19717}.

\bibitem[{Markley(1988)}]{Markley:1988}
Markley, F.~L., \enquote{Attitude Determination Using Vector Observations and
  the Singular Value Decomposition,} \emph{The Journal of the Astronautical
  Sciences}, Vol.~36, No.~3, 1988, pp. 245--258.

\bibitem[{Shuster(1989)}]{Shuster:1989}
Shuster, M.~D., \enquote{Maximum Likelihood Estimation of Spacecraft Attitude,}
  \emph{The Journal of the Astronautical Sciences}, Vol.~37, No.~1, 1989, pp.
  79--88.

\bibitem[{Mortari and Majji(2009)}]{Mortari:2009}
Mortari, D., and Majji, M., \enquote{Multiplicative Measurement Model,}
  \emph{{The Journal of the Astronautical Sciences}}, Vol.~57, No. 1 \& 2,
  2009, pp. 47--60.
\newblock \doi{10.1007/BF03321493}.

\bibitem[{Kaasalainen and Tanga(2004)}]{Kaasalainen:2004}
Kaasalainen, M., and Tanga, P., \enquote{Photocentre offset in ultraprecise
  astrometry: Implications for barycentre determination and asteroid
  modelling,} \emph{Astronomy and Astrophysics}, Vol. 416, No.~1, 2004, pp.
  367--373.
\newblock \doi{10.1051/0004-6361:20031711}.

\bibitem[{Olds and {et al.}(2022)}]{Olds:2022}
Olds, R.~D., and {et al.}, \enquote{The Use of Digital Terrain Models for
  Natural Feature Tracking at Asteroid Bennu,} \emph{The Planetary Science
  Journal}, Vol.~3, 2022.
\newblock \doi{10.3847/PSJ/ac5184}.

\bibitem[{Lowe(2004)}]{Lowe:2004}
Lowe, D.~G., \enquote{Distinctive Image Features from Scale-Invariant
  Keypoints,} \emph{{International Journal of Computer Vision}}, Vol.~60,
  No.~2, 2004, pp. 91--110.
\newblock \doi{10.1023/B:VISI.0000029664.99615.94}.

\bibitem[{Cheng et~al.(2006)Cheng, Crassidis, and Markley}]{Cheng:2006}
Cheng, Y., Crassidis, J.~L., and Markley, F.~L., \enquote{Attitude Estimation
  for Large Field-of-View Sensors,} \emph{{The Journal of the Astronautical
  Sciences}}, Vol.~54, No. 3 \& 4, 2006, pp. 433--448.
\newblock \doi{10.1007/BF03256499}.

\bibitem[{T\={u}s\={\i}(1891)}]{Tusi:lawofsines}
T\={u}s\={\i}, .-., Nas\={\i}r al-D\={\i}n Muhammad ibn~Muhammad,
  \emph{Trait\`{e} du quadrilat\`{e}re attribué a Nassiruddinel-Toussy,
  d'apr\`{e}s un manuscrit tir\'{e} de la biblioth\`{e}que de S.A. Edhem
  Pacha}, Typographie et lithographie Osmani\`{e}, Turkey, 1891.

\bibitem[{Aydin et~al.(2020)Aydin, Hammoudi, and Bakbouk}]{Al-Kashi:lowofcos}
Aydin, N., Hammoudi, L., and Bakbouk, G., \emph{Al-Kashi's Miftah al-Hisab,
  Volume II: Geometry}, Birkh\"{a}user Cham, 2020, p.~31.
\newblock \doi{10.1007/978-3-030-61330-3}.

\bibitem[{{Abdel-Aziz} and Karara(2015)}]{AbdelAziz:2015}
{Abdel-Aziz}, Y.~I., and Karara, H.~M., \enquote{Direct Linear Transformation
  from Comparator Coordinates into Object Space Coordinates in Close-Range
  Photogrammetry,} \emph{Photogrammetric Engineering \& Remote Sensing},
  Vol.~81, No.~2, 2015, pp. 103--107.
\newblock \doi{10.14358/PERS.81.2.103}.

\bibitem[{Christian(2019)}]{Christian:2019}
Christian, J.~A., \enquote{Autonomous Initial Orbit Determination with Optical
  Observations of Unknown Planetary Landmarks,} \emph{Journal of Spacecraft and
  Rockets}, Vol.~56, No.~1, 2019, pp. 211--220.
\newblock \doi{10.2514/1.A34259}.

\bibitem[{Gennery(1992)}]{Gennery:1992}
Gennery, D.~B., \emph{Calibration and Orientation of Cameras in Computer
  Vision}, Springer, 1992, Chaps. Least-Squares Camera Calibration Including
  Lens Distortion and Automatic Editing of Calibration Points, pp. 47--48.
\newblock \doi{10.1007/978-3-662-04567-1_5}.

\bibitem[{K and Li(2004)}]{Di:2004}
K, D., and Li, R., \enquote{CAHVOR camera model and its photogrammetric
  conversion for planetary applications,} \emph{Journal of Geophysical
  Research}, Vol. 109, 2004.
\newblock \doi{10.1029/2003JE002199}.

\bibitem[{Pl\"{u}cker(1865)}]{Plucker:1865}
Pl\"{u}cker, J., \enquote{On a new geometry of space,} \emph{Philosophical
  Transactions of the Royal Society}, Vol. 155, 1865.
\newblock \doi{10.1098/rstl.1865.0017}.

\bibitem[{Bartoli and Sturm(2005)}]{Bartoli:2005}
Bartoli, A., and Sturm, P., \enquote{Structure-from-motion using lines:
  Representation, triangulation, and bundle adjustment,} \emph{Computer Vision
  and Image Understanding}, Vol. 100, No.~3, 2005, pp. 416--441.
\newblock \doi{10.1016/j.cviu.2005.06.001}.

\bibitem[{Franzese and Topputo(2020)}]{Franzese:2020}
Franzese, V., and Topputo, F., \enquote{{Optimal Beacons Selection for
  Deep-Space Optical Navigation},} \emph{{The Journal of the Astronautical
  Sciences}}, Vol.~67, 2020, pp. 1775--1792.
\newblock \doi{10.1007/s40295-020-00242-z}.

\bibitem[{Shuster(1990)}]{Shuster:1990}
Shuster, M.~D., \enquote{Kalman Filtering of Spacecraft Attitude and the QUEST
  Model,} \emph{The Journal of the Astronautical Sciences}, Vol.~38, No.~3,
  1990, pp. 377--393.

\bibitem[{Fisher(1922)}]{Fisher:1922}
Fisher, R.~A., \enquote{On the mathematical foundations of theoretical
  statistics,} \emph{Philosophical Transactions of the Royal Society of London.
  Series A}, Vol. 222, 1922, pp. 309--368.
\newblock \doi{10.1098/rsta.1922.0009}.

\bibitem[{Tapley et~al.(2004)Tapley, Schutz, and Born}]{Tapley:2004}
Tapley, B., Schutz, B., and Born, G., \emph{Statistical Orbit Determination},
  Elsevier Academic Press, Amsterdam, 2004, pp. 172--173,190--194.
\newblock \doi{10.1016/B978-0-12-683630-1.X5019-X}.

\bibitem[{Hartley(1998)}]{Hartley:1998}
Hartley, R.~I., \enquote{Chirality,} \emph{International Journal of Computer
  Vision}, Vol.~26, No.~1, 1998, pp. 41--61.
\newblock \doi{10.1023/A:1007984508483}.

\bibitem[{Armstrong(1996)}]{Armstrong:1996}
Armstrong, M.~N., \enquote{Self-Calibration from Image Sequences,} Ph.D.
  thesis, University of Oxford, 1996.

\bibitem[{Sturm(1997)}]{Sturm:1997}
Sturm, P.~F., \enquote{Vision 3D Non Calibr\'{e}e: Contributions \`{a} la
  Reconstruction Projective et \'{E}tude des Mouvements Critiques pour
  L'Auti-Calibrage,} Ph.D. thesis, Institut National Polytechnique de Grenoble,
  1997.

\bibitem[{Kobylka and Christian(2021)}]{Kobylka:2021}
Kobylka, K.~R., and Christian, J.~A., \enquote{Methods in Triangulation for
  Image-Based Terrain Relative Navigation,} \emph{AAS/AIAA Space Flight
  Mechanics Meeting}, 2021.

\bibitem[{Goulb and {Van Loan}(2013)}]{Goulb:2013}
Goulb, G.~H., and {Van Loan}, C.~F., \emph{Matrix Computations, 4th Ed.}, Johns
  Hopkins University Press, 2013, pp. 260--268,304--306.

\bibitem[{Aldrich(1997)}]{Aldrich:1997}
Aldrich, J., \enquote{R. A. Fisher and the Making of Maximum Likelihood
  1912--1922,} \emph{Statistical Science}, Vol.~12, No.~3, 1997, pp. 162--176.
\newblock \doi{10.1214/ss/1030037906}.

\bibitem[{Gelb(1974)}]{Gelb:1974}
Gelb, A., \emph{Applied Optimal Estimation}, MIT Press, Cambridge, MA, 1974.

\bibitem[{Carr and Sobek(1980)}]{Carr:1980}
Carr, J.~R., and Sobek, J.~S., \enquote{Digital Scene Matching Area Correlator
  (DSMAC),} \emph{Proc. SPIE}, Vol. 0238, 1980, pp. 36--41.
\newblock \doi{10.1117/12.959130}.

\bibitem[{Golden(1980)}]{Golden:1980}
Golden, J.~P., \enquote{Terrain Contour Matching (TERCOM): A Cruise Missile
  Guidance Aid,} \emph{Proc. SPIE}, Vol. 0238, 1980, pp. 10--18.
\newblock \doi{10.1117/12.959127}.

\bibitem[{Kim and Sukkarieh(2004)}]{Kim:2004}
Kim, J., and Sukkarieh, S., \enquote{Autonomous airborne navigation in unknown
  terrain environments,} \emph{IEEE Transactions on Aerospace and Electronic
  Systems}, Vol.~40, No.~3, 2004, pp. 1031--1045.
\newblock \doi{10.1109/TAES.2004.1337472}.

\bibitem[{Cheng and Miller(2003)}]{Miller:2003}
Cheng, Y., and Miller, J., \enquote{Autonomous landmark based spacecraft
  navigation system,} \emph{AAS/AIAA Astrodynamics Specialist Conference},
  2003.

\bibitem[{{Y. Cheng} and {Ansar}(2005)}]{Cheng:2005}
{Y. Cheng}, and {Ansar}, A., \enquote{Landmark Based Position Estimation for
  Pinpoint Landing on Mars,} \emph{IEEE International Conference on Robotics
  and Automation}, 2005, pp. 1573--1578.
\newblock \doi{10.1109/ROBOT.2005.1570338}.

\bibitem[{Adams et~al.(2008)Adams, Criss, and Shankar}]{Adams:2008}
Adams, D., Criss, T., and Shankar, U., \enquote{Passive Optical Terrain
  Relative Navigation Using APLNav,} \emph{IEEE Aerospace Conference}, 2008,
  pp. 1--9.
\newblock \doi{10.1109/AERO.2008.4526303}.

\bibitem[{Steffes et~al.(2019)Steffes, Monterroza, Benhacine, and
  Mario}]{Steffes:2019}
Steffes, S.~R., Monterroza, F., Benhacine, L., and Mario, C., \enquote{Optical
  Terrain Relative Navigation Approaches to Lunar Orbit, Descent and Landing,}
  \emph{AIAA Scitech Forum}, 2019.
\newblock \doi{10.2514/6.2019-1178}.

\bibitem[{Acton(1996)}]{Acton:1996}
Acton, C., \enquote{{Ancillary data services of NASA's Navigation and Ancillary
  Information Facility},} \emph{Planetary and Space Sciences}, Vol.~44, No.~1,
  1996, pp. 65--70.
\newblock \doi{10.1016/0032-0633(95)00107-7}.

\bibitem[{Acton et~al.(2018)Acton, Bachman, Semenov, and Wright}]{Acton:2018}
Acton, C., Bachman, N., Semenov, B., and Wright, E., \enquote{{A look towards
  the future in the handling of space science mission geometry},}
  \emph{Planetary and Space Sciences}, Vol. 150, 2018, pp. 9--12.
\newblock \doi{10.1016/j.pss.2017.02.013}.

\bibitem[{Johnson et~al.(2017)Johnson, Aaron, Chang, Cheng, Montgomery, Mohan,
  Schroeder, Tweddle, Trawny, and Zheng}]{johnson:2017}
Johnson, A., Aaron, S., Chang, J., Cheng, Y., Montgomery, J., Mohan, S.,
  Schroeder, S., Tweddle, B., Trawny, N., and Zheng, J., \enquote{{The Lander
  Vision System for Mars 2020 Entry Descent and Landing},} \emph{AAS Guidance
  Navigation and Control Conference}, 2017.

\bibitem[{Maki and {et al.}(2020)}]{Maki:2020}
Maki, J.~N., and {et al.}, \enquote{The Mars 2020 Engineering Cameras and
  Microphone on the Perseverance Rover: A Next-Generation Imaging System for
  Mars Exploration,} \emph{Space Science Reviews}, Vol. 216, 2020.
\newblock \doi{10.1007/s11214-020-00765-9}.

\bibitem[{{National Academies of Sciences, Engineering, and
  Medicine}(2022)}]{PlanetaryDecadalSurvey:2022}
{National Academies of Sciences, Engineering, and Medicine}, \emph{Origins,
  Worlds, and Life: A Decadal Strategy for Planetary Science and Astrobiology
  2023-2032}, The National Academies Press, 2022, pp. S--7.
\newblock \doi{10.17226/26522}.

\bibitem[{Jacobson(2014)}]{Jacobson:2014}
Jacobson, R.~A., \enquote{The Orbits of the Uranian Satellites and Rings, the
  Gravity Field of the Uranian System, and the Orientation of the Pole of
  Uranus,} \emph{The Astronomical Journal}, Vol. 148, No.~5, 2014.
\newblock \doi{10.1088/0004-6256/148/5/76}.

\bibitem[{Lubey et~al.(2020)Lubey, Bhaskaran, Bradley, and
  Olikara}]{Lubey:2020}
Lubey, D.~P., Bhaskaran, S., Bradley, N., and Olikara, Z., \enquote{Ice Giant
  Exploration Via Autonomous Optical Navigation,} \emph{AAS/AIAA Astrodynamics
  Specialist Conference}, 2020.

\bibitem[{Bruzelius(1987)}]{Bruzelius:1987}
Bruzelius, C., \enquote{{The Construction of Notre-Dame in Paris},} \emph{{The
  Art Bulletin}}, Vol.~69, 1987, pp. 540--569.
\newblock \doi{10.1080/00043079.1987.10788458}.

\bibitem[{Snavely et~al.(2006)Snavely, Seitz, and Szeliski}]{Snavely:2006}
Snavely, N., Seitz, S.~M., and Szeliski, R., \enquote{Photo tourism: exploring
  photo collections in 3D,} \emph{ACM SIGGRAPH}, 2006.
\newblock \doi{10.1145/1179352.1141964}.

\bibitem[{Naasz and Moreau(2015)}]{Naasz:2012}
Naasz, B., and Moreau, M., \enquote{{Autonomous RPOD Challenges for the Coming
  Decade},} \emph{AAS Guidance Navigation and Control Conference}, 2015.

\bibitem[{Houbolt(1962)}]{Houbolt:1961}
Houbolt, J.~C., \enquote{Problems and Potentialities of Space Rendezvous,}
  \emph{Space Flight and Re-Entry Trajectories}, Springer, Vienna, 1962, pp.
  406--429.
\newblock \doi{10.1007/978-3-7091-5470-0_6}.

\bibitem[{Young and Alexander(1970)}]{Young:1970}
Young, K.~A., and Alexander, J.~D., \enquote{Apollo Lunar Rendezvous,}
  \emph{Journal of Spacecraft and Rockets}, Vol.~7, No.~9, 1970, pp.
  1083--1086.
\newblock \doi{10.2514/3.30106}.

\bibitem[{Goodman(2006)}]{Goodman:2006}
Goodman, J.~L., \enquote{History of Space Shuttle Rendezvous and Proximity
  Operations,} \emph{Journal of Spacecraft and Rockets}, Vol.~43, No.~5, 2006,
  pp. 944--959.
\newblock \doi{10.2514/1.19653}.

\bibitem[{Mattingly et~al.(2005)Mattingly, HAyati, and
  Udomkesmalee}]{Mattingly:2005}
Mattingly, R., HAyati, S., and Udomkesmalee, G., \enquote{Technology
  development plans for the Mars Sample Return mission,} 2005.
\newblock \doi{10.1109/AERO.2005.1559389}.

\bibitem[{Reed et~al.(2016)Reed, Smith, Naasz, and Pellegrino}]{Reed:2016}
Reed, B.~B., Smith, R.~C., Naasz, B., and Pellegrino, J., \enquote{The
  Restore-L Servicing Mission,} \emph{AIAA SPACE 2016 Forum}, 2016.
\newblock \doi{10.2514/6.2016-5478}.

\bibitem[{Sullivan and {D'Amico}(2017)}]{Sullivan:2017}
Sullivan, J., and {D'Amico}, S., \enquote{Nonlinear Kalman Filtering for
  Improved Angles-Only Navigation Using Relative Orbital Elements,}
  \emph{Journal of Guidance, Control, and Dynamics}, Vol.~40, No.~9, 2017, pp.
  2183--2200.
\newblock \doi{10.2514/1.G002719}.

\bibitem[{Vallado(2007)}]{Vallado:2007}
Vallado, D., \emph{Fundamentals of Astrodynamics and Applications, 3rd Ed.},
  Microcosm Press, Hawthorne, CA, 2007, pp. 389--417.

\bibitem[{Klein and Geller(2015)}]{Klein:2015}
Klein, I., and Geller, D., \enquote{Zero $\Delta$V Solution to the Angles-Only
  Range Observability Problem during Orbital Proximity Operations,}
  \emph{Itzhack Y. Bar-Itzhack Memorial Symposium on Estimation, Navigation,
  and Spacecraft Control}, 2015, pp. 351--369.
\newblock \doi{10.1007/978-3-662-44785-7_19}.

\end{thebibliography}

\end{document}